\documentclass[10pt]{article} 
\usepackage[accepted]{tmlr}

\usepackage{amsmath,amsfonts,bm}

\def\eqref#1{equation~\ref{#1}}

\def\1{\bm{1}}

\DeclareMathAlphabet{\mathsfit}{\encodingdefault}{\sfdefault}{m}{sl}
\SetMathAlphabet{\mathsfit}{bold}{\encodingdefault}{\sfdefault}{bx}{n}

\usepackage{hyperref}
\usepackage{url}
\usepackage{longtable,booktabs,array}
\usepackage{multirow}
\usepackage{calc} 
\usepackage[pdftex]{graphicx}
\graphicspath{ {./figures/} }
\usepackage{verbatim}
\usepackage{subfigure}

\title{Privacy Risks and Preservation Methods in Explainable \\Artificial Intelligence: A Scoping Review}

\author{\name Sonal Allana \email sallana@uoguelph.ca \\
      \addr School of Computer Science\\
      University of Guelph
      \AND
      \name Mohan Kankanhalli \email mohan@comp.nus.edu.sg \\
      \addr School of Computing\\
      National University of Singapore
      \AND
      \name Rozita Dara \email drozita@uoguelph.ca\\
      \addr School of Computer Science\\
      University of Guelph}

\def\month{10}  
\def\year{2025} 
\def\openreview{\url{https://openreview.net/forum?id=q9nykJfzku}} 

\begin{document}

\maketitle

\begin{abstract}
Explainable Artificial Intelligence (XAI) has emerged as a pillar of Trustworthy AI and aims to bring transparency in complex models that are opaque by nature. Despite the benefits of incorporating explanations in models, an urgent need is found in addressing the privacy concerns of providing this additional information to end users. In this article, we conduct a scoping review of existing literature to elicit details on the conflict between privacy and explainability. Using the standard methodology for scoping review, we extracted 57 articles from 1,943 studies published from January 2019 to December 2024. The review addresses 3 research questions to present readers with more understanding of the topic: (1) what are the privacy risks of releasing explanations in AI systems? (2) what current methods have researchers employed to achieve privacy preservation in XAI systems? (3) what constitutes a privacy preserving explanation? Based on the knowledge synthesized from the selected studies, we categorize the privacy risks and preservation methods in XAI and propose the characteristics of privacy preserving explanations to aid researchers and practitioners in understanding the requirements of XAI that is privacy compliant. Lastly, we identify the challenges in balancing privacy with other system desiderata and provide recommendations for achieving privacy preserving XAI. We expect that this review will shed light on the complex relationship of privacy and explainability, both being the fundamental principles of Trustworthy AI.
\end{abstract}

\section{Introduction}
\subsection{Paradigm shift in technology and the need for explanations}
Traditional software development processes have metamorphosed into
stable and reliable frameworks through decades of fine tuning by
software experts. These software systems are built on human designed
algorithms and produce a trace of the logic used to generate an output.
Even in complex systems, it is possible for software experts to analyze
the logic and generate an explanation for a specific result. During the
software development lifecycle, engineers focus on creating the
algorithm and validating using well designed test cases that closely
replicate real world scenarios. In contrast, modern AI systems do not
have an underlying human-written algorithm and learn from data fed to
them. This data-driven nature creates dependence of the system on data
quality \citep{merhiAssessmentBarriersImpacting2022} and introduces problems such as lack of fairness
when data is biased, or irrelevant results when data is incomplete or
outdated \citep{trocinResponsibleAIDigital2021}. During the AI development phase,
engineers access training datasets which may contain personally
identifiable or sensitive information about individuals. For neural
network systems, the development process often involves a trial-and-error
approach, where high accuracy is targeted by tweaking the
hyperparameters such as the learning rate, epochs, number of layers or
activation functions. The lack of an algorithm prevents engineers from
tracing through the AI system and interpreting the results. Thus, the
basic ability to be explainable and understand input-output behaviors,
which is critical to all computer systems \citep{sundararajanAxiomaticAttributionDeep2017},
is often out of reach of AI systems. Explanations for outcomes of AI are crucial in
high-risk applications \citep{mochaourabDemonstratorCounterfactualExplanations2023} in domains such as
healthcare \citep{dharChallengesDeepLearning2023,dwivediExplainableAIXAI2023,yangUnboxBlackboxMedical2022}, finance \citep{dwivediExplainableAIXAI2023, zhangExplainableArtificialIntelligence2022}, defense \citep{dwivediExplainableAIXAI2023}, justice \citep{deeksJudicialDemandExplainable2019}, energy and power \citep{machlevExplainableArtificialIntelligence2022} where the impact on human life and well-being is
significant \citep{karimiSurveyAlgorithmicRecourse2023, mcdermidArtificialIntelligenceExplainability2021,nassarBlockchainExplainableTrustworthy2020}  and the inability to do so deters their successful implementation \citep{nassarBlockchainExplainableTrustworthy2020,shrikumarLearningImportantFeatures2017,yangUnboxBlackboxMedical2022}.

Trustworthy AI strives to mitigate risks due to possible harms from the data-driven nature of AI systems. Trustworthiness is based
on foundation principles of reliability, validity, robustness, privacy,
explainability and fairness \citep{alzubaidiRiskFreeTrustworthyArtificial2023, tabassiAIRiskManagement2023} to boost user confidence in the system
outputs. Among these principles, explainability aims to bring the
much-needed transparency in opaque models and can be considered as a
non-functional requirement of a software system to mitigate opacity
\citep{chazetteExploringExplainabilityDefinition2021}. There are numerous benefits of including
explanations in AI models. Besides aiding data scientists in getting a
better understanding of the data \citep{hohmanGamutDesignProbe2019} and performing
required data cleansing \citep{chenEVFLExplainableVertical2022}, explanations can help
developers in detecting errors in input and determining features that
can be modified to change the outcome \citep{dattaAlgorithmicTransparencyQuantitative2016}. When
multiple models are available with similar accuracy, an explanation
method can help to choose between models \citep{dhurandharExplanationsBasedMissing2018}.
Interpretable models can enable knowledge discovery by detecting
knowledge or patterns that were missed by uninterpretable ones \citep{kimExamplesAreNot2016}. Since humans remain an important component in the
decision-making process as end-users and consumers of automated
decisions \citep{terziyanExplainableAIIndustry2022}, explanations can give them an
understanding of the model outcome, especially when they are adversely
affected by the decisions \citep{aliExplainableArtificialIntelligence2023}. Explainability can also facilitate
privacy awareness in end-users \citep{brunotteCanExplanationsSupport2021}, enabling them to
make right choices for their personal data and aid regulators and
compliance officers to understand the compliance of models \citep{mcdermidArtificialIntelligenceExplainability2021} with applicable regulations. With generative AI (Gen-AI) and
large language models (LLMs) entering mainstream, explanations
constitute an important design principle \citep{weiszGeneralDesignPrinciples2023} in
enabling a better mental model for users \citep{sunInvestigatingExplainabilityGenerative2022} and in
communicating its capabilities and limitations to them \citep{weiszGeneralDesignPrinciples2023}. It can also support users in effective prompt engineering to
determine the words that impact the output of a model \citep{mishraPromptAidPromptExploration2023} and in verification of generated content to mitigate the problem
of hallucinations \citep{schneiderExplainableGenerativeAI2024}.

\subsection{Challenges for privacy in explainability}

In many high risk application domains of AI, training models on sensitive personal information is inevitable
for usefulness of these systems \citep{veugenPrivacyPreservingContrastiveExplanations2022}. For instance, a
lung cancer detection model necessitates training on chest X-ray
images, which constitutes personal information of patients. Similarly, a
loan evaluation model of a bank, requires access to the financial
profiles of customers, which is also personal information of individuals. Usage of
personal data impacts the privacy of individuals when they are subject
to intentional or unintentional identification and exposure through
these systems. Some models are found to memorize data contained in the
input \citep{songMachineLearningModels2017} which can be exploited by adversaries for
extraction of personal information. Gen-AI models create new content
from large multi-modal datasets \citep{sunInvestigatingExplainabilityGenerative2022}
which could potentially contain sensitive personal information \citep{meskoImperativeRegulatoryOversight2023}. Due to such privacy risks involved, when personal data is used in training, testing, or inferencing of AI models, they become subject to data regulation and privacy acts \citep{icoExplainingDecisionsMade2020}.

Explainability is a foundational principle of Trustworthy AI, however,
recent research has determined that introducing explanations in AI systems is found to conflict with
the privacy requirements of the system. Explanation interfaces are found to give
adversaries an additional attack surface \citep{dudduInferringSensitiveAttributes2022,liuPleaseTellMe2024} to mine the information contained in the model. Privacy attacks can target explanations to
retrieve information about membership in the training set \citep{liuPleaseTellMe2024,narettoEvaluatingPrivacyExposure2022,shokriPrivacyRisksModel2021}, build surrogates
of the underlying model \citep{aivodjiModelExtractionCounterfactual2020,wangDualCFEfficientModel2022,yanExplanationLeaksExplanationguided2023}, infer sensitive attributes of individuals \citep{dudduInferringSensitiveAttributes2022,luoFeatureInferenceAttack2022} and reconstruct the training set \citep{shokriPrivacyRisksModel2021}. This leakage is demonstrated
across different types of XAI methods including those that are currently
used in commercial production systems. In addition to privacy attacks,
the content of explanations may also inadvertently expose information
that is proprietary \citep{milliModelReconstructionModel2019} and hence valuable and
confidential to organizations \citep{winikoffArtificialIntelligenceRight2021} or sensitive
to individuals, thus causing breach of data and privacy regulations.
Hence researchers have highlighted the urgent need of mitigating privacy
leakage through explanation interfaces \citep{luoFeatureInferenceAttack2022, patelModelExplanationsDifferential2022, yanExplanationLeaksExplanationguided2023}. Due to these concerns of the
privacy vulnerabilities of explanations, necessary privacy preservation
measures are required in XAI systems \citep{aivodjiModelExtractionCounterfactual2020,shokriPrivacyRisksModel2021,zhaoExploitingExplanationsModel2021}.

\subsection{Main contributions}

Previous research has identified that the privacy issues in
explainability are insufficiently studied \citep{liuPleaseTellMe2024,luoFeatureInferenceAttack2022,narettoEvaluatingPrivacyExposure2022} despite its criticality in achieving
safety in AI transparency. To the best of our knowledge, there is
currently no work that provides an in-depth understanding of the
conflict between privacy and explainability in AI. Hence, we focus this
article on these two fundamental desiderata of Trustworthy AI and
explore the landscape of privacy risks and preservation methods proposed
in literature in the context of XAI. The key questions that we have
designed to define the scope of this article are:

RQ1: What are the privacy risks of releasing explanations in AI systems?

RQ2: What current methods have researchers employed to achieve privacy
preservation in XAI systems?

RQ3: What constitutes a privacy preserving explanation?

We conducted a scoping review guided by RQ1 and RQ2. Based on the
knowledge gathered from the extracted studies, we propose
characteristics of privacy preserving XAI and outline them with the help
of practical use cases to answer RQ3. Our main contributions in this
article are as follows:
\begin{itemize}
\item\emph{Categorization of reported privacy risks in XAI:} We review the
  conflict between privacy and explainability in current literature and
  categorize the risks.
\item\emph{Identification of applicable privacy preservation methods in
  XAI:} We determine the privacy preservation methods that are applicable
  to XAI and report the progress achieved by researchers in integrating
  them in XAI systems.
\item\emph{Privacy preserving XAI characteristics:} We propose the
  desirable characteristics of privacy preserving XAI to provide
  researchers and practitioners the guidelines for achieving the trade-off
  between privacy, utility and explainability.
\end{itemize}
The rest of this article is organized as follows. Section~\ref{section2} presents a
brief background on XAI including its definition, evolution,
categorization of explanation approaches and related reviews. In Section~\ref{section3}, we present the details of the scoping review methodology for
extracting studies relevant to our research questions. Sections~\ref{section4} and \ref{section5}
synthesize the results from the scoping review. In Section~\ref{section4}, we
consolidate both intentional and unintentional privacy risks of
explanations to answer RQ1. In Section~\ref{section5}, we elaborate the use of
privacy preserving methods on explanations and the existing works that
utilize them in response to RQ2. Section~\ref{section6} proposes the characteristics
of privacy preserving XAI and answers RQ3. We conclude the article by
discussing the results, and highlight the open issues, challenges, and
recommendations for future work in Section~\ref{section7} and conclusions in Section~\ref{section8}.

\clearpage
\section{Background}\label{section2}

\subsection{Definition of XAI}\label{section2.1}

In 2017, DARPA kickstarted its 4-year XAI program to accelerate research
in the development of explanation methods and interfaces to enhance
understanding and trust of end-users \citep{gunningDARPAExplainableArtificial2019}. The program
defined XAI as ``AI systems that can explain their rationale to a human
user, characterize their strengths and weaknesses, and convey an
understanding of how they will behave in the future'' \citep{gunningDARPAExplainableArtificial2019}. The study established users' preference for systems with
explanations over systems that provided only decisions. \citet{ribeiroWhyShouldTrust2016} refer to explanations of predictions as qualitative artifacts
that provide the relationship between an input instance and the output
prediction.

\subsection{Evolution of XAI and emergence of privacy
concerns}\label{section2.2}

\begin{figure}[h]
\begin{center}
\includegraphics[width=0.8\linewidth]{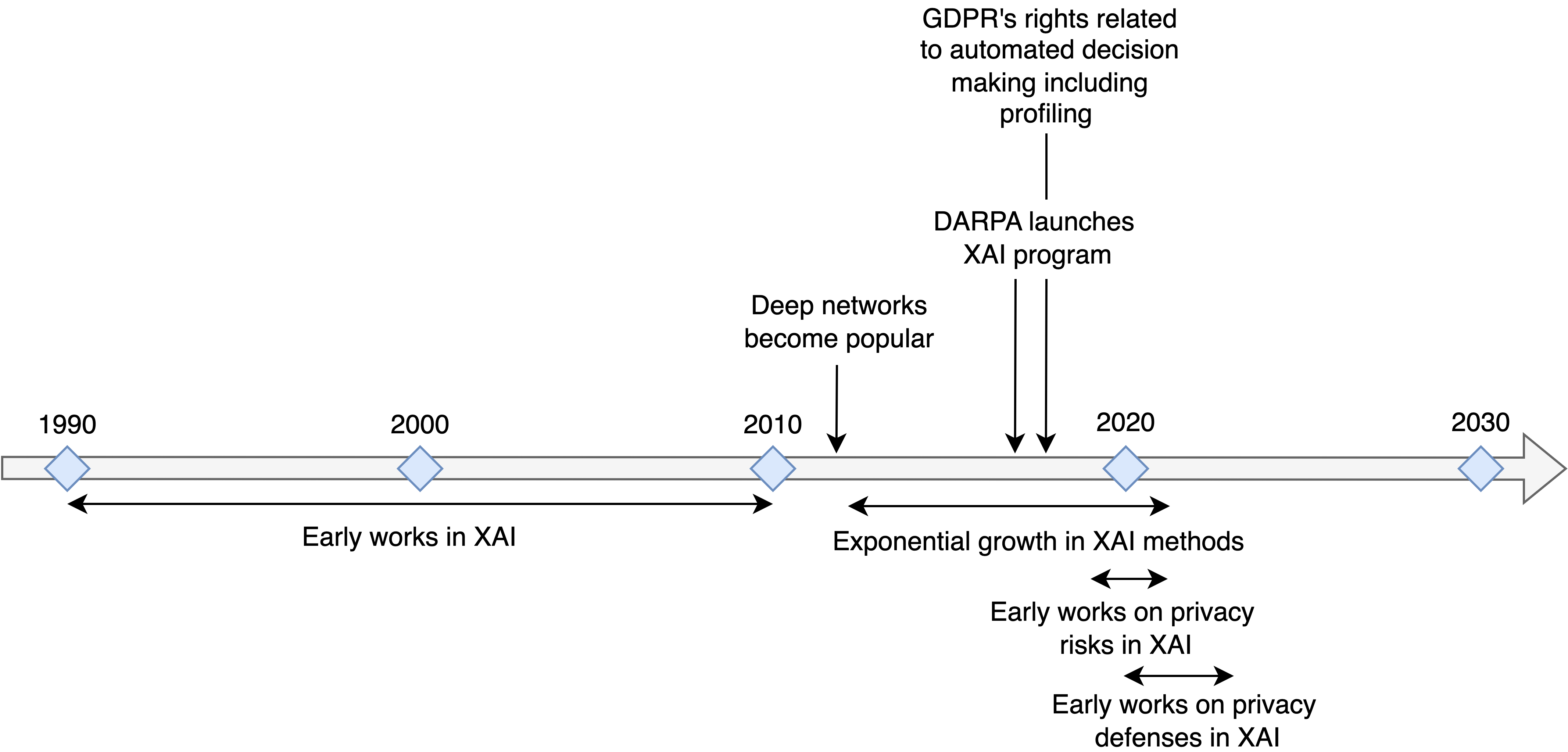}
\end{center}
\caption{Key milestones and emergence of privacy
attacks/defenses in XAI.}
\label{fig:fig1}
\end{figure}

The field of explainability can be traced to the early 1990s, driven by
the lack of transparency in black-box models. Early contributions \citep{benitezAreArtificialNeural1997, cravenExtractingTreeStructuredRepresentations1995, liminfuRuleGenerationNeural1994, milareApproachExplainNeural2002, torresExtractingTreesTrained2005} proposed different techniques for
extracting interpretable representations from these systems. The rise of deep learning and the
improvement in the predictive performance of black-box systems, propelled complex
uninterpretable systems into mainstream usage. However, their use in
critical domains remains problematic due to their lack of transparency.
Regulatory frameworks such as the General Data Protection Regulation \citep{gdprArt22GDPR2016}, specifically the provisions on individuals' rights related to automated
decision-making including profiling, intensified the demand for transparent, explainable models thus resulting in a rapid growth in the field of explainability. \\
However, the introduction of transparency through XAI methods has also exposed new vectors for privacy leakage through explanation
interfaces. Early studies \citep{milliModelReconstructionModel2019,shokriExploitingTransparencyMeasures2020,shokriPrivacyRisksModel2021}  described privacy attacks on the training data and the
underlying model. In response, researchers have begun to explore
defense mechanisms and pioneering works in this field \citep{harderInterpretableDifferentiallyPrivate2020,patelModelExplanationsDifferential2022} have proposed various strategies for generating privacy
preserved explanations. Despite these efforts, privacy risks in XAI remain an open research problem, with novel attacks being identified and defense strategies being actively investigated. 
Figure~\ref{fig:fig1} outlines the key milestones in the evolution of XAI and
highlights the emergence of privacy issues and proposed
defenses.

\subsection{Categorization of XAI}\label{section2.3}

In recent years, several different XAI methods have been proposed.
Broadly, explainability can be achieved using inherently interpretable
models or applying post-hoc methods on trained models \citep{harderInterpretableDifferentiallyPrivate2020}. Methods specific to certain model types and capabilities, are
referred to as model-specific while those independent of the model are
referred to as model-agnostic \citep{dwivediExplainableAIXAI2023}. In this
subsection, we discuss the main categories into which XAI methods are
grouped in existing literature (Table~\ref{tab:table1}), based on the underlying
mechanism used to derive explanations. We also include categories suitable for Gen-AI (Sections ~\ref{section2.3.5} to ~\ref{section2.3.8}) that explore new explanation paradigms for large models \citep{sarkar2024explaining}. Since there is a broad spectrum
of available explainability methods, we limit ourselves to a selection
of methods to give readers sufficient understanding of the terminologies
used in subsequent sections. For a comprehensive review of XAI
categories and methods, we refer the reader to other related reviews
listed in Section~\ref{section2.4}.

\subsubsection{Interpretable methods}\label{section2.3.1}
These AI models are understandable by design \citep{arrietaExplainableArtificialIntelligence2020}.
They have embedded rules or transparent architecture that facilitates
the understanding of the input-output logic of the system. They are also
referred to as white-box or transparent models. Decision trees, Bayesian
networks, linear/logistic regression, k-nearest neighbours, rule based
systems and general additive models \citep{arrietaExplainableArtificialIntelligence2020,molnarInterpretableMachineLearning2023,rawalRecentAdvancesTrustworthy2022} are some examples of interpretable models. Though these models are promising in aiding the
understandability of a system, they have limitations. The primary deterrent to
their successful adoption as explainable-by-design methods, is their
lower accuracy \citep{blanco-justiciaMachineLearningExplainability2020,elzeinPrivaTreeCollaborativePrivacyPreserving2024,gunningDARPAExplainableArtificial2019} compared to better performing black-box models
such as deep learning systems. They also lack
natural language explanations, making them unsuitable for use by
non-technical users \citep{biranHumanCentricJustificationMachine2017}. 
Nonetheless, due to their intrinsically transparent architecture,
interpretable models are often used as surrogates for black-box models
\citep{mcdermidArtificialIntelligenceExplainability2021}. The use of multiple surrogate models
is found to facilitate the availability of different types of explanations \citep{dwivediExplainableAIXAI2023} improving the overall interpretability of the system. 

\subsubsection{Example-based methods}\label{section2.3.2}
These methods use examples, i.e., data instances as samples to explain
the model \citep{mcdermidArtificialIntelligenceExplainability2021}. The instances may be from the
training set or generated by the method \citep{jimenez-lunaDrugDiscoveryExplainable2020,liSurveyDatadrivenKnowledgeaware2020}. These methods are also referred to as
record-based \citep{shokriExploitingTransparencyMeasures2020}, instance-based \citep{jimenez-lunaDrugDiscoveryExplainable2020} or case-based \citep{montenegroPrivacyPreservingGenerativeAdversarial2021} methods in literature.
They can complement feature-based methods to aid understandability of
the end users \citep{jiaPredictionWeaningMechanical2021} and also improve the
interpretability of complex distributions \citep{kimExamplesAreNot2016}. They are
intuitive and natural in their ability to provide explanations to humans
\citep{jimenez-lunaDrugDiscoveryExplainable2020}. Some methods in this category are anchors \citep{ribeiroAnchorsHighPrecisionModelAgnostic2018}, contrastive explanations \citep{dhurandharExplanationsBasedMissing2018}, counterfactuals \citep{wachterCounterfactualExplanationsOpening2017}, influence functions \citep{kohUnderstandingBlackboxPredictions2017} and prototypes and criticisms \citep{kimExamplesAreNot2016}.

\subsubsection{Knowledge-based methods}\label{section2.3.3}
These methods utilize knowledge representation techniques in machine
learning (ML) models to enhance interpretability \citep{tiddiKnowledgeGraphsTools2022}. The integration of
background knowledge \citep{hitzlerNeurosymbolicApproachesArtificial2022} facilitates the incorporation of
contextual information \citep{lecueRoleKnowledgeGraphs2020,paezPragmaticTurnExplainable2019}, thus increasing the trustworthiness
of explanations. The emerging field of neuro-symbolic \citep{hitzlerNeurosymbolicApproachesArtificial2022} or in-between
methods \citep{ilkouSymbolicVsSubsymbolic2020} explores the 
integration of symbolic AI approaches rooted in knowledge representation and
reasoning with subsymbolic or connectionist based approaches \citep{hitzlerNeuralsymbolicIntegrationSemantic2020}.  \\
Another knowledge-based approach is the use of semantic web technologies for semantic
interpretation and automated reasoning from structured knowledge bases \citep{seeligerSemanticWebTechnologies2019}. Knowledge graphs and ontologies are the common
tools that can be deployed to support explainability. Knowledge graphs have
applicability in pre-model and post-model explainabilty contexts for feature extraction, relation identification, inferencing and reasoning \citep{rajabiKnowledgegraphbasedExplainableAI2022}. The field of semantic web technologies in
explainability is attractive because of its potential in creating
knowledge-rich explanations without compromising the model performance \citep{seeligerSemanticWebTechnologies2019}.

\subsubsection{Feature-based methods}\label{section2.3.4}
These explanation methods score or measure the effect of individual
input features on the output of the model \citep{arrietaExplainableArtificialIntelligence2020,bhattEvaluatingAggregatingFeaturebased2020,dwivediExplainableAIXAI2023,strobelDataPrivacyTrustworthy2022}. They are
also referred to as feature importance \citep{mcdermidArtificialIntelligenceExplainability2021}, feature
relevance \citep{arrietaExplainableArtificialIntelligence2020} or attribution-based \citep{liuPleaseTellMe2024} methods. They are based on the attribution problem which is the
distribution of the output of a model for a specific input to its base
features \citep{sundararajanManyShapleyValues2020}. Two important categories of
feature-based methods identified in literature are perturbation and
backpropagation-based methods \citep{anconaBetterUnderstandingGradientbased2018,mcdermidArtificialIntelligenceExplainability2021}.
\begin{itemize}
\item
  \emph{Perturbation-based methods} remove, alter, or mask an input
  feature or set of features and observe the difference with the
  original output \citep{mcdermidArtificialIntelligenceExplainability2021}. Some perturbation-based
  methods are LIME \citep{ribeiroWhyShouldTrust2016}, permutation feature
  importance \citep{breimanRandomForests2001}, SHAP \citep{lundbergUnifiedApproachInterpreting2017} and MASK \citep{fongInterpretableExplanationsBlack2017}.
\item
  \emph{Backpropagation-based methods} compute input attributions in
  forward and backward passes of the network \citep{anconaBetterUnderstandingGradientbased2018}. The use of the gradient of the output with the respective input features
  \citep{mcdermidArtificialIntelligenceExplainability2021,strobelDataPrivacyTrustworthy2022} is a common approach
  in these methods and is referred to as gradient-based approach.
  Methods used on images that determine the global importance of pixels,
  generate saliency maps, and are referred to as pixel-level attribution
  methods \citep{kapishnikovXRAIBetterAttributions2019,molnarInterpretableMachineLearning2023}. Some examples of
  backpropagation-based methods are gradient \citep{simonyanDeepConvolutionalNetworks2014},
  gradient x input \citep{shrikumarNotJustBlack2017},
  guided backpropagation \citep{springenbergStrivingSimplicityAll2015} and integrated
  gradients \citep{sundararajanAxiomaticAttributionDeep2017}.
\end{itemize}

\subsubsection{Concept-based methods}\label{section2.3.5}
Concept-based methods aim to uncover high-level concepts above features, pixels and characters \citep{ghorbani2019towards}, that influence the behaviour of a model \citep{bereska2024mechanistic}. The use of human understandable concepts result in naturally interpretable explanations. Desiderata such as meaningfulness, coherence and importance are proposed for this category of explanations \citep{ghorbani2019towards}. 
Concept activation vectors (CAVs) \citep{kim2018interpretability} is a concept-based method that leverages linear classifiers to determine the presence or absence of concepts corresponding to a set of input examples. The use of concept activation regions \citep{crabbe2022concept} enhances CAVs and allows the underlying examples to be distributed across the latent space of the model, resulting in the  discovery of global explanations. In another approach, concept bottleneck models \citep{koh2020concept} predict intermediate human-specified concepts in the model and utilise these for making predictions. The models also support human interventions on these concepts and facilitate the generation of counterfactuals, thus enabling human-model collaboration \citep{koh2020concept}.

\subsubsection{Probing-based methods}\label{section2.3.6}
 Probing-based methods use classifiers or probes to determine the knowledge captured by a model \citep{sajjad2022neuron, schneider2024explainable}. The technique involves using the internal activations of networks to train probing classifiers for prediction of embedded properties \citep{alain2017probing, belinkov2022probes}. However, such probing experiments require the properties to be known apriori along with clear specifications of the original/probing tasks, the models and datasets used. The outcomes of these experiments can be applied for tuning the model on certain tasks or determining the encoded information for downstream tasks. 

\subsubsection{Neuron activation methods}\label{section2.3.7}
These methods focus on the behaviour of neurons that are responsible for specific outcomes or represent learned linguistic properties \citep{zhaoExplainabilityLargeLanguage2024}. Proposed methods in this category, such as linguistic correlation analysis and cross-model correlation analysis, identify and analyse such neurons for post-hoc explainability \citep{dalvi2019neuronact}. Individual or groups of neurons may be discovered as salient for a property and ranked based their importance to the model's task \citep{dalvi2019neuronact}. These methods may also experience trade-offs between accuracy and selectivity \citep{antverg2022neuronact}. 

\subsubsection{Mechanistic interpretability}\label{section2.3.8}
Mechanistic interpretability \citep{bereska2024mechanistic} seeks to understand the model outputs by reverse engineering \citep{olah2022mechanistic} the system. Analogous to understanding complex computer programs, this approach aims at dissecting network activations into independently understandable units. It considers the analysis of individual components of the system, such as features, connections and neurons, to determine the causes of outputs. This approach to interpretability includes techniques that could be applied before, during or after the training process. Techniques, such as feature visualisation \citep{zimmermann2021mechint} and sparse autoencoders \citep{cunningham2024sae}, are some of the methods from this category. 

\textbf{Table 1} Broad XAI categories and a selection of early works.
\begin{longtable}[]{@{}
  >{\raggedright\arraybackslash}p{(\linewidth - 8\tabcolsep) * \real{0.1472}}
  >{\raggedright\arraybackslash}p{(\linewidth - 8\tabcolsep) * \real{0.2138}}
  >{\raggedright\arraybackslash}p{(\linewidth - 8\tabcolsep) * \real{0.2296}}
  >{\raggedright\arraybackslash}p{(\linewidth - 8\tabcolsep) * \real{0.1641}}
  >{\raggedright\arraybackslash}p{(\linewidth - 8\tabcolsep) * \real{0.2453}}@{}}
\toprule\noalign{}
\label{tab:table1}
\begin{minipage}[b]{\linewidth}\centering
\textbf{XAI Category}
\end{minipage} &
\multicolumn{2}{>{\centering\arraybackslash}p{(\linewidth - 8\tabcolsep) * \real{0.4434} + 2\tabcolsep}}{%
\begin{minipage}[b]{\linewidth}\centering
\textbf{XAI Method}
\end{minipage}} & \begin{minipage}[b]{\linewidth}\centering
\textbf{Model-specific/agnostic}
\end{minipage} & \begin{minipage}[b]{\linewidth}\centering
\textbf{Study}
\end{minipage} \\
\midrule\noalign{}
\endhead
\bottomrule\noalign{}
\endlastfoot
Interpretable &
\multicolumn{2}{>{\raggedright\arraybackslash}p{(\linewidth - 8\tabcolsep) * \real{0.4434} + 2\tabcolsep}}{%
Decision trees, Bayesian networks, linear/logistic regression, k-nearest
neighbours, rule-based systems, general additive models} &
Model-specific & - \\
\cline{1-5}
\multirow[t]{5}{=}{Example-based} &
\multicolumn{2}{>{\raggedright\arraybackslash}p{(\linewidth - 8\tabcolsep) * \real{0.4434} + 2\tabcolsep}}{%
Anchors} & Model-agnostic & \citet{ribeiroAnchorsHighPrecisionModelAgnostic2018} \\
\cline{2-5}
&
\multicolumn{2}{>{\raggedright\arraybackslash}p{(\linewidth - 8\tabcolsep) * \real{0.4434} + 2\tabcolsep}}{%
Contrastive explanations} & Model-agnostic & \citet{dhurandharExplanationsBasedMissing2018} \\
\cline{2-5}
&
\multicolumn{2}{>{\raggedright\arraybackslash}p{(\linewidth - 8\tabcolsep) * \real{0.4434} + 2\tabcolsep}}{%
Counterfactuals} & Model-agnostic & \citet{wachterCounterfactualExplanationsOpening2017} \\
\cline{2-5}
&
\multicolumn{2}{>{\raggedright\arraybackslash}p{(\linewidth - 8\tabcolsep) * \real{0.4434} + 2\tabcolsep}}{%
Influence functions} & Model-agnostic & \citet{kohUnderstandingBlackboxPredictions2017} \\
\cline{2-5}
&
\multicolumn{2}{>{\raggedright\arraybackslash}p{(\linewidth - 8\tabcolsep) * \real{0.4434} + 2\tabcolsep}}{%
Prototypes and criticisms} & Model-agnostic & \citet{kimExamplesAreNot2016} \\
\cline{1-5}
\multirow[t]{2}{=}{Knowledge-based} &
\multicolumn{2}{>{\raggedright\arraybackslash}p{(\linewidth - 8\tabcolsep) * \real{0.4434} + 2\tabcolsep}}{%
Semantic web technologies} & Model-agnostic & \citet{seeligerSemanticWebTechnologies2019} \\
\cline{2-5}
&
\multicolumn{2}{>{\raggedright\arraybackslash}p{(\linewidth - 8\tabcolsep) * \real{0.4434} + 2\tabcolsep}}{%
Neuro-symbolic approaches} & Model-specific & \citet{hitzlerNeurosymbolicApproachesArtificial2022} \\
\cline{1-5}
\multirow[t]{8}{=}{Feature-based} & \multirow[t]{4}{=}{Perturbation-based} &
LIME & Model-agnostic & \citet{ribeiroWhyShouldTrust2016} \\
\cline{3-5}
& & Permutation Feature Importance & Model-agnostic & \citet{breimanRandomForests2001} \\
\cline{3-5}
& & SHAP & Model-agnostic & \citet{lundbergUnifiedApproachInterpreting2017} \\
\cline{3-5}
& & MASK & Model-agnostic & \citet{fongInterpretableExplanationsBlack2017} \\
\cline{2-5}
& \multirow[t]{4}{=}{Backpropagation-based} & Gradient & Model-specific &
\citet{simonyanDeepConvolutionalNetworks2014} \\
\cline{3-5}
& & Gradient x Input & Model-specific & \citet{shrikumarNotJustBlack2017} \\
\cline{3-5}
& & Guided Backpropagation & Model-specific & \citet{springenbergStrivingSimplicityAll2015} \\
\cline{3-5}
& & Integrated Gradients & Model-specific & \citet{sundararajanAxiomaticAttributionDeep2017} \\
\cline{1-5}
Concept-based &
\multicolumn{2}{>{\raggedright\arraybackslash}p{(\linewidth - 8\tabcolsep) * \real{0.4434} + 2\tabcolsep}}{%
Concept activation vectors} &
Model-specific & \citet{kim2018interpretability} \\
\cline{2-5}
&
\multicolumn{2}{>{\raggedright\arraybackslash}p{(\linewidth - 8\tabcolsep) * \real{0.4434} + 2\tabcolsep}}{%
Concept bottleneck models} & Model-specific & \citet{koh2020concept} \\
\cline{2-5}
&
\multicolumn{2}{>{\raggedright\arraybackslash}p{(\linewidth - 8\tabcolsep) * \real{0.4434} + 2\tabcolsep}}{%
Concept activation regions} & Model-specific & \citet{crabbe2022concept} \\
\cline{1-5}
Probing-based &
\multicolumn{2}{>{\raggedright\arraybackslash}p{(\linewidth - 8\tabcolsep) * \real{0.4434} + 2\tabcolsep}}{%
Probing experiments} &
Model-specific & \citet{alain2017probing} \\
\cline{1-5}
Neuron activation &
\multicolumn{2}{>{\raggedright\arraybackslash}p{(\linewidth - 8\tabcolsep) * \real{0.4434} + 2\tabcolsep}}{%
Linguistic correlation analysis} &
Model-specific & \citet{dalvi2019neuronact} \\
\cline{2-5}
&
\multicolumn{2}{>{\raggedright\arraybackslash}p{(\linewidth - 8\tabcolsep) * \real{0.4434} + 2\tabcolsep}}{%
Cross-model correlation analysis} & Model-specific & \citet{dalvi2019neuronact} \\
\cline{1-5}
Mechanistic interpretability &
\multicolumn{2}{>{\raggedright\arraybackslash}p{(\linewidth - 8\tabcolsep) * \real{0.4434} + 2\tabcolsep}}{%
Feature visualisation} &
Model-specific & \citet{zimmermann2021mechint} \\
\cline{2-5}
&
\multicolumn{2}{>{\raggedright\arraybackslash}p{(\linewidth - 8\tabcolsep) * \real{0.4434} + 2\tabcolsep}}{%
Sparse autoencoders} & Model-specific & \citet{cunningham2024sae} 
\end{longtable}

\subsection{Related reviews}\label{section2.4}
XAI is currently an active research area and detailed reviews have captured the state of the art in the field. Though current literature has reviews covering different aspects of XAI, to the best of our knowledge there is a lack of comprehensive review that considers the tension of privacy with explainability. Our work addresses this gap and offers a unique contribution compared to other existing reviews. In this subsection, we identify related reviews on XAI and summarize their focus areas.

An in-depth overview of the core concepts and taxonomies in XAI was provided by \citet{arrietaExplainableArtificialIntelligence2020}. \citet{mohseniMultidisciplinarySurveyFramework2021} conducted an interdisciplinary survey and proposed a comprehensive framework for design and
evaluation of XAI methods. \citet{dwivediExplainableAIXAI2023} covered a wide breadth of explanation algorithms, programming frameworks and software toolkits for XAI development. \citet{aliExplainableArtificialIntelligence2023} examined explainability through the
lens of trustworthiness detailing evaluation metrics, available software packages and XAI datasets. \citet{bodriaBenchmarkingSurveyExplanation2023} systematically categorized explanation methods and benchmarked prominent methods using quantitative metrics. \citet{muralidharElementsThatInfluence2023} reviewed transparency elements from human computer interaction (HCI) in the context of explanations while \citet{cambriaSurveyXAINatural2023} investigated presentation methods and usage of natural language with XAI. 

Beyond these broad surveys, domain specific reviews have also emerged. For example, XAI in healthcare has been surveyed by \citet{payrovnaziriExplainableArtificialIntelligence2020} and \citet{yangUnboxBlackboxMedical2022}; in cybersecurity by \citet{capuanoExplainableArtificialIntelligence2022} and in energy and
power systems by \citet{machlevExplainableArtificialIntelligence2022}. Methodology focussed reviews also exist, covering counterfactuals \citep{guidottiCounterfactualExplanationsHow2022},
data-driven knowledge-aware XAI systems \citep{liSurveyDatadrivenKnowledgeaware2020},
knowledge-graph based XAI \citep{rajabiKnowledgegraphbasedExplainableAI2022, tiddiKnowledgeGraphsTools2022} and semantic web technologies for explanations \citep{seeligerSemanticWebTechnologies2019}. Recent advances include the intersection of explainability and federated learning (FL), termed as Federated XAI (Fed-XAI), reviewed by \citet{lopez-blancoFederatedLearningExplainable2023} and categorisation of explanation techniques for transformer-based language models based on training paradigms as surveyed by \citet{zhaoExplainabilityLargeLanguage2024}. A comprehensive review of mechanistic interpretabililty in LLMs for achieving AI safety was conducted by \citet{bereska2024mechanistic}.

Recent literature has increasingly highlighted the potential of malicious exploitation of XAI interfaces. \citet{baniecki2024adversarial} present a survey of adversarial attacks in XAI and discuss corresponding defenses. While their study addresses a specific type of XAI privacy attack, its primary emphasis is on non-privacy attacks including data poisoning, backdoor attacks, model manipulation and attacks on fairness metrics through explanations.  A categorisation of XAI-aware attacks was proposed in another recent work \citep{Noppel2024adversarial, Noppel2024SoK} based on the outputs that are preserved in the attack. The work also categorised attacks based on the scope of explanation alteration and adversary capabilities. However, this work targets the robustness of post-hoc XAI and though the proposed categorisation is relevant to privacy attacks, it does not directly focus on the privacy aspect of XAI safety.

The focus of this review diverges from prior reviews by specifically examining the privacy risks that target the identification and exposure of personal or sensitive information through the misuse of XAI interfaces in AI systems. Further, we review strategies used by researchers in
mitigating the privacy leakage in XAI. An early work by \citet{spartalis2023balancing}, provides a short review on XAI privacy risks and applicable defense mechanisms. The review also identifies security risks such as evasion and poisoning through XAI interfaces. Though this work raises awareness for further research in this area, it mainly focuses on early privacy attacks and discusses limited defense mechanisms. A recent review \citep{nguyen2025privacy} explores the intersection of privacy and explainability through attacks and countermeasures, but differs from our review in its methodology and lacks a detailed examination of the core requirements underpinning the design of privacy preserving XAI. Our review employs an established scoping
review methodology guided by clearly defined research questions. The resulting taxonomy
of XAI privacy risks and corresponding mitigation methods are distilled
from the understanding of existing literature across the privacy and XAI communities. This methodology enables a structured and rigorous approach to addressing the research questions through the analysis of the selected studies. 

\section{Method}\label{section3}

We conducted a scoping review based on the Preferred Reporting Items for
Systematic reviews and Meta-Analyses extension for Scoping Reviews
(PRISMA-ScR) \citep{triccoPRISMAExtensionScoping2018}. This section elaborates the
process followed, the proposed taxonomy of XAI privacy risks and the identified research trends.

\subsection{Literature selection and extraction }\label{section3.1}

\begin{figure}[h]
\begin{center}
\includegraphics[width=0.5\linewidth]{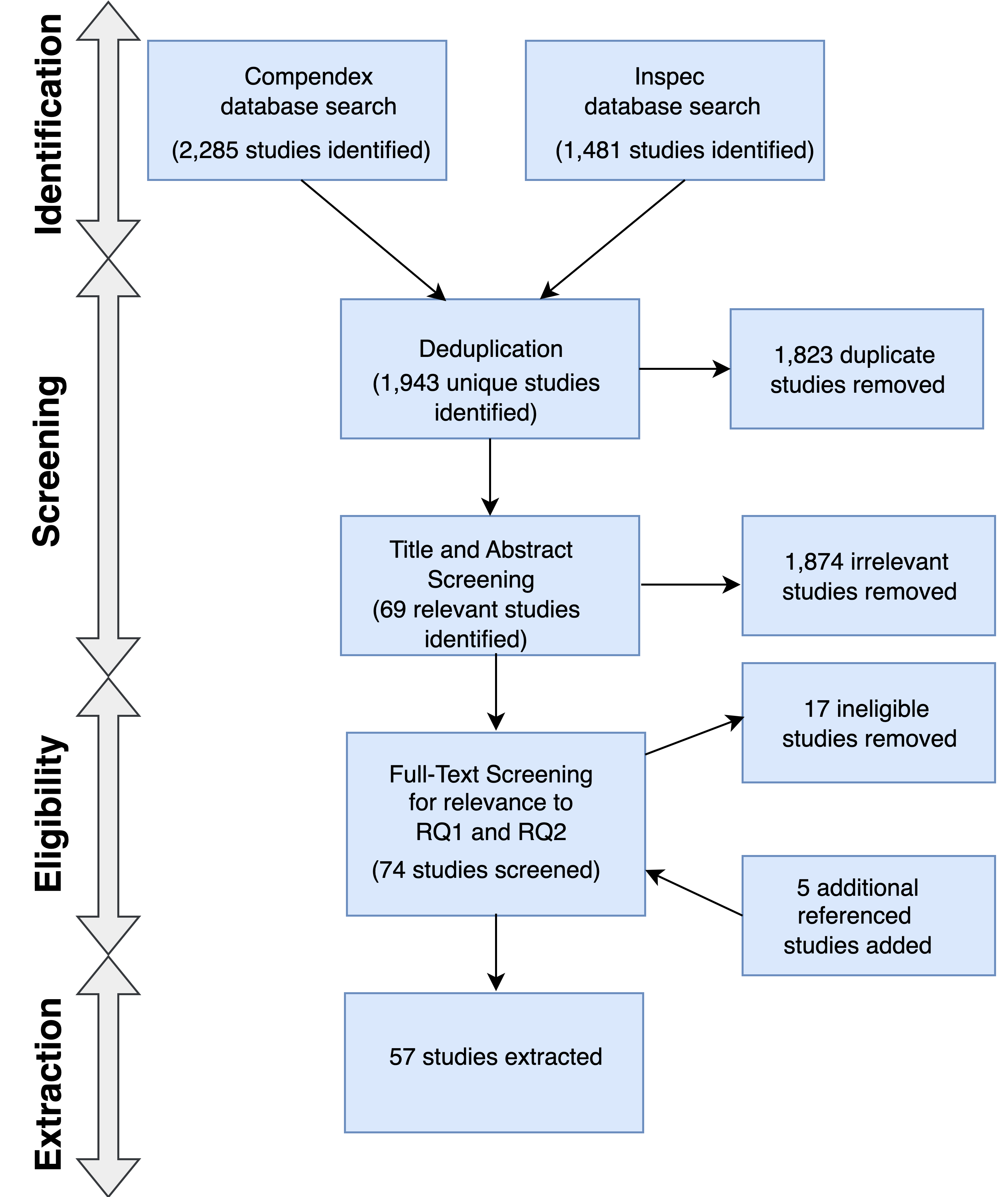}
\end{center}
\caption{Scoping review process as per PRISMA-ScR.}
\label{fig:fig2}
\end{figure}

A 4-step process was employed comprising of identification, screening,
eligibility, and extraction, as illustrated in Figure~\ref{fig:fig2}. In the initial step of identification, Elsevier Engineering Village \citep{elsevierb.v.EngineeringVillage} search platform was used and the search was conducted on Compendex
and Inspec databases. These databases index publications from leading computer
science publishers, including IEEE, ACM, Springer and Elsevier. A researcher formulated the search string, using the two main concepts of privacy and
explainability, with the help of a librarian. This was applied on the title, subject, and abstract fields. The reseacher worked with the librarian to set the search, inclusion and exclusion criteria as detailed below for reproducibility: 

\begin{itemize}
\item
  \emph{Search string:} (privacy OR confidential* OR ``membership
  inference'' OR ``model inversion'' OR ``model extract*'' OR ``model
  reconstruct*'' OR ``property inference'') AND (explainab* OR explanat*
  OR interpretab* OR XAI OR recourse OR ``transparency report'').
\item
  \emph{Period of publication:} January 1, 2019, to December 31, 2024.
  The start year was chosen based on the seminal works \citep{milliModelReconstructionModel2019,shokriExploitingTransparencyMeasures2020,shokriPrivacyRisksModel2021} published on this topic.
\item
  \emph{Date of most recent search:} Jan 6, 2025
\item
  \emph{Type of publications included:} journal articles, conference
  articles, book chapters, articles in press.
\item
  \emph{Type of publications excluded:} preprints, unpublished papers,
  dissertations, books, standards, report chapters, notes, report
  reviews, editorials, erratum, retracted documents. This criteria was required to ensure the extraction of high quality and rigorously peer reviewed articles. 
\end{itemize}

\begin{itemize}
\item
  \emph{Language:} English
\item
  \emph{Inclusion criteria:} Study should describe at least one privacy
  risk or privacy preservation method in XAI.
\end{itemize}

The search results comprising of 3,766 studies were exported from
Engineering Village and imported into Covidence \citep{Covidence}
review management software by one researcher. During the import process, the software
merged duplicate studies from the databases, retaining
only unique records. After deduplication, 1,943 studies were forwarded
for screening wherein the title and abstract were examined by the researcher to determine
relevance to RQ1 or RQ2 while considering the inclusion criteria. Out of 1,943
studies, 69 studies were moved to the next step to determine eligibility
wherein the full text of the identified articles were examined with
respect to RQ1 and RQ2. This step was conducted by one researcher and independently verified by a second researcher. Since the focus of this article is privacy leakage through XAI interfaces, studies that addressed adversarial XAI (such as backdoor, poisoning, etc.) without overlap on privacy were eliminated. Studies that discussed general privacy issues in ML without XAI context were also eliminated. Related survey papers that appeared in the search results were not extracted but used to determine any additional studies relevant to the research questions. These steps resulted in removal of 17 studies and addition of 5 studies identified through forward and backward searches. The removal and addition of studies were discussed and verified by two researchers. Thus, overall the 4-step scoping review process resulted in extraction of 57 studies.

\subsection{Proposed taxonomy of privacy risks in XAI}\label{taxonomy}

\begin{figure}[h]
\begin{center}
\includegraphics[width=0.6\linewidth]{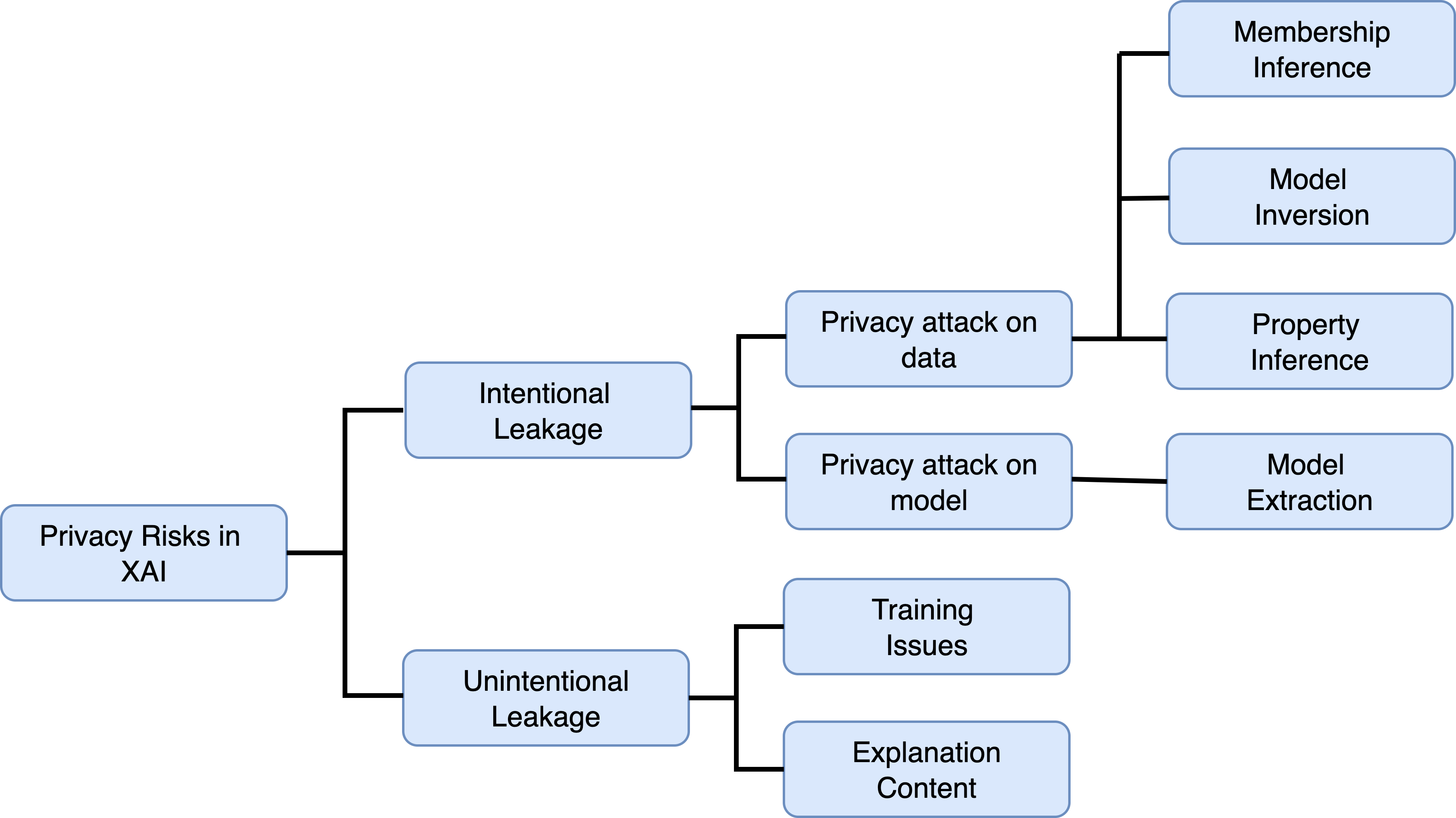}
\end{center}
\caption{Proposed taxonomy of privacy risks in XAI.}
\label{fig:fig3}
\end{figure}

Each extracted study was categorized under the appropriate research
question. The categorisation of studies under RQ1, led to the identification of 3 main types of privacy attacks reported in XAI. These attacks are due to the malicious intent of adversaries and we propose to term this leakage as intentional. This leakage can target data (training or query) or models (Figure~\ref{fig:fig3}). A fourth type of privacy attack in literature was added to the taxonomy due to its applicability to intentional leakage through explanations, even though it is not currently reported in XAI. 

Though all studies extracted under RQ1 belonged to intentional leakage, we note that not all privacy leakage in XAI is due to malicious intent. Hence we propose another category of leakage termed as unintentional. This type of inadvertent leakage can be caused due to training problems, such as overfitting, or design flaws, such as lack of appropriate role-based release of explanations. These problems can be unintentionally introduced in XAI systems and further exploited for intentional leakage by adversaries. In Sections~\ref{section4.2} and ~\ref{section4.3}, we elaborate further on intentional and unintentional leakages respectively.

\subsection{Research trends}\label{research-trends}

\begin{figure}[t]
    \centering
    \subfigure[]{\label{fig:fig4a}
    \includegraphics[width=0.4\textwidth]{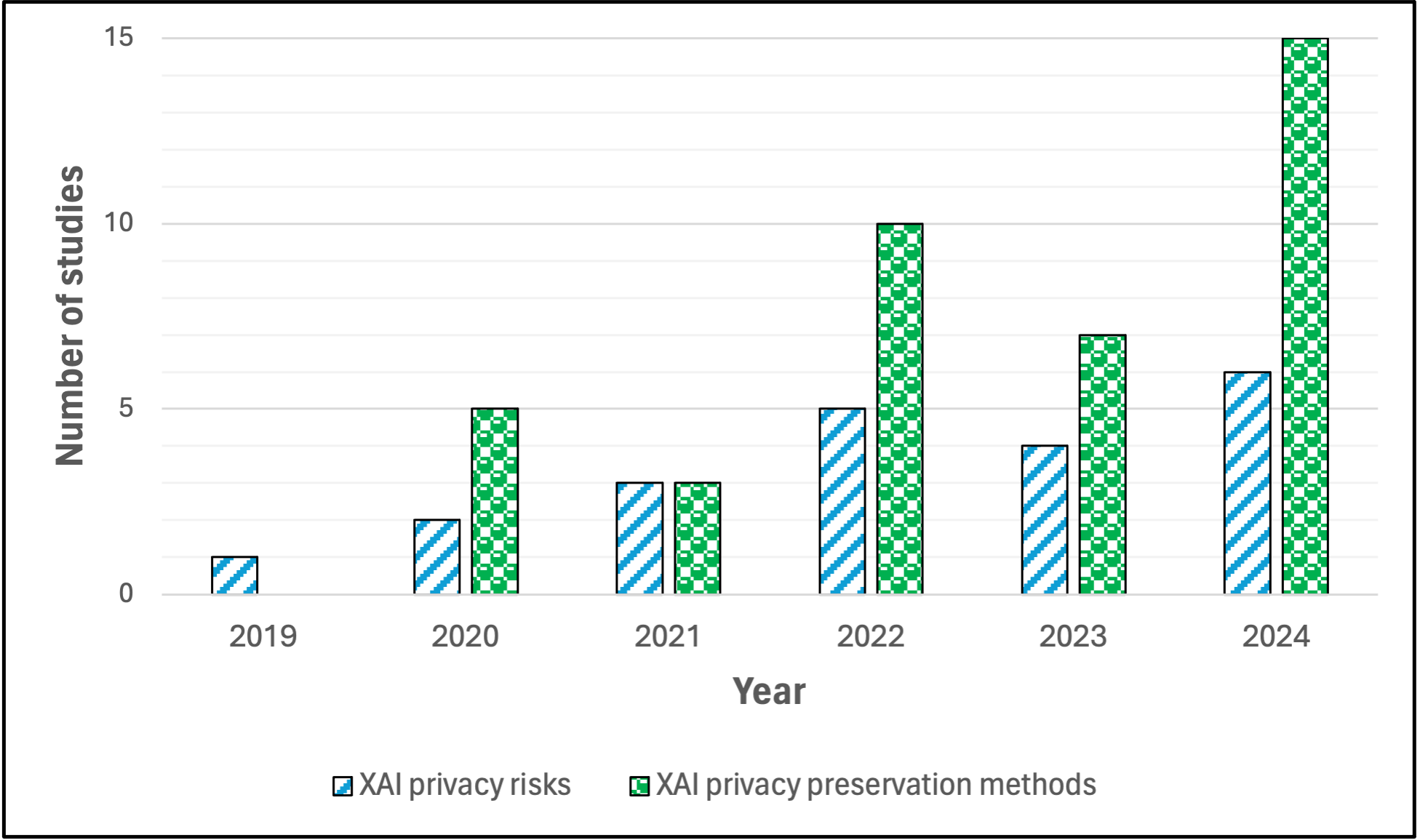}}
    \subfigure[]{\label{fig:fig4b}
    \includegraphics[width=0.4\textwidth]{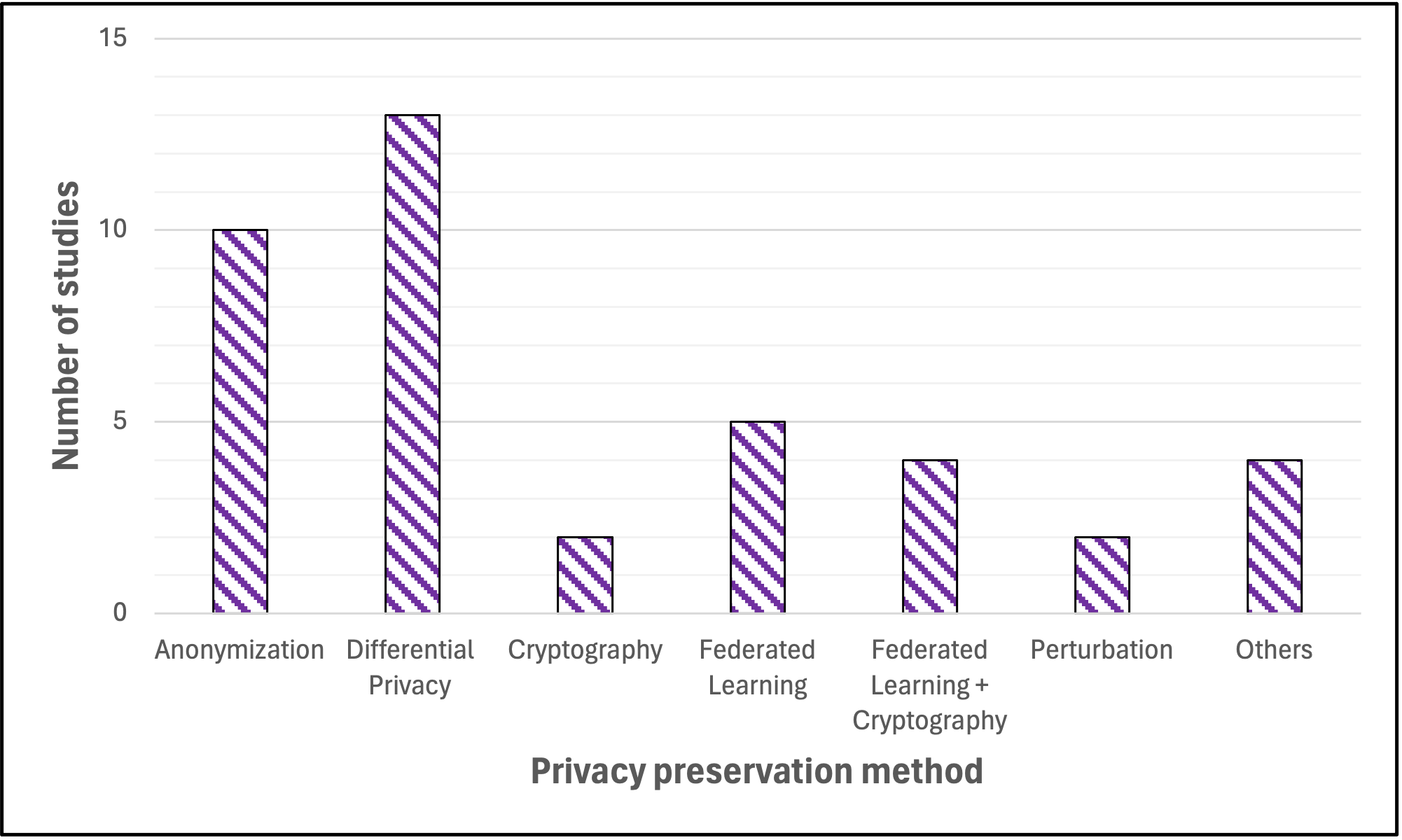}} 
    \subfigure[]{\label{fig:fig4c}
    \includegraphics[width=0.4\textwidth]{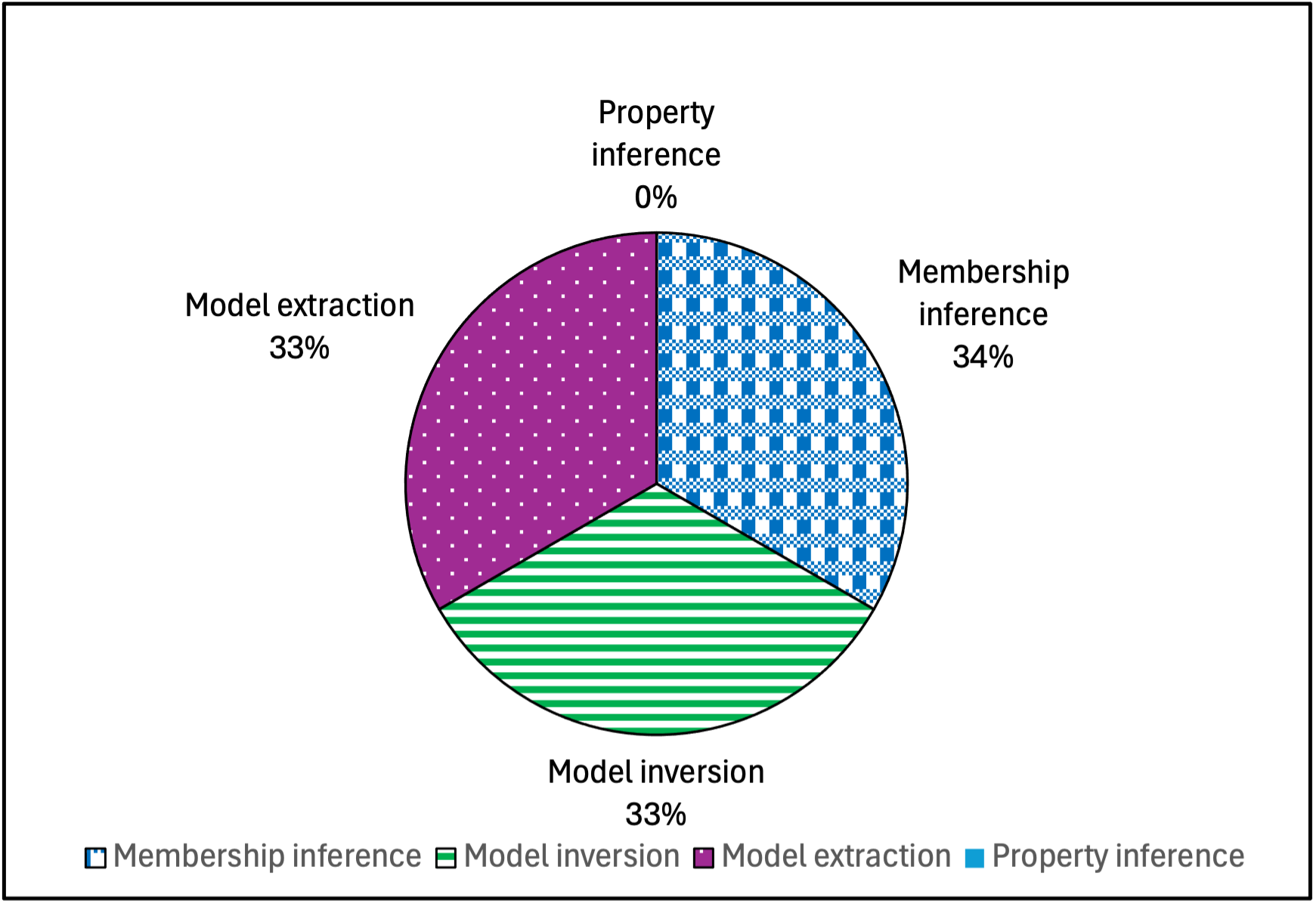}}
    \subfigure[]{\label{fig:fig4d}
    \includegraphics[width=0.4\textwidth]{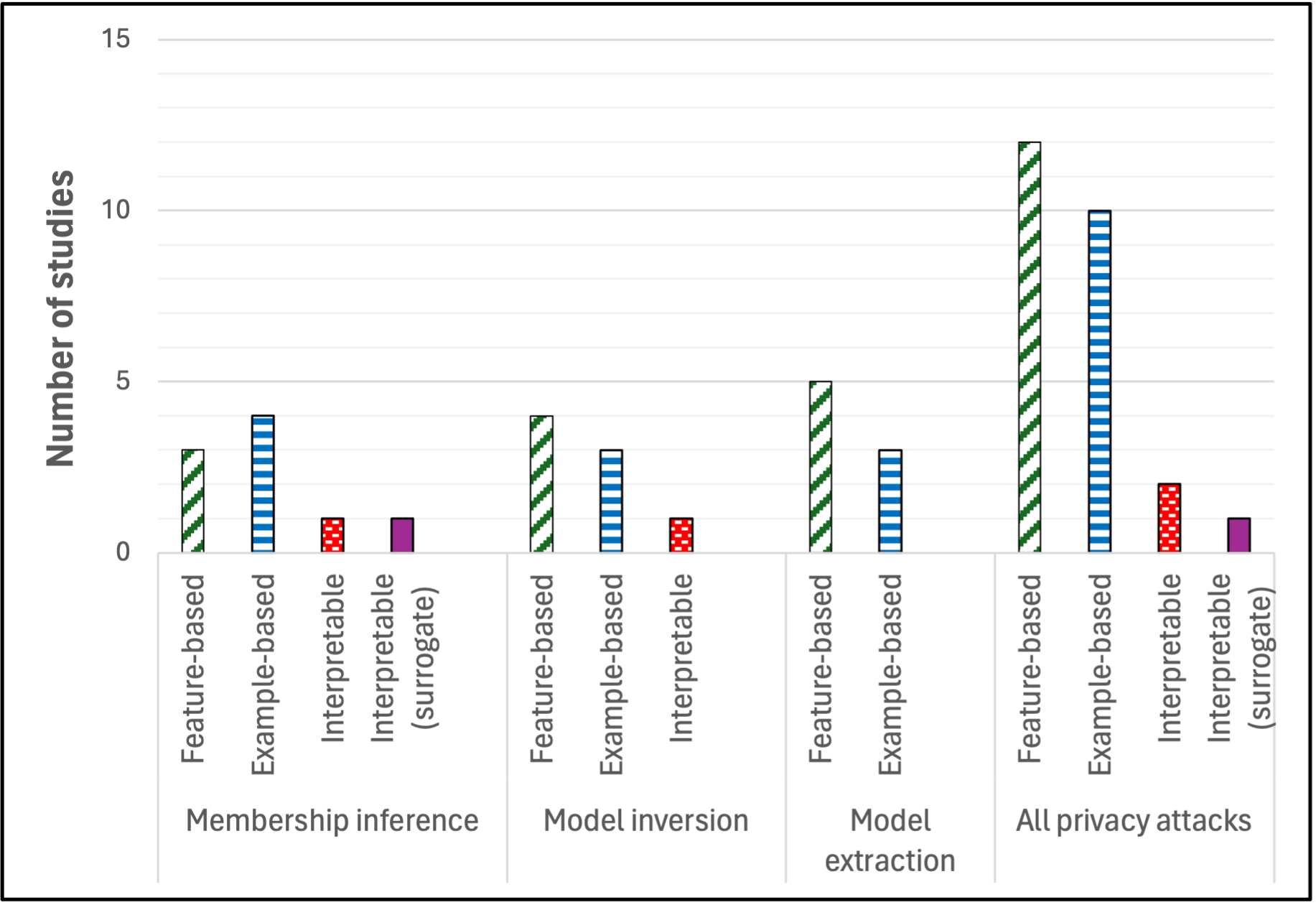}}
    \caption{Research trends identified from extracted studies (a) Studies on XAI privacy risks and preservation methods (b) Privacy preservation methods in XAI (c) Privacy attacks in XAI (d) Privacy attacks by XAI categories.}
    \label{fig:fig4}
\end{figure}

The distribution of the extracted studies for RQ1 (i.e., XAI privacy
risks) and RQ2 (i.e., XAI privacy preservation) by year, can be seen in
Figure~\ref{fig:fig4a}. An upward trend in the reported privacy risks associated with XAI methods is evident over the period under review. Correspondingly, there has been a noticeable increase in the number of studies exploring the use of various privacy
preservation methods in XAI as observed from Figure~\ref{fig:fig4b}. Among these techniques, differential
privacy and anonymization emerge as the most commonly employed approaches. With respect to the identified privacy risks, three types of attacks,
namely, membership inference, model inversion and model extraction, appear with comparable frequency across literature (Figure~\ref{fig:fig4c}). Notably, property inference attacks have not been examined in the context of XAI systems. Figure~\ref{fig:fig4d} presents the categories of XAI targeted by different privacy attacks. Feature-based and example-based XAI are more frequently targeted to
such attacks in comparison to interpretable methods. 

\section{Privacy Risks in XAI}\label{section4}

Traditionally privacy is referred to as the ``right to be left alone'' \citep{warrenRightPrivacy1890} and the ``claim of individuals, groups, or
institutions to determine for themselves when, how, and to what extent
information about them is communicated to others'' \citep{westinPrivacyFreedom1967}. In the
modern context, with availability, collection, and collation of copious
information about individuals through online and offline sources, the
concept of information privacy is more applicable and refers to the
ability of individuals to exert control on their own data \citep{curzonPrivacyArtificialIntelligence2021}. \citet{clarkeInternetPrivacyConcerns1999} has defined information privacy as the
``claims of individuals that data about themselves should generally not
be available to other individuals and organizations, and that, where
data is possessed by another party, the individual must be able to
exercise a substantial degree of control over that data and its use''.
In this article, we refer to this latter definition of privacy.

Trustworthy AI is built on the foundational principle of explainability, which supports the gaining of insights into the decision making processes of black-box AI systems \citep{tabassiAIRiskManagement2023}. However, the
relationship between privacy and explainability has contrasting aspects.
On the one hand, explainability aids privacy in several ways such as in creating privacy
awareness in users \citep{brunotteCanExplanationsSupport2021}, in ascertaining that privacy
of a system is achieved \citep{doshi-velezRigorousScienceInterpretable2017,muftuogluPrivacyPreservingMechanismsExplainability2022}, and in determining correlations with identifiable data for
removal \citep{hohmanGamutDesignProbe2019}. On the other hand, explanations can reveal
sensitive information contained in models and training data \citep{harderInterpretableDifferentiallyPrivate2020,rawalRecentAdvancesTrustworthy2022,zhaoExploitingExplanationsModel2021} thus leading to
privacy risks \citep{kuppaAdversarialXAIMethods2021}. Thus, there are conflicting
outcomes \citep{guerra-manzanaresPrivacyPreservingMachineLearning2023,nguyenXRandDifferentiallyPrivate2023,sandersonImplementingResponsibleAI2023,spartalis2023balancing} of including explainability as a non-functional requirement in AI systems. 

\subsection{Threat model}\label{section4.1}

\begin{figure}[h]
\begin{center}
\includegraphics[width=0.6\linewidth]{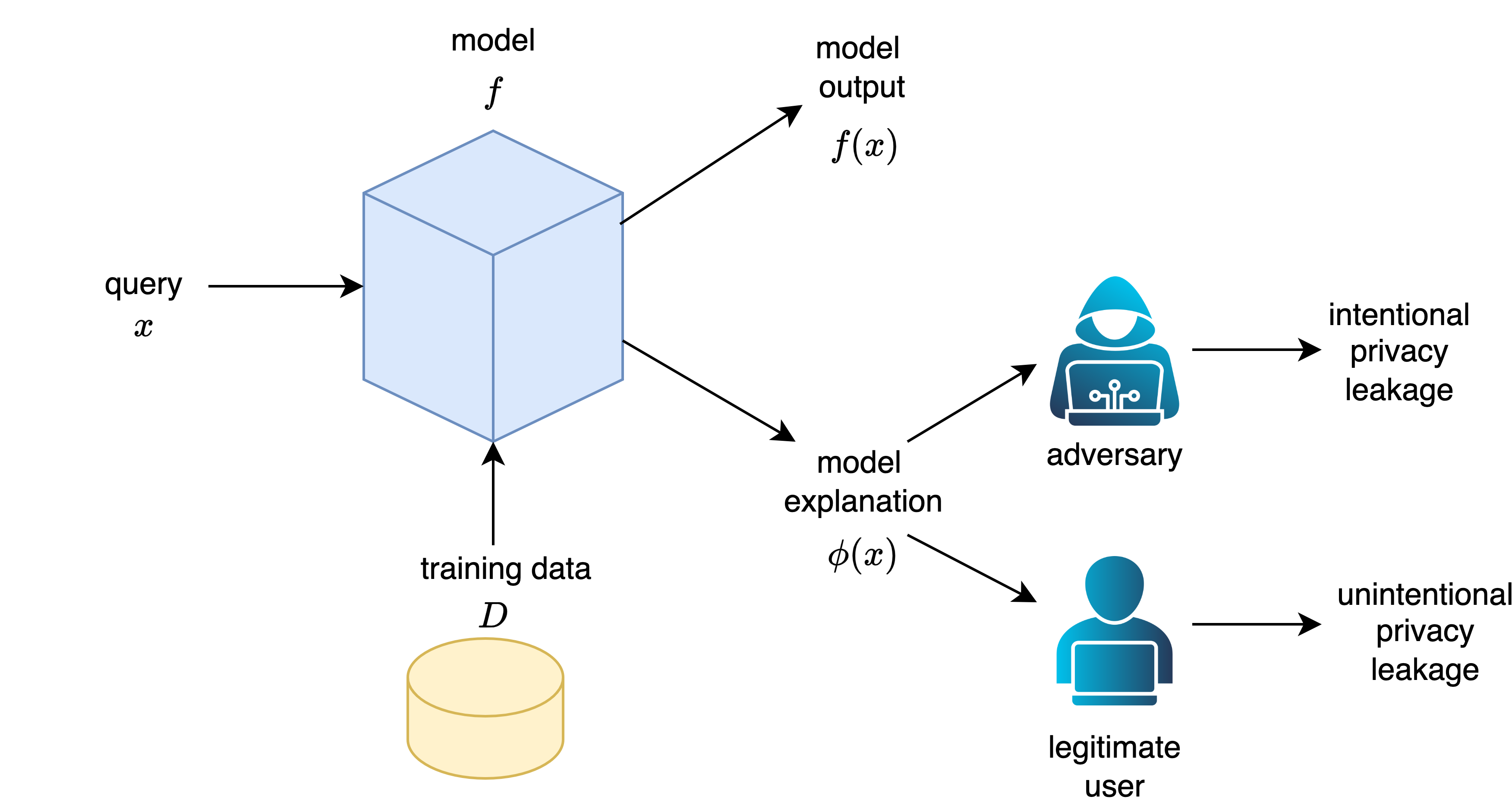}
\end{center}
\caption{Threat model of XAI.}
\label{fig:fig5}
\end{figure}

To discuss the threat model of XAI, we consider a target AI model, \textit{f}, with a corresponding explanation function,  \textit{\( \phi \)}. For an input query, \textit{x}, the system generates an output, \textit{f(x)}, such as prediction, classification or an artifact of a Gen-AI system, and a corresponding explanation, \textit{\( \phi \)(x)} (Figure~\ref{fig:fig5}). Though the explanation interface is made available for the end-users of the system, adversaries can also secure different levels of access to the XAI system during the stages of the AI lifecycle \citep{shahriarSurveyPrivacyRisks2023}.  In white-box access, adversaries possess information about the model internals such as architecture and hyperparameters \citep{liuWhenMachineLearning2022,zhangSurveyPrivacyInference2023}. In black-box access, adversaries
access the input/output of the system \citep{liuWhenMachineLearning2022} without any knowledge of the training process or model internals \citep{rigakiSurveyPrivacyAttacks2023}. Any intermediate access between white-box and black-box is referred to as gray-box \citep{jegorovaSurveyLeakagePrivacy2022}. The adversaries may act as passive observers \citep{jegorovaSurveyLeakagePrivacy2022} and use the model outputs for launching privacy attacks, or they may actively interfere in the training process of the model \citep{nasrComprehensivePrivacyAnalysis2019}.

\subsection{Intentional privacy leakage}\label{section4.2}
This subsection reviews intentional risks in the form of privacy attacks launched by adversaries.  As XAI systems are fundamentally AI models augmented with explainability features, they remain susceptible to malicious threats that affect conventional AI models. Prior
research has identified security attacks, such as evasion
and poisoning \citep{pitropakisTaxonomySurveyAttacks2019}, that compromise the integrity of AI. Adversarial attacks on XAI such as input manipulations \citep{dombrowski2019explanations,zhang2020interpretable}, model manipulations \citep{heo2019fooling} and explanation-aware backdoors \citep{noppel2023backdoor} are also discussed in current literature. However, the present article focusses on privacy attacks that aim to compromise the personal data of individuals or the confidentiality of the underlying model. In the XAI context, model explanations
further aid \citep{zhaoExploitingExplanationsModel2021} the identification or exposure of
personal information of individuals or the intellectual property of the
model owner. Table~\ref{tab:table2} provides an overview of these intentional privacy risks through XAI and the studies addressing them.

\subsubsection{Membership inference}\label{section4.2.1}

\begin{figure}[h]
\begin{center}
\includegraphics[width=0.6\linewidth]{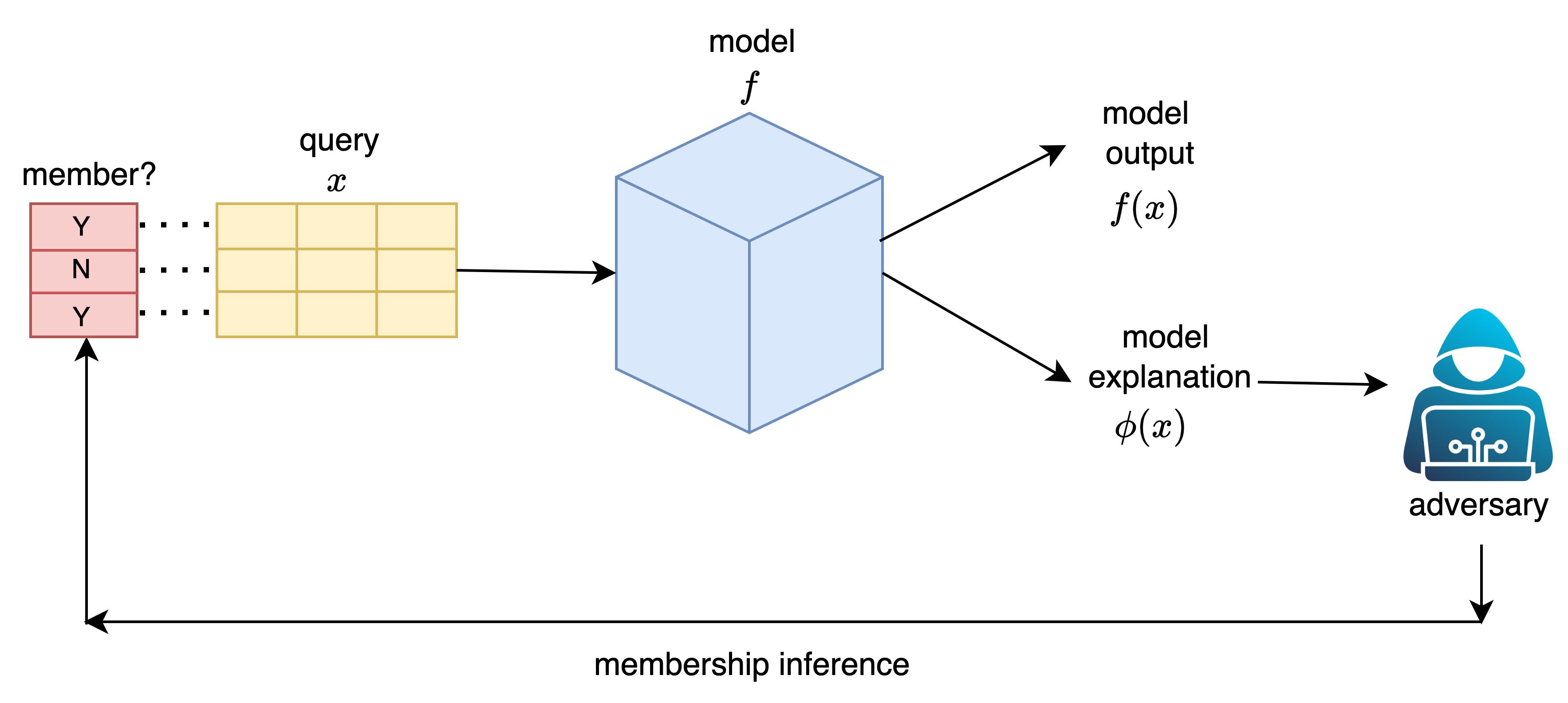}
\end{center}
\caption{Membership inference exploiting explanations.}
\label{fig:fig6}
\end{figure}

This is a privacy risk of identification of an individual in the
training set of a model \citep{shokriMembershipInferenceAttacks2017,zarifzadehLowCostHighPowerMembership2024}
(Figure~\ref{fig:fig6}). An adversary can execute this attack with black-box or
white-box access to the model \citep{vealeAlgorithmsThatRemember2018} after it has been
deployed. A membership inference model can be expressed as the following binary classifier \citep{kuppaAdversarialXAIMethods2021} when the model output, $f(x)$, is available:
\begin{equation}
A_{MemInf}: x,f(x) \rightarrow \{member, non\mbox{-}member\}
\end{equation}
When explanations, $\phi(x)$, are available, they can be alternatively used to differentiate members from non-members using the following inference model: 
\begin{equation}
A_{MemInfExp}: x,\phi(x) \rightarrow \{member, non\mbox{-}member\}
\end{equation}

The seminal work on membership inference in feature-based and
example-based XAI systems was presented by \citet{shokriExploitingTransparencyMeasures2020, shokriPrivacyRisksModel2021}. The study used various backpropagation and perturbation methods
to show the vulnerability of feature-based systems. The proposed attack
used variances in the prediction and explanation vectors to
differentiate between members and non-members. \citet{liuPleaseTellMe2024}
introduced a membership inference on feature-based XAI using model
performance and robustness metrics. The study observed higher loss in
confidence on perturbation of important features for members and
utilized this observation in training an attack model, in addition to
using the performance loss from the model. \citet{maLabelOnlyMembershipInference2024} extended membership inference to label-only settings on  Shapley
value explanations. This method, which builds on earlier work on label-only attacks using hard prediction labels \citep{choquette-chooLabelOnlyMembershipInference2021}, improved neighbourhood sampling using explanations thus reducing the number of queries.

In the example-based category, \citet{shokriExploitingTransparencyMeasures2020, shokriPrivacyRisksModel2021} investigated
influence functions on logistic regression models. Since influence functions generate explanations in the form of actual datapoints, the study observed that attackers could obtain certainty about membership,
thus leading to stronger attacks. More recently, \citet{cohenMembershipInferenceAttack2024} considered
self-influence functions instead, that show the influence of a datapoint
on its own prediction. The proposed attack required white-box access
to the target model parameters, activations, and gradients. The
selection of an appropriate threshold range for self-influence scores associated with members
was critical for this attack and was achieved by maximizing the balanced
accuracy on the training set.

\citet{kuppaAdversarialXAIMethods2021} used a different type of example-based
explanation, namely, counterfactuals, for membership inference. The
authors trained shadow models using counterfactual samples and auxiliary
datasets. A threshold on the difference in predictions of
the attack and target models was used to determine membership.
\citet{pawelczykPrivacyRisksAlgorithmic2023} also targeted counterfactuals and proposed two
types of attacks. The first relied on the distances between datapoints
and their counterfactuals to differentiate between members and
non-members. The second used a loss-based approach using a
likelihood-ratio test \citep{carliniMembershipInferenceAttacks2022} that improved the attack.

Interpretable models using decision trees, and surrogate models created
using Trepan algorithm \citep{cravenExtractingTreeStructuredRepresentations1995}, were evaluated for
membership inference by \citet{narettoEvaluatingPrivacyExposure2022}. The study also examined
the effect of overfitting on the attack. The success of
membership inference was determined to be higher on both interpretable and surrogate models compared to black-box models. Further, surrogates of overfitted models exhibited higher susceptibiliy to the attack than those derived from well-regularized
models.

Membership inference attacks in machine learning models have been explored
extensively in existing literature \citep{huMembershipInferenceAttacks2022} and attack strategies have
exploited confidence scores and predictions \citep{shokriMembershipInferenceAttacks2017}.
However, the recent attacks that exploit explanations suggest that XAI
interfaces provide a new avenue for adversaries to launch this attack.
Such attacks have targeted feature-based, example-based, and
interpretable (including surrogates) XAI methods. The effectiveness of the attack is influenced by factors such as
dataset type \citep{shokriPrivacyRisksModel2021}, dimension \citep{pawelczykPrivacyRisksAlgorithmic2023, shokriPrivacyRisksModel2021}, model architecture \citep{shokriPrivacyRisksModel2021} and
overfitting \citep{pawelczykPrivacyRisksAlgorithmic2023}. Some attacks have proven effective in the absence of knowledge of the training dataset or target architectures \citep{liuPleaseTellMe2024}, underscoring their practical threat potential.

While interpretable models are often recommended as surrogates for explaining
black-box models \citep{mcdermidArtificialIntelligenceExplainability2021}, as demonstrated by
these attacks, the layer of interpretability can introduce a backdoor
to the target system and lead to privacy leaks \citep{narettoEvaluatingPrivacyExposure2022}.
In the example-based category, influence functions expose data instances, particularly outliers, due to their distinct characteristics
and higher influence on the training process \citep{shokriPrivacyRisksModel2021}.
Among feature-based methods, those using perturbations exhibit higher
resilience to membership inference due to use of out-of-distribution
points, however, this can also result in reduced explanation fidelity (\citep{shokriPrivacyRisksModel2021}. Conversely, feature-based methods with better explanation quality are
also found to be susceptible to higher leakage \citep{liuPleaseTellMe2024}
suggesting a conflict between privacy and utility.

\subsubsection{Model inversion}\label{section4.2.2}
\begin{figure}[h]
\begin{center}
\includegraphics[width=0.7\linewidth]{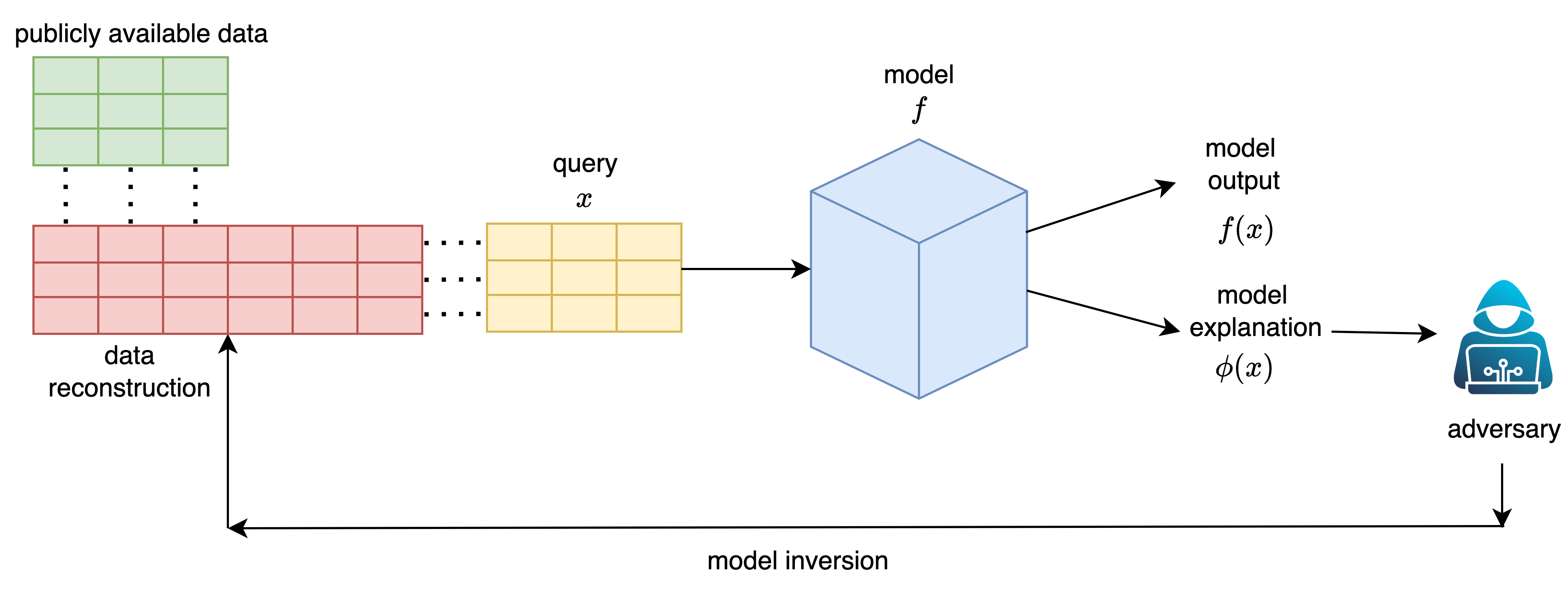}
\end{center}
\caption{Model inversion exploiting explanations.}
\label{fig:fig7}
\end{figure}

This category of privacy risk can result in reconstruction of datapoints partly or completely from outputs \citep{zhangSurveyPrivacyInference2023} (Figure~\ref{fig:fig7}). These attacks can be conducted with black-box or white-box access to the
model \citep{fredriksonModelInversionAttacks2015,vealeAlgorithmsThatRemember2018} after it has been
deployed in production systems. Attribute inference is a type of data reconstruction that can determine the values of certain attributes,
generally those sensitive to individuals \citep{yeomPrivacyRiskMachine2018} from outputs. When the model output, $f(x)$, is used for inversion, the inference model can be expressed as follows:
\begin{equation}
A_{ModInv}: f(x) \rightarrow x
\end{equation}
When explanations are available, they can be alternatively exploited to reconstruct datapoints using the following inference model:
\begin{equation}
A_{ModInvExp}: \phi(x) \rightarrow x
\end{equation}
Model inversion attacks have been documented in XAI on example-based,
feature-based, and interpretable systems. \citet{shokriExploitingTransparencyMeasures2020, shokriPrivacyRisksModel2021} demonstrated a data reconstruction attack on influence functions in
logistic regression models and found the attack dependent on data
dimensionality. The authors designed different heuristics for low and
high dimension data to improve coverage and efficiently recover more
training points. \citet{goethalsPrivacyIssueCounterfactual2023} showed an explanation linkage
attack using native counterfactuals generated from actual instances of
the training data. The attack demonstrated the vulnerability of
counterfactuals in leaking private attributes.

Private images were found to be susceptible to recovery through saliency
maps by \citet{zhaoExploitingExplanationsModel2021} leading to inadvertent exposure. The study found XAI
systems that provided class-specific multiple explanations particularly prone to leakage. The authors also used attention transfer to highlight similar risks for non-explanation models. Other studies \citep{dudduInferringSensitiveAttributes2022,luoFeatureInferenceAttack2022} have focused on attribute
inference of tabular data using feature-based XAI. The former trained
attack models using predictions and explanations to infer sensitive
features. The latter used Shapley values and effectively executed the
attack with limited number of queries on cloud ML services. \citet{tomaCombinationsAIModels2024} further showed that the efficacy of the proposed attack
was dependent on the combination of black-box architecture and XAI
method. Their findings indicate that linear models using Shapley values were particularly vulnerable
to attribute inference.

\citet{ferryProbabilisticDatasetReconstruction2024} designed a probabilistic white-box attack applicable
to transparent models, such as decision trees and rule lists, and
quantified the information about the training data contained in the
model. The work found that models built using greedy algorithms leak
more information compared to those built using optimal strategies. The
authors also observed the attack's capacity for misuse in launching
other inference attacks such as membership and property inference.

The above attacks from current literature demonstrate the potential of misuse of explanations APIs, resulting in a two-way flow \citep{vealeAlgorithmsThatRemember2018} where inputs can be determined from outputs. Model
explanations provide an effective attack surface compared to predictions \citep{dudduInferringSensitiveAttributes2022,zhaoExploitingExplanationsModel2021}, indicating the contradiction between the need for explanations in
Trustworthy AI and protecting privacy \citep{zhaoExploitingExplanationsModel2021}. Data
reconstruction attacks impact active users of AI systems, thus putting end-users at risk \citep{zhaoExploitingExplanationsModel2021} and having a higher impact. In certain proposed
model inversion attacks, sensitive attributes can be retrieved from
models trained on non-sensitive attributes \citep{dudduInferringSensitiveAttributes2022} while
other proposed attacks demonstrate higher leakage from more important
attributes \citep{luoFeatureInferenceAttack2022} and recovery of entire training
datasets \citep{shokriPrivacyRisksModel2021}. In addition, the above works highlight
the misuse of XAI techniques even for models that do not provide
explanations \citep{zhaoExploitingExplanationsModel2021}.

A tension is also found between preserving privacy and maintaining utility of the XAI system.
For instance, synthetic counterfactuals created by perturbing actual
samples are shown to provide resilience to inversion in comparison to native
counterfactuals. However, their usage is found to affect the plausibility
and runtime of explanations \citep{goethalsPrivacyIssueCounterfactual2023}, suggesting
degrading utility. Similarly, the use of multiple diverse explanations are usually recommended \citep{voFeaturebasedLearningDiverse2023} for improving understandibility of explanations, however they are also found to contribute to leakage of
privacy. Consequently restricting the access to explanation APIs has been suggested as a countermeasure \citep{zhaoExploitingExplanationsModel2021}, however such restrictions may reduce the utility to end-users.

\subsubsection{Model extraction}\label{section4.2.3}

\begin{figure}[h]
\begin{center}
\includegraphics[width=0.7\linewidth]{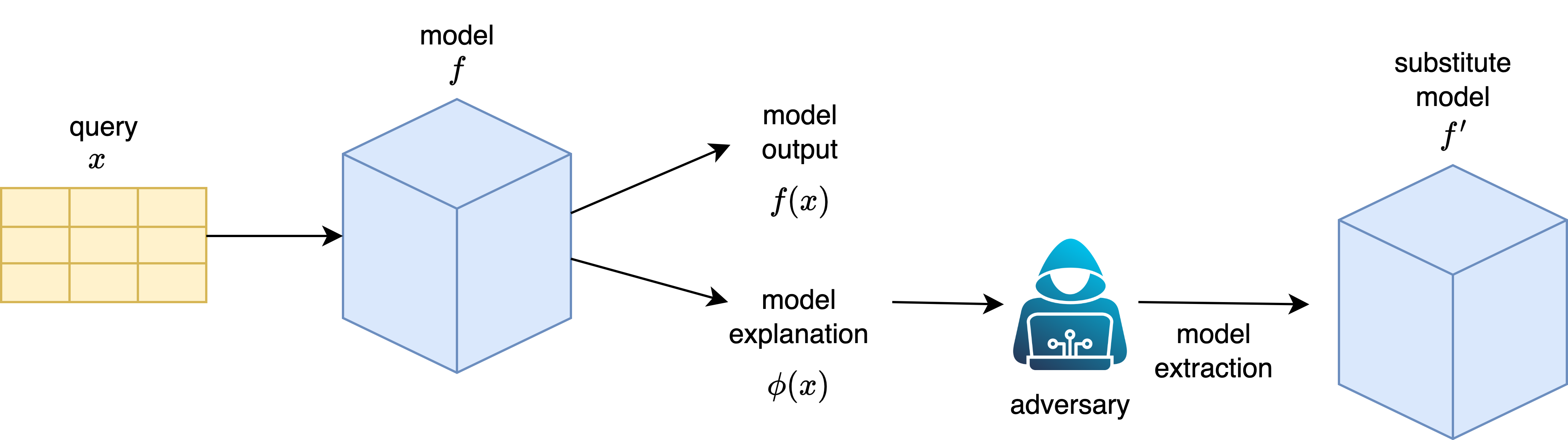}
\end{center}
\caption{Model extraction exploiting explanations.}
\label{fig:fig8}
\end{figure}

In this attack, the target model is replicated to a significant degree of accuracy and fidelity
\citep{jegorovaSurveyLeakagePrivacy2022}, thus breaching the confidentiality and the intellectual property of the model owner (Figure~\ref{fig:fig8}). In a typical model extraction attack, the adversary has
black-box access to a deployed victim model and uses an unlabeled
dataset to query it, thus building an attack dataset \citep{yanExplanationLeaksExplanationguided2023} for cloning the model. In contrast, data-free model extraction leverages generative
models to synthesize the datasets, which is advantageous when the input data is difficult to obtain \citep{miuraMEGEXDataFreeModel2024}. When the model output, $f(x)$, is utilized for this attack, the extraction of the target model function, $f'$, can be expressed as follows:
\begin{equation}
A_{ModExt}: x,f(x) \rightarrow f'
\end{equation}
When explanations are utilised for the purpose of extraction, the attack model can be expressed as follows:
\begin{equation}
A_{ModExtExp}: x,\phi(x) \rightarrow f'
\end{equation}
Since the extracted models can further leak
personal data through membership inference and model inversion \citep{songMachineLearningModels2017}, this attack is usually used as a starting point
for initiating other types of attacks \citep{aivodjiModelExtractionCounterfactual2020,miuraMEGEXDataFreeModel2024}. 

Model extraction attacks using explanations have been proposed across feature-based and
example-based XAI methods. In the seminal work on the topic, gradient
explanations, in the form of saliency maps, were found to be vulnerable
by \citet{milliModelReconstructionModel2019}. The use of explanations improved the attack by
reducing the number of queries compared to attacks relying solely on model predictions.
\citet{miuraMEGEXDataFreeModel2024} also leveraged gradient-based explanations but used
data-free approach to train generative models for creating the attack
dataset. The inclusion of explanations improved the quality of the
generative model, and the accuracy of the cloned model improved with
the diversity of the generated samples. Similarly, \citet{yanExplanationbasedDatafreeModel2023} employed data-free extraction wherein explanation loss was used to guide the generative model.
Accuracy of the cloned model was improved by matching the victim model's
predictions and explanations.

A different approach of extraction on feature-based XAI, used multitask learning to learn both
classification and explanation tasks of the victim model \citep{yanExplainableModelExtraction2022}. Further, a model agnostic technique on gradient and pertubation-based XAI \citep{yanExplanationLeaksExplanationguided2023}, showed that explanations provided auxiliary information that enabled more efficient attacks, reducing the query budget compared to prediction-only strategies. The attack could
also be applied to non-explanation models and achieved accuracy
equivalent to those of explanation models. 

Besides the above extraction attacks targeting various feature-based
XAI, from the example-based category, counterfactuals have been mainly
targeted for this attack. In an extraction attack proposed by \citet{aivodjiModelExtractionCounterfactual2020}, they were used to approximate the decision boundary of the
victim model with high accuracy and fidelity under low query budgets.
Multiple and diverse counterfactuals were found to aid the extraction
process by divulging additional information to adversaries. An
improvisation of the attack, to reduce the number of queries further,
mitigate decision boundary shift and achieve higher agreement with the
victim model, was proposed by \citet{wangDualCFEfficientModel2022}. The method used the
original counterfactual explanation with its own counterfactual as training
pairs, to extract additional datapoints to train the cloned model. In another approach, \citet{kuppaAdversarialXAIMethods2021} proposed iterative querying of the victim model to
capture the training data distribution. The method utilized distillation
loss to transfer knowledge from the victim to the cloned model and was found to be successful due to the optimization of various
properties such as diversity, proximity, feasibility, and sparsity.

As demonstrated by the aforementioned attacks, explanation-based extraction attacks offer substantial advantages over traditional prediction-only approaches by facilitating model replication with reduced number
of queries \citep{milliModelReconstructionModel2019,miuraMEGEXDataFreeModel2024}. The reduction in
the number of queries benefits the adversary, especially in pay-by-query models. Certain attacks are also
possible with partial knowledge of the data distribution \citep{aivodjiModelExtractionCounterfactual2020} or in absence of overlap between the attack and training datasets \citep{yanExplainableModelExtraction2022}. In addition, in scenarios where
attackers do not possess the input datasets, data-free extraction
attacks are possible and the use of explanations is shown to improve the attack accuracy \citep{miuraMEGEXDataFreeModel2024}. Moreover, the diversity of
the generated input datasets in such attacks is found to improve
the accuracy of the cloned models \citep{miuraMEGEXDataFreeModel2024}.

In addition to the direct threat to explanation models, XAI techniques
can be misused for extraction of non-explanation models \citep{yanExplanationLeaksExplanationguided2023}. The use of diverse explanations, intended to build
user trust in explanations, can lead to further leakage of
privacy \citep{aivodjiModelExtractionCounterfactual2020}. Similarly, the optimization of counterfactuals to
satisfy various properties to improve explanation quality, can reveal information to adversaries about class-specific decision boundaries
thus aiding the attack \citep{kuppaAdversarialXAIMethods2021} and leading to the
conflict of explainability and privacy with utility.

\subsubsection{Property inference}\label{section4.2.4}

\begin{figure}[h]
\begin{center}
\includegraphics[width=0.7\linewidth]{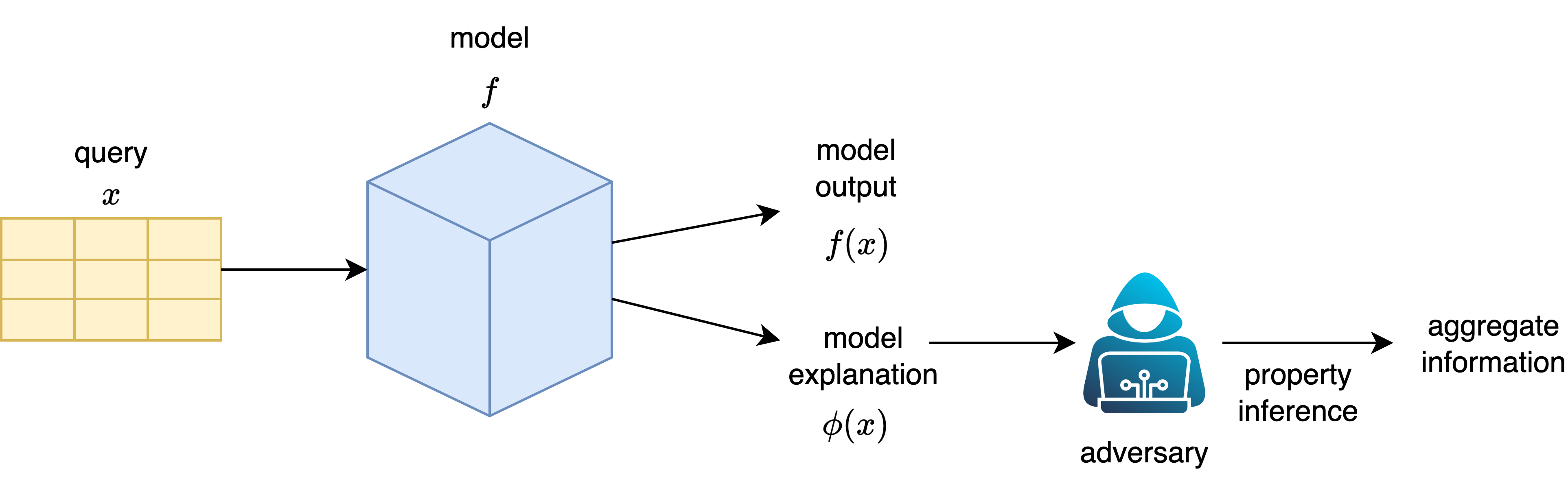}
\end{center}
\caption{Property inference exploiting explanations.}
\label{fig:fig9}
\end{figure}

This type of risk pertains to inference of properties from the training data such as
global statistics or aggregates \citep{mahloujifarPropertyInferencePoisoning2022}, which model
owners did not intend on sharing \citep{ganjuPropertyInferenceAttacks2018} (Figure~\ref{fig:fig9}). The inferred property may not correspond to features in the training data or be correlated to any feature. A property inference model that uses the model outputs, $f(x)$, to infer such property, $p$, can be expressed as the following:
\begin{equation}
A_{PropInf}: f(x) \rightarrow p
\end{equation}
When explanations, $\phi(x)$, are available, the inference model can be expressed as follows:
\begin{equation}
A_{PropInfExp}: \phi(x) \rightarrow p
\end{equation}

Although property inference is a known issue in AI models, to the best of our knowledge, no attacks have yet been documented that exploit explanations for this purpose.

\clearpage
\textbf{Table 2} Studies on intentional privacy leakage in XAI systems.
\begin{longtable}[]{@{}
  >{\raggedright\arraybackslash}p{(\linewidth - 6\tabcolsep) * \real{0.1641}}
  >{\raggedright\arraybackslash}p{(\linewidth - 6\tabcolsep) * \real{0.1804}}
  >{\raggedright\arraybackslash}p{(\linewidth - 6\tabcolsep) * \real{0.3773}}
  >{\raggedright\arraybackslash}p{(\linewidth - 6\tabcolsep) * \real{0.2781}}@{}}
\toprule\noalign{}
\label{tab:table2}
\begin{minipage}[b]{\linewidth}\centering
\textbf{Privacy Risk}
\end{minipage} & \begin{minipage}[b]{\linewidth}\centering
\textbf{XAI Category}
\end{minipage} & \begin{minipage}[b]{\linewidth}\centering
\textbf{XAI Method}
\end{minipage} & \begin{minipage}[b]{\linewidth}\centering
\textbf{Study}
\end{minipage} \\
\midrule\noalign{}
\endhead
\bottomrule\noalign{}
\endlastfoot
\multirow[t]{8}{=}{Membership inference} & Interpretable & Decision tree &
\citet{narettoEvaluatingPrivacyExposure2022} \\
\cline{2-4}
& Interpretable (surrogate) & Trepan & \citet{narettoEvaluatingPrivacyExposure2022} \\
\cline{2-4}
& \multirow[t]{3}{=}{Example-based} & Influence functions & \citet{shokriExploitingTransparencyMeasures2020, shokriPrivacyRisksModel2021} \\
\cline{3-4}
& & Counterfactuals & \citet{kuppaAdversarialXAIMethods2021,pawelczykPrivacyRisksAlgorithmic2023} \\
\cline{3-4}
& & Self-influence functions & \citet{cohenMembershipInferenceAttack2024} \\
& \multirow{3}{=}{Feature-based} & Gradient, integrated gradients,
guided backpropagation, LRP, LIME, SmoothGrad & \citet{shokriExploitingTransparencyMeasures2020, shokriPrivacyRisksModel2021} \\
\cline{3-4}
& & Integrated gradients, SmoothGrad, VarGrad, Grad-CAM++, SHAP, LIME &
\citet{liuPleaseTellMe2024} \\
\cline{3-4}
& & Shapley values & \citet{maLabelOnlyMembershipInference2024} \\
\cline{1-4}
\multirow[t]{6}{=}{Model inversion} & Interpretable & Decision tree, rule
list & \citet{ferryProbabilisticDatasetReconstruction2024} \\
\cline{2-4}
& \multirow[t]{2}{=}{Example-based} & Influence functions & \citet{shokriExploitingTransparencyMeasures2020, shokriPrivacyRisksModel2021} \\
\cline{3-4}
& & Counterfactuals (native) & \citep{goethalsPrivacyIssueCounterfactual2023} \\
\cline{2-4}
& \multirow[t]{3}{=}{Feature-based} & Gradient, gradient x input, class
activation maps (CAM), Grad-CAM, LRP & \citep{zhaoExploitingExplanationsModel2021} \\
\cline{3-4}
& & Integrated gradients, DeepLIFT, GradientSHAP, SmoothGrad & \citet{dudduInferringSensitiveAttributes2022} \\
\cline{3-4}
& & Shapley values & \citet{luoFeatureInferenceAttack2022, tomaCombinationsAIModels2024} \\
\cline{1-4}
\multirow[t]{6}{=}{Model extraction} & Example-based & Counterfactuals &
\citet{aivodjiModelExtractionCounterfactual2020,kuppaAdversarialXAIMethods2021,wangDualCFEfficientModel2022}\\
\cline{2-4}
& \multirow[t]{5}{=}{Feature-based} & Gradient & \citet{milliModelReconstructionModel2019} \\
\cline{3-4}
& & Gradient, Grad-CAM, MASK & \citet{yanExplainableModelExtraction2022} \\
\cline{3-4}
& & Gradient, Grad-CAM, MASK, LIME & \citet{yanExplanationLeaksExplanationguided2023} \\
\cline{3-4}
& & Grad-CAM, LIME & \citet{yanExplanationbasedDatafreeModel2023} \\
\cline{3-4}
& & Gradient, integrated gradients, SmoothGrad & \citep{miuraMEGEXDataFreeModel2024} \\
\cline{1-4}
Property inference & Not reported & Not reported & - \\
\end{longtable}

\subsection{Unintentional privacy leakage}\label{section4.3}

This subsection discusses unintentional privacy leakage in XAI that
occur without malicious intent \citep{jegorovaSurveyLeakagePrivacy2022}. Certain leakages
can occur due to the natural mechanisms of the training process or through the content of explanations. These may occur during the different AI lifecycle phases or during the course of an explanation \citep{rawalRecentAdvancesTrustworthy2022}.

\subsubsection{Training issues}\label{section4.3.1}

Training issues such as, overfitting and memorization, identified in AI
models can lead to privacy leakage. Overfitting is found to aid membership and attribute
inference attacks \citep{yeomPrivacyRiskMachine2018}. Memorization leads to the model
remembering subsets of training data \citep{songMachineLearningModels2017} and occurs
during training before overfitting begins \citep{jegorovaSurveyLeakagePrivacy2022}. It
can cause leakage when data owners deploy models with codebases and training pipelines developed by third parties, such as in MLaaS, allowing sensitive information to be
leaked from training data \citep{songMachineLearningModels2017}.

\subsubsection{Explanation content}\label{section4.3.2}

The content of explanations may contain values of sensitive
fields. For instance, in example-based explanations such as influence
functions, training datapoints potentially containing sensitive fields,
are directly revealed to end-users. \citet{karimiSurveyAlgorithmicRecourse2023} provide another
example of unintentional leakage through contrastive explanations, which
can lead to inference of sensitive details of individuals whose partial
attributes are known. Additionally, interpretable models used as surrogates, can
reveal properties of the training data or additional information about
the black-box beyond what the model owner intended to share \citep{blanco-justiciaMachineLearningExplainability2020}.
In addition to direct content-based leakage, risks may also arise from the inadvertent exposure of explanations to unauthorised users \citep{kuppaAdversarialXAIMethods2021}. For example, during troubleshooting of error
cases, developers or quality engineers may unintentionally access
sensitive personal information in the explanation. Moreover, even when direct identifiers are absent, explanations that contain proxy or correlated features can still enable indirect inference. 

\section{Privacy Preservation Methods in XAI}\label{section5}

\begin{figure}[h]
\begin{center}
\includegraphics[width=0.7\linewidth]{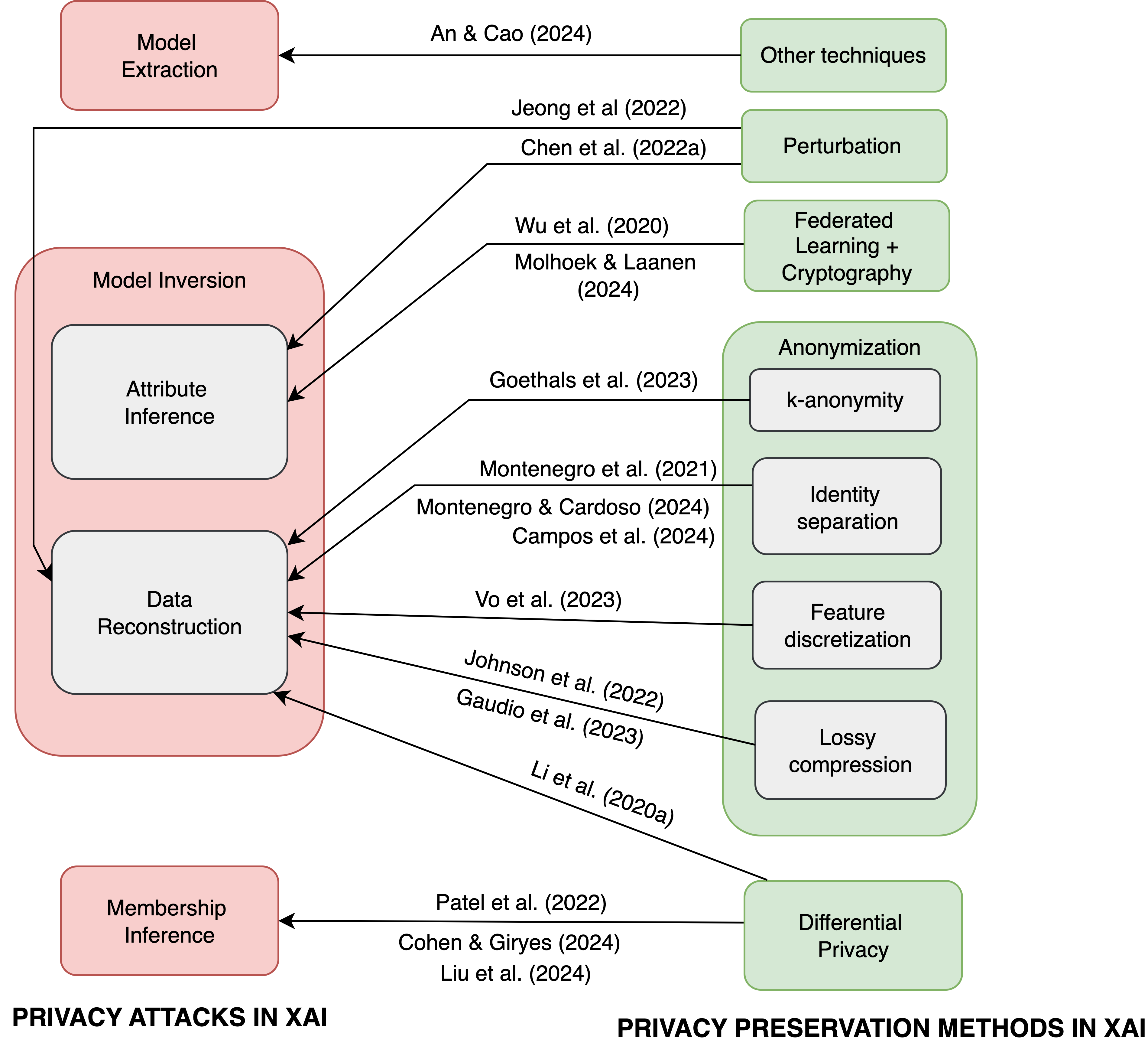}
\end{center}
\caption{Proposed privacy preservation methods for specific
privacy attacks in XAI.}
\label{fig:fig10}
\end{figure}

To address the privacy risks outlined in Section~\ref{section4}, a growing body of research has emphasized the need for privacy preserving XAI techniques \citep{aivodjiModelExtractionCounterfactual2020,shokriPrivacyRisksModel2021,zhaoExploitingExplanationsModel2021}. In response to these concerns, several studies have proposed methods to generate explanations while  mitigating privacy concerns. Many of these approaches draw upon established principles and methods from the broader domain of privacy preserving ML (PPML), adapting them to specific challenges posed by explanation techniques. This section synthesizes the key contributions in literature to enhance the privacy guarantees of XAI systems in alignment with the objectives of RQ2. We categorize the proposed solutions under the main
approaches in PPML. Table~\ref{tab:table3} summarizes these approaches and methods, and
Figure~\ref{fig:fig10} presents a consolidated mapping of privacy preserving strategies to specific types of privacy attacks discussed earlier.

\subsection{Differential privacy}\label{section5.1}

Differential privacy (DP) \citep{dworkOurDataOurselves2006} is a widely recognised method that
provides a quantifiable definition of privacy and the incremental
privacy loss from publishing confidential data \citep{mckaybowenPhilosophyDifferentialPrivacy2021}. A mechanism is differentially private if it can hide
the participation of any single individual in a dataset \citep{harderInterpretableDifferentiallyPrivate2020}. This can be achieved by using noise and is typically associated
with an adverse effect on the accuracy of the system \citep{harderInterpretableDifferentiallyPrivate2020}. By adjusting the privacy budget, $\epsilon$, from 0 to $\infty$, practitioners
can manage this trade-off between privacy and accuracy \citep{mckaybowenPhilosophyDifferentialPrivacy2021}. Given its robust privacy guarantees, early research
in privacy preserving explanations has adopted DP using various
strategies to safeguard the training data in interpretable,
feature-based, and example-based XAI. In the context of XAI, an
explanation is differentially private if it can obscure any single
individual in the training dataset \citep{patelModelExplanationsDifferential2022}.
The technique can be applied at various stages, including the explanation generation algorithm \citep{patelModelExplanationsDifferential2022}, the training process of the target model \citep{cohenMembershipInferenceAttack2024,liuPleaseTellMe2024} or directly on the training data \citep{bozorgpanahExplainableMachineLearning2024,ezzeddineDifferentialPrivacyAnomaly2024}.

Decision trees are popular due to their simplicity and inherent
interpretable qualities, however they are prone to privacy leakage \citep{ferryProbabilisticDatasetReconstruction2024,narettoEvaluatingPrivacyExposure2022}. A number of algorithms have been developed for
building decision trees based on DP guarantees \citep{fletcherDecisionTreeClassification2020} offering  different trade-offs in privacy and
utility. The interpretability of private trees and their subsequent
usefulness to XAI users, depend on factors such as the privacy budget
per query, tree depth, pruning strategies and termination criteria
\citep{fletcherDecisionTreeClassification2020}. \citet{noriAccuracyInterpretabilityDifferential2021} applied DP to another type of
interpretable model, namely, Explainable Boosting Machines (EBM), to prevent privacy leakage of training data. The resulting privatized
system demonstrated good accuracy at low privacy budgets while facilitating correction of errors introduced by noise, the removal of bias and the 
enforcement of constraints such as monotonicity. Building on this, \citet{baekDifferentiallyPrivateExplainable2024}
further optimized the utilization of privacy budget in these models to
improve accuracy, incorporating techniques such as gradient error optimization and pruning of
non-essential features.

\citet{harderInterpretableDifferentiallyPrivate2020} proposed an interpretable model using differentially private locally linear maps with Gaussian mechanism per output class. The filters learned by the model from input images were observed
to have higher interpretability compared to feature-based methods, such
as integrated gradients and SmoothGrad. However, increasing the number
of such maps per class dropped accuracy due to the distribution of
privacy budget over additional parameters. In a different approach, \citet{liDifferentialPrivacyPreservation2020} proposed an interpretable
model in the form of feedforward-designed convolutional neural network
(FF-CNN) made privacy preserving by using DP
guarantees on subspace approximation with adjusted bias (Saab). Their findings indicated the the integration of DP was effective in mitigating the risk of reconstruction
of input images while maintaining classification accuracy.

For feature-based explanations generated by local linear approximations
around the point of interest, \citet{patelModelExplanationsDifferential2022} introduced a
differentially private approach for loss calculation in the explanation
algorithm. The study also proposed an adaptive method of reusing
previous explanations for prudent usage of the privacy budget. \citet{nguyenXRandDifferentiallyPrivate2023} employed local DP to restrict adversaries from learning the
top influential features through aggregated scores in feature-based XAI.
While originally proposed to defend against a backdoor security attack
exploiting explanations \citep{severiExplanationGuidedBackdoorPoisoning2021}, the random perturbation
of influential features under local DP guarantees was observed to
preserve the privacy of those features while maintaining explanation
fidelity. \citet{bozorgpanahExplainableMachineLearning2024} also applied local DP to mask the
training dataset and investigated its impact on privacy and utility of
feature-based explanations. They introduced an irregularity metric to
measure the feature distortion due to privatization of the original
dataset and the change in explanation values. Their findings indicate that the use of additive noise
on the training dataset caused irregularities, thereby
reducing the utility of the explanations. In a related study, \citet{ezzeddineDifferentialPrivacyAnomaly2024} added
calibrated noise to training datasets and evaluated the impact on SHAP
explanations using various distance metrics. They observed the change in SHAP values
in the privatized systems correlated with the privacy
budget and data dependent. \citet{abbasiFurtherInsightsBalancing2024} used a different approach on the data and employed DP
for synthetic data generation for training of different
model architectures. They used similarity scores to track the change in
explanations while utility loss evaluated the drop in accuracy, thus
quantifying the triad of privacy, utility and explainability.

In addition to the aforementioned studies, researchers have also examined the use of DP
in the training process of the target model in feature-based \citep{liuPleaseTellMe2024} and example-based \citep{cohenMembershipInferenceAttack2024,mochaourabDemonstratorCounterfactualExplanations2023} XAI. These investigations have determined mitigation against membership inference attacks for
high privacy budgets \citep{cohenMembershipInferenceAttack2024, liuPleaseTellMe2024}. The introduction of DP
noise serves as a regularization mechanism for target models \citep{noriAccuracyInterpretabilityDifferential2021}
and its mathematical guarantee enables quantification of privacy, making
it a compelling choice as a privacy enhancing technology. In the context of XAI, in
addition to its application in mitigating membership inference from
explanations \citep{cohenMembershipInferenceAttack2024, liuPleaseTellMe2024, patelModelExplanationsDifferential2022}, it is also found to mitigate reconstruction of sensitive inputs
\citep{liDifferentialPrivacyPreservation2020}. However, the improvement in mitigation of attacks
at high privacy budgets and hence degrading accuracy \citep{cohenMembershipInferenceAttack2024, liuPleaseTellMe2024} can be a setback to the use of this
technique. In addition to adversely affecting accuracy, its introduction
can also deteriorate explanation quality in terms of fidelity \citep{liuPleaseTellMe2024,patelModelExplanationsDifferential2022} with pronounced effects on
minority groups \citep{patelModelExplanationsDifferential2022}. The technique is also ineffective
against attribute inference attacks when there are existing strong
correlations between different attributes \citep{chenAchievingTransparencyReport2022}. In such cases, the privacy enhancing benefits of DP may be insufficient to prevent adversaries from inferring sensitive attributes. 

Algorithms such as DPSaab \citep{liDifferentialPrivacyPreservation2020}, have been observed to offer a more favourable trade-off between accuracy and privacy. Practitioners can employ strategies such
as reusing previously generated differentially private explanations to utilize the privacy budget effectively \citep{patelModelExplanationsDifferential2022}. Methods that use local DP, are observed to achieve high faithfulness
of explanations with privacy \citep{nguyenXRandDifferentiallyPrivate2023}, thus demonstrating
that it is possible to balance multiple desirable properties. Hence
judicious use of this technique can ensure that privacy is
achieved while maintaining reasonable utility of the model and
explanations.

\subsection{Cryptography}\label{section5.2}

Cryptographic protocols for privacy preservation in ML use secure
algorithms to protect the target model and data. Prominent methods in this
category include homomorphic encryption, secret sharing, and secure
multi-party computation \citep{yinComprehensiveSurveyPrivacypreserving2022}. In XAI, these techniques have seen limited application in interpretable and example-based systems. They have also been used in conjunction with other privacy preserving
techniques such as federated learning \citep{elzeinPrivaTreeCollaborativePrivacyPreserving2024,molhoekSecureCounterfactualExplanations2024,wuPrivacyPreservingVertical2020,wuFalconPrivacyPreservingInterpretable2023}.

For interpretable tree-based models, \citet{zhaoEfficientPrivacypreservingTreebased2023} proposed an
additive homomorphic scheme for model owners and query users, for pushing
the encrypted model and query data to cloud service providers for
inferencing. Adding perturbations to the inference results and query
data ensured privacy protection of these assets while maintaining accuracy
comparable to non-private inference. In the example-based category,
\citet{veugenPrivacyPreservingContrastiveExplanations2022} proposed a cryptographic method with secure
multi-party computation to generate contrastive explanations, while
protecting private training data and confidentiality of the model. The
algorithm securely trained a binary decision tree to generate fact and
foil leaves, which were used as explanations for a query datapoint. Additionally, a synthetic datapoint from the foil leaf was provided to the
end-user to enhance explainability.

Cryptographic methods, such as homomorphic encryption, introduce significant 
computational complexity in the system \citep{liuWhenMachineLearning2022}. The use of
encryption can deteriorate model transparency, limiting the ability of data scientists to
correct errors, inspect data, add features or fine
tune the model \citep{dowlinCryptoNetsApplyingNeural2016}. Therefore it is essential to implement cryptographic
protocols in XAI system components in the right use cases to complement
other privacy preservation techniques or when other techniques are
infeasible or costly.

\subsection{Anonymization}\label{section5.3}

Anonymization refers to the process of transformation of data \citep{majeedAnonymizationTechniquesPrivacy2021} to obscure the distinctive features of individuals, thus safeguarding
their privacy. The process is associated with the removal or
modification of direct and quasi-identifiers \citep{majeedAnonymizationTechniquesPrivacy2021}, that
can uniquely identify individuals. Various methods of anonymization
are used in practice, such as k-anonymity, l-diversity, and t-closeness \citep{yinComprehensiveSurveyPrivacypreserving2022}. In XAI, different anonymization techniques have been demonstrated in
example-based, feature-based, and interpretable methods. Techniques such as disentangled representation learning and lossy
compression have been applied on sensitive visual data, such as medical images, to
generate privatized explainable-by-design representations.

A dataset is considered k-anonymous if every record is indistinguishable
from k-1 other records \citep{sweeneyKanonymityModelProtecting2002}, thus providing a measure of the
risk of re-identification of records. K-anonymity can be achieved using
methods such as suppression and generalization \citep{sweeneyAchievingKanonymityPrivacy2002}, which obscure the data to remove identifiable features. Though
traditionally this technique is applied to target datasets for
protection, \citet{goethalsPrivacyIssueCounterfactual2023} proposed its usage on native
counterfactuals, that are actual datapoints from the training dataset,
for protection against model inversion attack through explanations. This
strategy of generating k-anonymous counterfactuals was shown to result in
lower information loss and higher validity, outperforming
counterfactuals derived from k-anonymized datasets. Further analysis by \citet{berningTradeoffPrivacyQuality2024} determined that the 
effectiveness of k-anonymous counterfactuals is confined to dense areas of the dataset. Its
offered privacy protection was also found disproportionate to the value of k. \citet{voFeaturebasedLearningDiverse2023} highlighted another limitation, namely, the computational overhead of generating these counterfactuals requiring
querying the explainer for a large number of counterfactuals. The authors proposed an alternative strategy of
privatizing diverse counterfactuals by discretization of continuous
features. This technique is closely related to generalization in
privacy preserving data mining and is particularly effective against linkage attacks.

K-anonymity using microaggregation has  been applied on the training dataset in
feature-based XAI by \citet{bozorgpanahExplainableMachineLearning2024}. The study found explanations from  non-private and private datasets largely aligned, with minor
irregularities observed. The alignment indicated that utility was effectively preserved after
privatization. Similarly, \citet{blanco-justiciaMachineLearningExplainability2020} applied
microaggregation to build local tree-based surrogate explanations from
clusters around an instance to be explained. The method enforced k-anonymity by
restricting the cluster size and incorporated shallow trees to enable comprehensibility.

An emerging area of study focusses on providing explanations while protecting privacy in the medical
domain, where privacy of patients' visual data is crucial.
The primary aim of such techniques is transformation of private data through
removal of identifying features while retaining explanatory evidence.
Strategies such as the use of autoencoders for disentanglement of
identifiable characteristics \citep{montenegroAnonymizingMedicalCasebased2024}, Siamese network for increasing identity distance between
original and privatized images \citep{montenegroPrivacyPreservingGenerativeAdversarial2021} and latent diffusion models for generating synthetic images \citet{camposLatentDiffusionModels2024} are proposed. The use of lossy compression by pixel sampling is also observed to remove identification information while being post-hoc explainable  \citep{gaudioDeepFixCXExplainablePrivacypreserving2023,johnsonHeartSpotPrivatizedExplainable2022}. This approach also has an added advantage of reducing the image size significantly, thus making medical
training datasets smaller \citep{gaudioDeepFixCXExplainablePrivacypreserving2023}.

In critical domains, such as healthcare, anonymizing training and query data
can assist in protecting identifiable information. However, the applied
techniques should preserve the output quality for utility to diverse end-users \citep{camposLatentDiffusionModels2024}.  Unlike DP, anonymization
techniques lack proven guarantees, however despite DP's theoretical
guarantees, it is unable to scale beyond low resolution image data \citep{camposLatentDiffusionModels2024}. Lossy compression alternatively provides an effective way of
privatizing image data, with the benefits of achieving both privacy and
explainability while reducing training dataset sizes. It thus enables data
sharing with multiple parties in non-private settings \citep{gaudioDeepFixCXExplainablePrivacypreserving2023} and can serve as an explanation generation method for sensitive image data.

Anonymization techniques, such as k-anonymity, protect privacy of individuals by mitigating
re-identification and linkage attacks \citep{voFeaturebasedLearningDiverse2023}. When applying
k-anonymity, selecting an appropriate value of k is critical in striking the right balance between accuracy and
privacy risk level \citep{bozorgpanahExplainableMachineLearning2024}. While higher
values of k enhance privacy, explainability may be adversely
affected \citep{berningTradeoffPrivacyQuality2024, blanco-justiciaMachineLearningExplainability2020}. Moreover, the actual level of privacy may also not scale with increasing values of k \citep{berningTradeoffPrivacyQuality2024}. Hence the selection of k that achieves the right trade-off in
privacy, explainability and utility is important. Additionally,
k-anonymity has limitations such as its dependence on data
characteristics, susceptibility to homogeneity attack \citep{berningTradeoffPrivacyQuality2024}, and its vulnerability to privacy leakage when background
knowledge is available or diversity is lacking in the private attributes
\citep{goethalsPrivacyIssueCounterfactual2023}. Other techniques such as l-diversity and
t-closeness may address some of these challenges, though their applicability to explanations remain unexplored. 

The generation of synthetic data for privacy preserving data analysis is
explored in previous non-XAI works \citep{boedihardjoPrivacySyntheticData2023,liuPrivacyPreservingSyntheticData2022}. Generating synthetic data that is private, accurate and
preserves properties of the true data is a known challenge and NP-hard
in the worst case \citep{ullmanPCPsHardnessGenerating2011}. When models are trained on
such data, the explanations through XAI tools are expected to be
inherently privacy preserving, hence this approach can be a promising direction in preserving privacy in explainable systems.

\subsection{Perturbation}\label{section5.4}

Perturbation of sensitive data is a widely recognized technique in the field of privacy
preserving data publishing \citep{tranComprehensiveSurveyTaxonomy2024,yinComprehensiveSurveyPrivacypreserving2022}. When
explanations contain sensitive information, obfuscating the contents through perturbations can
prevent direct exposure. This technique can also be applied to
stem indirect leakage of inferencing sensitive attributes through
explanations.

\citet{chenAchievingTransparencyReport2022} proposed a generic privacy preserving mechanism
applicable to different XAI types such as feature-based and
interpretable surrogates. The proposed method perturbed the decision
mapping of an algorithm prior to public release of transparency reports.
To mitigate privacy leakage while upholding utility, the authors defined
a maximum confidence measure in the inference of sensitive attributes of
data subjects and a utility measure in terms of faithfulness. \citet{jeongLearningGenerateInversionResistant2022} applied perturbations on saliency map explanations as a defense mechanism for model inversion in image models. The proposed framework comprised of a two-player minimax game between
inversion and noise injector networks, in which the inversion network
attempted to reconstruct images from saliency maps and the noise injector
perturbed explanations to counter the inversion. The use of multiple
evaluation metrics to differentiate between original and
reconstructed images facilitated the quantification of the privacy of the defense
mechanism.

For the prevention of privacy leakage in XAI, researchers have attempted
perturbation of two types of model outputs, namely, predictions and
explanations. Adding perturbations to model predictions, such as the
strategy of adding noise to output confidence scores used by MemGuard
\citep{jiaMemGuardDefendingBlackBox2019}, is found ineffective in mitigating membership
inference through explanations \citep{liuPleaseTellMe2024}. Perturbation of
explanations is also insufficient in defending against data-free
model extraction based on explanations \citep{yanExplanationbasedDatafreeModel2023}.
However, the strategy has shown promising results in countering model
inversion. The use of perturbation techniques at the explanation
interface is also attractive due to its ease of implementation that requires no
retraining of the model \citep{jeongLearningGenerateInversionResistant2022}. Nevertheless, large magnitude noise can degrade the usefulness of
explanations \citep{jeongLearningGenerateInversionResistant2022}, hence perturbations should be carefully
calibrated to minimize any adverse impact on explanation quality.

\subsection{Federated learning}\label{section5.5}

Among the distributed privacy enhancing techniques available in PPML, Federated
learning (FL) is an architectural solution \citep{elmestariPreservingDataPrivacy2024}
that enables training of local models on user devices and exchange of
model parameters with a centralized server that co-ordinates the
training of a shared global model \citep{konecnyFederatedLearningStrategies2016}. It thus
enables collaborative learning while keeping users' private data at the
source \citep{guerra-manzanaresPrivacyPreservingMachineLearning2023} and mitigates the privacy risk
of multiple parties sharing their sensitive data with other parties \citep{elzeinPrivaTreeCollaborativePrivacyPreserving2024} or a centralized server \citep{zhuHorizontalFederatedLearning2022}. In
horizontal federated learning (HFL), local datasets have the same
feature space but contain different samples while vertical federated learning
(VFL), involves datasets with different feature spaces but overlaps in samples
\citep{fiosinaInterpretablePrivacyPreservingCollaborative2022}.

To address both privacy and explainability in Trustworthy AI, the
combination of FL and XAI, i.e., Fed-XAI is suggested \citep{barcenaFedXAIFederatedLearning2022,corcuerabarcenaEnablingFederatedLearning2023,lopez-blancoFederatedLearningExplainable2023} and
refers to the federated learning of XAI models. Many approaches of
Fed-XAI using HFL and VFL are proposed in literature. \citet{fiosinaInterpretablePrivacyPreservingCollaborative2022}
used a HFL approach for forecasting taxi trip duration and applied
feature-based explainability methods post-hoc. \citet{chenEVFLExplainableVertical2022}
used an explainable VFL framework to optimize counterfactual
explanations using a representative query distributed on multiple
parties. Both setups demonstrate the use of post-hoc explainability
tools in a distributed environment, with FL serving as a privacy
preserving setup for collaborative learning of sensitive data owned by
multiple parties. Fed-XAI architectures have also leveraged
interpretable models locally, such as fuzzy rule-based classifiers
\citep{daoleTrustworthyAIHeterogeneous2024}, Takagi-Sugeno \citep{zhuHorizontalFederatedLearning2022} and
Takagi-Sugeno-Kang \citep{corcuerabarcenaEnablingFederatedLearning2023} fuzzy rule-based
models. In these setups, interpretability is achieved using underlying
explainable-by-design \citep{corcuerabarcenaEnablingFederatedLearning2023} models.

Though FL aids privacy by default, it is prone to reconstruction and
inferencing attacks \citep{mothukuriSurveySecurityPrivacy2021,zhangSurveyTrustworthyFederated2024}. The
sharing of gradients and model parameters, communication mechanism and
aggregation process can lead to leakage of privacy of the participating
clients \citep{zhangSurveyTrustworthyFederated2024}. Hence researchers have proposed
integration of other privacy preserving techniques, such as
cryptography, with FL methods. In one such work, \citet{molhoekSecureCounterfactualExplanations2024} generated synthetic data on vertically partitioned data in a FL
two-party setup. Counterfactuals built from this synthetic data using
secure multi-party computation, were ranked and shared with both
parties and were found to be resilient to attribute inference. \citet{elzeinPrivaTreeCollaborativePrivacyPreserving2024} proposed a HFL structure using decision tree models,
wherein a global decision tree was collaboratively trained by
participants and additive
secret-sharing was used in aggregation of intermediate results. A VFL technique, Falcon \citep{wuFalconPrivacyPreservingInterpretable2023}, utilized a
hybrid approach consisting of partially homomorphic encryption (PHE) and
additive secret sharing for exchange of intermediate computations. Another setup, Pivot \citep{wuPrivacyPreservingVertical2020}, proposed as part of
Falcon, used threshold
partially homomorphic encryption (TPHE) and additive secret sharing to protect privacy of intermediate exchanges. Though these works
successfully integrate cryptographic techniques with FL, research has
also determined that the use of cryptographic methods in FL reduces the centralized server's ability to differentiate true model parameters leading to backdoor attacks \citep{guoADFLPoisoningAttack2022}.
Hence appropriate defense frameworks, such as trust evaluation schemes
\citep{guoTFLDTTrustEvaluation2023}, should be incorporated for protection of the FL
system from malicious users.

FL enables the training of AI models from diverse, private, and
high-quality data \citep{zhuHorizontalFederatedLearning2022} located at client systems. It
reduces the footprint of user data in the network \citep{mothukuriSurveySecurityPrivacy2021} by keeping data at the source and avoids transmission and
storage of sensitive information in a centralized location when multiple
parties are involved \citep{wangMeasureContributionParticipants2019}. Despite its benefits, in
its current form FL faces challenges for its risk-free adoption
\citep{mothukuriSurveySecurityPrivacy2021} including ensuring privacy constraints, merging
of local XAI models and dealing with large data streaming that can lead
to concept drifts \citep{barcenaFedXAIFederatedLearning2022}. The introduction of XAI
methods in the FL architecture, can also further increase the
vulnerability of the system to privacy attacks through explanations.
Thus Fed-XAI presently cannot guarantee privacy preservation through XAI
components and further research to develop strategies to
stem inadvertent privacy leakage through explanations is essential.

\subsection{Other techniques}\label{section5.6}

In addition to the main privacy preservation methods commonly employed in PPML, certain
non-standard techniques have also been explored to mitigate privacy leakage in certain types of XAI. These approaches aim to enhance privacy preservation by adopting alternative strategies including limiting access to training data, obscuring
features, or providing an abstraction of the target models. While not traditionally classified under formal privacy methods, these approaches contribute to reducing privacy leakage and complement other methods. 

A client-centric, data-driven approach of generating counterfactuals was
proposed by \citet{anCounterfactualExplanationWill2024} by leveraging previous inferences
retrieved by the model user. Due to the generation of counterfactuals
locally at the client, the method was shown to be resilient to model
extraction while achieving desirable properties such as diversity and
succinctness. In another approach to create an interpretable model from a neural network, \citet{martonExplainingNeuralNetworks2024} described a data-free strategy of
distilling the function represented by the model. The method used synthetic data to train a set of
neural networks and extracted the parameters to train an
Interpretation-Net with an output representation in the form of
surrogate decision trees.

Using a knowledge-based approach, \citet{rozanecKnowledgeGraphbasedRich2022} applied semantic
technologies in the form of domain specific ontology and knowledge
graphs, to enhance explanations and describe features on a higher
conceptual level. This enabled delinking explanations from features,
thus preserving the confidentiality of the underlying model. Further,
the integration of feature-based XAI such as LIME, enabled the system to
determine features important for predictions. \citet{terziyanExplainableAIIndustry2022} also applied semantic techniques to build XAI consisting of decision trees
and rules generated from targeted points around the decision boundary of
black-box models, without accessing the original training data. Due to
the interoperability of semantic rules, the method enabled usage in 
decentralized setup for collaborative decision making, without
individual parties sharing private local data.

These works demonstrate the application of data-free and knowledge-driven
techniques in XAI to build privacy-by-design systems for protection of training data and the confidentiality
of the model. By disconnecting features from the model and 
creating abstraction layers for
generation of explanations \citep{rozanecKnowledgeGraphbasedRich2022}, these strategies are helpful in protecting the underlying assets.

\textbf{Table 3} Privacy preserving methods applied to XAI systems.

\begin{longtable}[]{@{}
  >{\raggedright\arraybackslash}p{(\linewidth - 8\tabcolsep) * \real{0.1586}}
  >{\raggedright\arraybackslash}p{(\linewidth - 8\tabcolsep) * \real{0.2137}}
  >{\raggedright\arraybackslash}p{(\linewidth - 8\tabcolsep) * \real{0.1455}}
  >{\raggedright\arraybackslash}p{(\linewidth - 8\tabcolsep) * \real{0.2589}}
  >{\raggedright\arraybackslash}p{(\linewidth - 8\tabcolsep) * \real{0.2233}}@{}}
\toprule\noalign{}
\label{tab:table3}
\begin{minipage}[b]{\linewidth}\centering
\textbf{Privacy Preservation Category}
\end{minipage} & \begin{minipage}[b]{\linewidth}\centering
\textbf{Privacy Preserving Algorithm}
\end{minipage} & \begin{minipage}[b]{\linewidth}\centering
\textbf{Protected Asset}
\end{minipage} & \begin{minipage}[b]{\linewidth}\centering
\textbf{XAI Category (Method)}
\end{minipage} & \begin{minipage}[b]{\linewidth}\centering
\textbf{Study}
\end{minipage} \\
\midrule\noalign{}
\endhead
\bottomrule\noalign{}
\endlastfoot
\multirow[t]{12}{=}{Differential privacy} & Various DP training algorithms
& Training data & Interpretable (decision trees) & \citet{fletcherDecisionTreeClassification2020} \\
\cline{2-5}
& DP locally linear maps & Training data & Interpretable (locally linear
maps) & \citet{harderInterpretableDifferentiallyPrivate2020} \\
\cline{2-5}
& DPSaab & Training data & Interpretable (FF-CNN) & \citet{liDifferentialPrivacyPreservation2020} \\
\cline{2-5}
& DP-EBM & Training data & Interpretable (EBM) & \citet{baekDifferentiallyPrivateExplainable2024, noriAccuracyInterpretabilityDifferential2021} \\
\cline{2-5}
& DP explanation generation & Training and query data & Feature-based
methods using local linear approximations (LIME, etc.) & \citet{patelModelExplanationsDifferential2022} \\
\cline{2-5}
& Local DP & Training data & Feature-based methods that aggregate scores
(SHAP, etc.) & \citet{nguyenXRandDifferentiallyPrivate2023} \\
\cline{2-5}
& DP trained SVM & Training data & Example-based (counterfactuals) &
\citet{mochaourabDemonstratorCounterfactualExplanations2023} \\
\cline{2-5}
& DP-SGD & Training data & Feature-based (Grad-CAM) & \citet{liuPleaseTellMe2024} \\
\cline{2-5}
& DP-RMSProp & Training data & Example-based (self-influence functions)
& \citet{cohenMembershipInferenceAttack2024} \\
\cline{2-5}
& Local DP & Training data & Feature-based (TreeSHAP) & \citet{bozorgpanahExplainableMachineLearning2024} \\
& Local DP & Training data & Feature-based (SHAP) & \citet{ezzeddineDifferentialPrivacyAnomaly2024} \\
\cline{2-5}
& DP-WGAN (Wasserstein GAN) & Training data & Various XAI methods from
DALEX framework\footnote{DALEX framework is available on
\url{https://github.com/modeloriented/dalex}} & \citet{abbasiFurtherInsightsBalancing2024} \\
\hline
\multirow[t]{2}{=}{Cryptography} & Privacy preserving foil trees & Training
data, model & Example-based (contrastive explanations) & \citet{veugenPrivacyPreservingContrastiveExplanations2022} \\
\cline{2-5}
& Additive homomorphic encryption & Query data, inference results, model
& Interpretable (tree-based models) & \citet{zhaoEfficientPrivacypreservingTreebased2023} \\
\hline
\multirow[t]{9}{=}{Anonymization} & Microaggregation (MDAV) & Training
data, model & Interpretable (decision trees) & \citet{blanco-justiciaMachineLearningExplainability2020} \\
\cline{2-5}
& Privacy preserving generative model & Training data & Example-based
(case-based) & \citet{montenegroPrivacyPreservingGenerativeAdversarial2021} \\
\cline{2-5}
& HeartSpot

(lossy compression) & Training data & Feature-based

(saliency maps) & \citet{johnsonHeartSpotPrivatizedExplainable2022} \\
\cline{2-5}
& Discretization of features (generalization) & Training data &
Example-based (counterfactuals) & \citet{voFeaturebasedLearningDiverse2023} \\
\cline{2-5}
& CF-K

(k-anonymity of counterfactuals) & Training data & Example-based (native
counterfactuals) & \citet{berningTradeoffPrivacyQuality2024,goethalsPrivacyIssueCounterfactual2023} \\
\cline{2-5}
& DeepFixCX

(lossy compression) & Training data & Feature-based (saliency maps) &
\citet{gaudioDeepFixCXExplainablePrivacypreserving2023} \\
\cline{2-5}
& Microaggregation (MDAV) & Training data & Feature-based (TreeSHAP) &
\citet{bozorgpanahExplainableMachineLearning2024} \\
\cline{2-5}
& Disentangled representation learning & Training data & Example-based
(case-based) & \citet{montenegroAnonymizingMedicalCasebased2024} \\
\cline{2-5}
& Latent diffusion models & Training data & Example-based (case-based) &
\citet{camposLatentDiffusionModels2024} \\
\hline
\multirow[t]{2}{=}{Perturbation} & GNIME & Training and query data &
Feature-based (saliency maps) & \citet{jeongLearningGenerateInversionResistant2022} \\
\cline{2-5}
& Linear-Time Optimal Privacy Scheme & Training and query data & Various
XAI methods (interpretable surrogates, feature-based, etc.) & \citet{chenAchievingTransparencyReport2022} \\
\hline
\multirow[t]{7}{=}{Federated learning} & Pivot (VFL, additive secret
sharing, TPHE) & Training data & Tree-based models (transparent) & \citet{wuPrivacyPreservingVertical2020} \\
\cline{2-5}
& HFL & Training data & Feature-based methods (DeepLIFT, integrated
gradients, LIME, etc.) & \citet{fiosinaInterpretablePrivacyPreservingCollaborative2022} \\
\cline{2-5}
& HFL & Training data & Interpretable (Takagi-Sugeno,Takagi--
Sugeno--Kang, fuzzy rule-based classifier) & \citet{corcuerabarcenaEnablingFederatedLearning2023,daoleTrustworthyAIHeterogeneous2024, zhuHorizontalFederatedLearning2022} \\
\cline{2-5}
& VFL & Training data & Counterfactuals & \citet{chenEVFLExplainableVertical2022} \\
\cline{2-5}
& Falcon (VFL, additive secret sharing, PHE) & Explanations and training
data & Feature-based (LIME) & \citet{wuFalconPrivacyPreservingInterpretable2023} \\
\cline{2-5}
& PrivaTree

(HFL, additive secret sharing) & Training data & Decision trees
(transparent) & \citet{elzeinPrivaTreeCollaborativePrivacyPreserving2024} \\
\cline{2-5}
& VFL, SMC, Synthetic data & Query data & Example-based
(counterfactuals) & \citet{molhoekSecureCounterfactualExplanations2024} \\
\hline
\multirow[t]{4}{=}{Other techniques} & Semantic XAI & Training data &
Interpretable (decision trees, semantic rules) & \citet{terziyanExplainableAIIndustry2022} \\
\cline{2-5}
& Semantic technologies (knowledge graphs, ontologies) & Model &
Feature-based (LIME, etc.) & \citet{rozanecKnowledgeGraphbasedRich2022} \\
\cline{2-5}
& Guarded counterfactuals & Training data, model & Example-based
(counterfactuals) & \citet{anCounterfactualExplanationWill2024} \\
\cline{2-5}
& Interpretation-Nets & Training data & Interpretable (decision trees) &
\citet{martonExplainingNeuralNetworks2024} \\
\end{longtable}

\section{Privacy Preserving XAI
Characteristics}\label{section6}

In the preceding sections, we have examined the privacy risks in XAI arising from both
intentional and unintentional causes. We have also reviewed applicable
privacy preserving methods to safeguard the additional attack surface exposed by
explanations. In this section, drawing on the insights from investigation into RQ1 and RQ2, we aim to address RQ3 by identifying key
characteristics that XAI should possess to mitigate the identified risks. These
characteristics provide a framework for understanding the essential properties of privacy preserving XAI,
taking into account the vulnerable assets that require protection and the various stakeholders
involved during the AI lifecycle. The proposed characteristics offer guidelines to
both researchers and practitioners to assess the effectiveness of existing privacy preserving XAI methods and guide the development of new approaches that prioritize privacy by design. By incorporating these qualities, XAI can strive to achieve the optimal balance between the triad of privacy, explainability and
utility.

\begin{figure}[h]
\begin{center}
\includegraphics[width=0.9\linewidth]{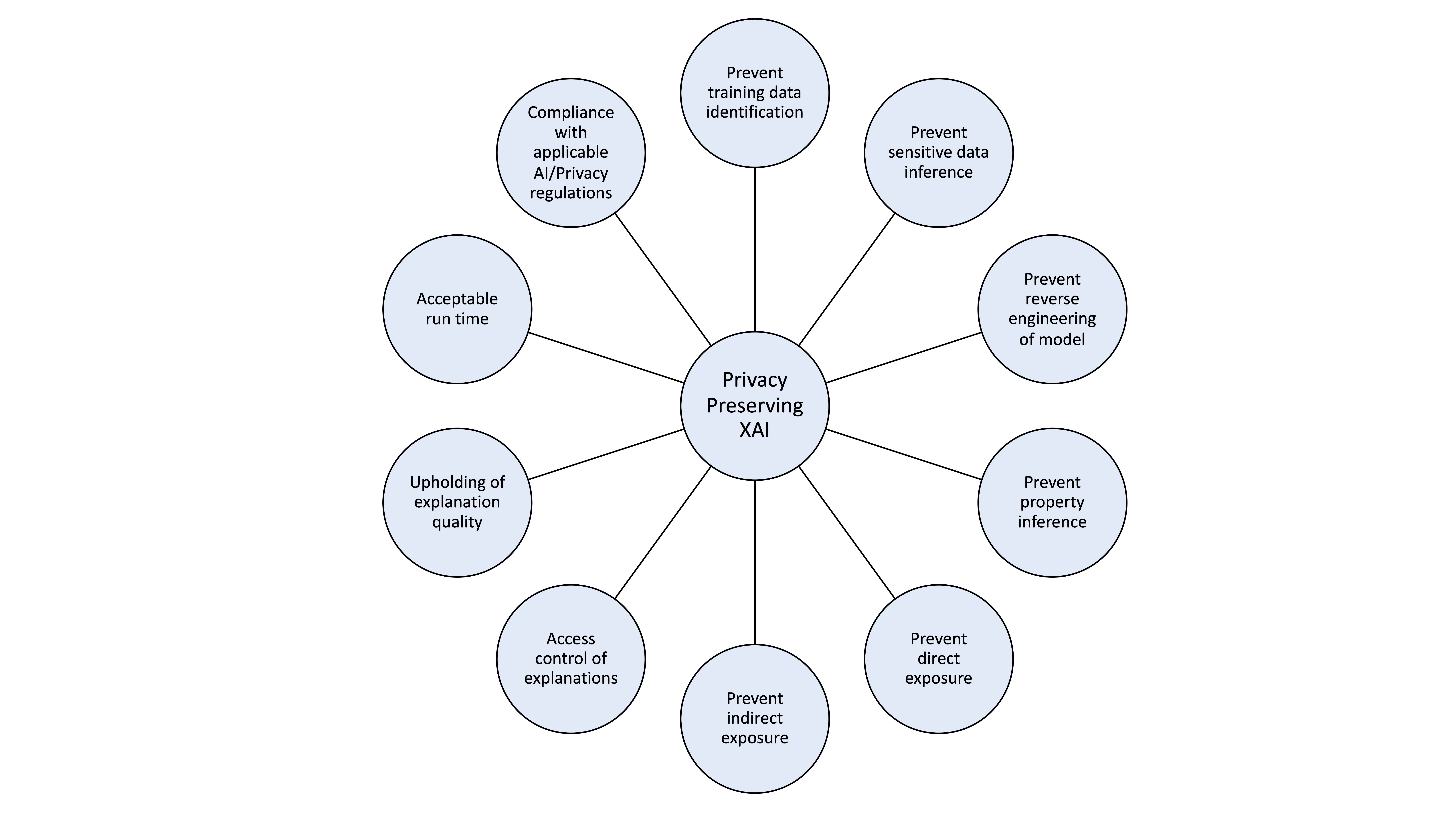}
\end{center}
\caption{Proposed privacy preserving XAI characteristics.}
\label{fig:fig11}
\end{figure}

We present the characteristics (Figure~\ref{fig:fig11}) by considering three use
cases outlined in Table~\ref{tab:table4}. To facilitate understanding, a simplified
example of an online loan application system that leverages an AI model with XAI capabilities is considered. The
system uses seven input features where salary, net worth, and age, are
protected features that require privacy preservation. The use cases
describe the following scenarios:

\begin{itemize}
\item
  Use Case 1 considers intentional privacy leakage through an
  adversary.
\item
  Use Case 2 involves interaction of a layman end-user, i.e., a bank's
  customer, with the XAI system. The end-user is provided an explanation of an automated decision
  directly through the system and indirectly through a human. Let's
  assume that in this use case, the loan was denied because the
  applicant salary was \textless{} 40K and age was \textgreater{} 50
  years.
\item
  Use Case 3 considers the interaction of technical support, i.e., AI developer and quality engineer, with the XAI system.
\end{itemize}

\textbf{Table 4} Use cases for privacy preserving XAI in an online loan
application system.

\begin{longtable}[]{@{}
  >{\raggedright\arraybackslash}p{(\linewidth - 2\tabcolsep) * \real{0.25}}
  >{\raggedright\arraybackslash}p{(\linewidth - 2\tabcolsep) * \real{0.75}}@{}}
\toprule\noalign{}
\label{tab:table4}
\begin{minipage}[b]{\linewidth}\centering
\textbf{Property}
\end{minipage} & \begin{minipage}[b]{\linewidth}\centering
\textbf{Details}
\end{minipage} \\
\midrule\noalign{}
\endhead
\bottomrule\noalign{}
\endlastfoot
System & Online loan application system \\
\cline{2-2}
Model owner & Bank \\
\cline{2-2}
Model input features & salary, net worth, age, length of credit history,
occupation, working hours per week, education \\
\cline{2-2}
Sensitive features & salary, net worth, age \\
\hline
Use Case 1 & Adversary with black-box access to the system. \\
\cline{2-2}
Actors & Adversary \\
\cline{2-2}
Overview & An adversary secures black-box access to the bank's model
through the online application system. The adversary attempts different
queries and observes the outputs generated by the system. \\
\cline{2-2}
Query data & (i) randomly generated queries.

(ii) targeted queries using prior information. \\
\hline
Use Case 2 & Customer accessing explanation of the application
outcome. \\
\cline{2-2}
Actors & Customer, bank executive \\
\cline{2-2}
Overview & A customer submits an online application for a loan and is
given a denied result. The customer is provided with:

(i) an automated explanation.

(ii) a consultation with a bank executive to discuss the result. \\
\cline{2-2}
Query data & salary = \$35K, net worth = \$75K, age = 55 years, length
of credit history = 30 years, occupation = office executive, working
hours per week = 25, education = diploma. \\
\hline
Use Case 3 & AI developer accessing explanation for troubleshooting a
reported error case and a quality engineer subsequently validating the
system updates. \\
\cline{2-2}
Actors & AI developer, quality engineer \\
\cline{2-2}
Overview & An error is reported on a specific query and a developer
updates the model during debugging. The developer accesses the
explanation of the error case to verify the results. Finally, a quality
engineer validates the system updates with another round of testing. \\
\cline{2-2}
Query data & Synthetic query similar to the error case requiring
troubleshooting. \\
\end{longtable}

We propose ten characteristics of privacy preserving XAI that aim to balance
privacy, explainability and utility. The first six characteristics are derived from
privacy attacks and unintentional leakage discussed in Section~\ref{section4}. The
remaining four characteristics are focussed on addressing performance issues and ensuring regulatory
compliance. The proposed characteristics are as follows:

\subsection{Prevent training data identification}

XAI tools should be designed such that the identification of individuals used in training
the model is not facilitated through them. In Use Case 1, if the adversary has access to a specific
individual's input details and retrieves the corresponding outputs
including the outcome and explanations, no additional advantage should
be provided through explanations in determining if the individual was
used in training the bank's model. Thus, the explanations should be
resilient to membership inference (Section~\ref{section4.2.1}).

\subsection{Prevent sensitive data inference}

XAI tools should be designed to prevent reconstruction or inference of sensitive
attributes of individuals. In Use Case 1, if the adversary has access to the
non-sensitive features of an individual and the outcome of the loan
application but is unaware of any sensitive feature such as salary or
age, the explanations provided should not aid in inferring these
sensitive features of the individual. Thus, the explanations should be
resilient to model inversion (Section~\ref{section4.2.2}).

\subsection{Prevent reverse engineering of model}

XAI tools should be designed to prevent reverse engineering of the model
functionalities. In Use Case 1, the adversary, by querying the bank's
model and by inspecting the explanations, should be unable to build a
surrogate of the original model. Thus, the explanations should be
resilient to model extraction (Section~\ref{section4.2.3}).

\subsection{Prevent property inference}

XAI tools should be designed to prevent the inferencing of aggregate properties of the
training data. In Use Case 1, by using targeted queries on the bank's
model, the adversary should be unable to exploit explanations in
determining group properties such as the ratio of old and young
customers in the training data or the distribution of wealthy and average
income training participants. Thus, the explanations should be resilient
to property inference (Section~\ref{section4.2.4}).

\subsection{Prevent direct exposure}

Explanations generated by XAI tools should not disclose personally
identifiable and/or sensitive information to unauthorized individuals
\citep{chenAchievingTransparencyReport2022}. In
Use Case 2, when the customer seeks an explanation on his/her
application outcome, the explanation might indicate the failure to meet
respective thresholds of \$40K for salary and 50 years for age.
Revealing actual values of protected features would breach the
customer's privacy when accessed by other actors, such as the
bank executive during the customer's consultation. The customer may,
however, subsequently provide consent to the executive to retrieve their
personal and financial details from the bank's records for
consultation.

\subsection{Prevent indirect exposure}

The content of the provided explanations should not indirectly expose
personally identifiable and/or sensitive information through correlated
or proxy features to unauthorized individuals. In Use Case 2, if the
explanation discloses a non-sensitive attribute such as the length of credit history, to actors other than the customer, it could indirectly lead
to exposure of a sensitive attribute, such as age, due to the strong correlation between the two attributes.

\subsection{Access control of explanations}

The content of explanations should be accessible only to authorized users
\citep{blanco-justiciaMachineLearningExplainability2020,kuppaAdversarialXAIMethods2021} with details provided on need-to-know basis. In Use Case 2,
the customer is authorized to access his/her own explanation as it pertains to their specific application outcome. The bank
executive should be authorized to access the explanation only if a human
intermediary is required to enhance the process of explanation for the
customer. In Use Case 3, the AI developer and quality engineer should be
permitted to access explanations and outcomes only for synthetic queries
generated to simulate specific error cases rather than for
real production data.

\subsection{Upholding of explanation quality}

The quality of explanations should not be compromised by the introduction of
privacy preservation measures. Explanations must remain useful and meaningful
\citep{shokriPrivacyRisksModel2021} to target stakeholders. In Use Cases 2 and 3, the
details contained in the explanations to respective users should assist
them in completing their tasks effectively and/or help them interpret
the outcome of the AI system.

\subsection{Acceptable run time}

The run time of XAI methods, being an important evaluation metric
\citep{bodriaBenchmarkingSurveyExplanation2023}, should not deteriorate by introduction of privacy
preservation measures. In Use Cases 2 and 3, the explanation recipients
should see the outputs within an acceptable timeframe. A long turnaround
time may lead to the explanations becoming ineffective for the task at
hand.

\subsection{Compliance with applicable AI/privacy regulations}

XAI being an AI system, should comply with applicable AI and privacy
regulations specfic to the jurisdiction in which it operates. For instance, if the XAI is
deployed in Canada with Canadian residents as the target users, it must adhere to the provisions of the Artificial Intelligence and Data Act \citep{aidaActEnactConsumer2022}. Users should be clearly informed of the XAI capabilities of the system
including the types and content of explanations and the third parties with whom
the explanations may be shared. Furthermore, appropriate consent must be obtained from users, as required by applicable laws and regulations.

\section{Discussion}\label{section7}

In this section, we summarize the results of our work, open issues, and challenges in the field. Additionally, we offer 
recommendations for future research directions to advance the development and deployment of privacy preserving XAI systems.

\subsection{Summary and implications}\label{section7.1}
The comprehensive review of existing literature facilitated a synthesis of current knowledge on
the conflict between privacy and explainability, both being important
pillars of Trustworthy AI. The analyzed	 studies demonstrate that the
additional information provided in the form of explanations can benefit adversaries in launching privacy attacks in XAI. We identified and
categorized certain types of privacy leakage due to malicious intent of
adversaries as intentional causes. These include membership
inference, model inversion and model extraction, all of which are demonstrated on
explanations generated using different methods. These attacks pose a
threat to the privacy of individuals whose data is contained in the training set,
thus increasing the risk of identification of individuals or exposure of their sensitive
information. Moreover, reconstruction of data from inference-time queries renders active XAI users vulnerable to similar privacy breaches \citep{zhaoEfficientPrivacypreservingTreebased2023,zhaoExploitingExplanationsModel2021}. The
threat of model extraction through explanations, targets the
confidentiality and intellectual property of model owners. While property
inference can expose sensitive aggregates or group properties of the
training data, our review found no evidence of such attacks targeting XAI
systems specifically. Beyond privacy attacks, ML
models exhibit inherent privacy vulnerabilities, such as memorization of training
data or overfitting, which can lead to various inferencing attacks. These
privacy problems are inherited in XAI systems, and we have categorized
them as unintentional causes. Additionally, the explanation content can be at direct threat of
privacy breaches by unauthorized users due to lack of access control, or
through proxies and correlated fields. Such vectors further compound the privacy risks associated with deploying XAI technologies in sensitive domains.

Due to the growing concerns surrounding privacy risks of explainability, researchers have
proposed defense mechanisms for privacy preservation with
explanations. This review identifies that techniques, such as DP and
anonymization, are extensively explored in this context, as evidenced by the
number of studies that have employed these methods. However, there remains limited exploration of alternative approaches, including knowledge integration, cryptography, and perturbation, all of which present promising avenues for enhancing privacy preservation. Hence there is scope to utilize these
underexplored techniques to achieve privacy in XAI systems. In
distributed environments, Fed-XAI attempts to achieve explainability
while preserving privacy of local data, yet its current implementations are insufficient to guarantee privacy in the generated explanations. 

The investigation of privacy risks and preservation methods in XAI has led to the identification of key characteristics that privacy preserving explanations must exhibit. In addition to being resilient to privacy attacks and
preventing both direct and indirect exposure of sensitive information,
explanations should satisfy performance and utility constraints. Given that
explanations may contain potentially identifiable data and may be
subject to legal and regulatory frameworks, they are required to comply with the applicable
AI and privacy laws within the jurisdiction. This article identifies and highlights a gap in the research of methods within the 
field of XAI, i.e., explainability methods should be designed
considering privacy as a system requirement. The findings of this paper can be utilized by researchers to understand state-of-the-art privacy attacks and corresponding preservation methods. Practitioners can leverage these insights to enhance their understanding of the privacy risks associated with XAI and identify potential solutions to mitigate those risks across various XAI methods.

\subsection{Design considerations for privacy preserving XAI}\label{section7.2}
Many unintentional leaks identified in this review can be avoided through responsible design practices. Safeguards in data handling such as data minimisation, deidentification and anonymisation of training data, or the use of synthetic data for training models when sensitive or personally identifiable information is involved, can help minimize the risks. During training, regularisation techniques can prevent overfitting and memorisation. The use of private training algorithms, such as differentially private training, can provide a privacy guarantee against leakages. At the output interface, the explanation API can include role-based access of explanation content to restrict the availability of sensitive information to end-users on a need-to-know basis. The use of logging at output for monitoring, tracing and privacy audits can further aid the accountability of the system. 

Steps can also be taken to prevent intentional leaks by restricting explanation APIs to authorised users. Restraining unlimited query access and setting the number of queries depending on users' roles can prevent misuse of explanation APIs by adversaries. Query monitoring can further help detection of anomalies. In addition to securing the interfaces, model owners should give due consideration to the type of access provided to users. When the model is available through an API, black-box privacy attacks are possible. However, when the model as a whole is released to users, model owners have no control on its usage and white-box privacy attacks become possible, thus giving additional avenues to adversaries to penetrate the system. Developers can use documented XAI privacy attacks for testing of systems during development for evaluation of the risk of privacy breaches. These evaluations can assist in making decisions about integration of appropriate privacy preservation methods as needed.

\subsection{Open issues and challenges}\label{section7.3}

Based on the privacy risks and mitigation methods surveyed, key open issues and challenges have been identified that require further attention. These challenges underscore the complexity of balancing privacy with the need for explainability in AI systems. In particular, the following issues remain critical: 
\begin{itemize}
\item{\textbf{Lack of user-centricity in development of privacy preserving XAI:}}
End-users are an integral and inseparable component of XAI as they directly engage and draw insights from the content generated by these systems. While current XAI methods are predominantly model-centric, focussing on model development, evaluation and audit processes \citep{kaplanUnifiedPracticalUsercentric2024}, there is an increasing need for a shift towards a user-centric approach. When considering privacy as a system requirement or using existing privacy preserving solutions with explainability methods, human users are absent from the design and development of these approaches. 

\item{\textbf{Lack of standardised approach for privacy evaluation of XAI:}}
Existing XAI literature provides evaluation metrics to assess aspects such as faithfulness, complexity, localisation, randomisation and robustness \citep{hedstromQuantusExplainableAI2023}. However, though privacy is fundamental in Trustworthy AI, currently there is a lack of standardised approach to evaluate the privacy of explanations. The lack of suitable quantitative metrics in this area prevents an evaluaton of safety with respect to privacy or comparison between methods for selection of those that satisfy the safety requirements of an application domain.  

\item{\textbf{Trade-oﬀs in privacy, explainability and utility:}}
The introduction of privacy enhancing technologies often result in adverse eﬀects on model utility \citep{ezzeddineDifferentialPrivacyAnomaly2024, harderInterpretableDifferentiallyPrivate2020} and the quality of explanations. For instance, the perturbation of classifier weights of support vector machines for privacy preservation is observed to deteriorate the classification accuracy and credibility of counterfactuals \citep{mochaourabDemonstratorCounterfactualExplanations2023}. Similarly, the use of diﬀerential privacy in neural network models is found to lower its accuracy \citep{blanco-justiciaCriticalReviewUse2023} and explanation quality \citep{liuPleaseTellMe2024}. Thus when targeting these 3 properties simultaneously in an XAI system, there are often trade-offs involved. 

\item{\textbf{Lack of XAI methods that are privacy preserving by design:}}
As emphasized by \citet{hoepmanPrivacyDesignStrategies2014}, privacy is a core property of computer systems that requires addressing from system design phase, rather than treated as an add-on. However, current explainability approaches do not consider privacy as a system requirement. Due to the  overlook of this safety aspect, design flaws may be introduced which can lead to unintentional privacy leakage. Adversaries may exploit these flaws to cause intentional leakage.

\item{\textbf{Lack of privacy preserving XAI for Gen-AI and LLMs:}}
XAI research has mainly focused on discriminative models that produce decision boundaries, and there has been limited work on developing explainability methods for Gen-AI and LLMs \citep{schneiderExplainableGenerativeAI2024,sunInvestigatingExplainabilityGenerative2022, weiszGeneralDesignPrinciples2023}. Due to the complex structure and vast number of parameters in these models, traditional explainability methods become impractical to them \citep{zhaoExplainabilityLargeLanguage2024}. These models have privacy issues, such as memorization of training data that escalates as the models become larger \citep{carliniExtractingTrainingData2021}. In addition, downstream private datasets used for in-context learning in LLMs, are found to be susceptible to membership inference \citep{duanPrivacyRiskIncontext2023}. When creating explainability approaches for these systems, privacy should be considered as an integral requirement.
\end{itemize}

\subsection{Recommendations for future work}\label{section7.4}
Based on the open issues and challenges outlined in the previous subsection, we have the following recommendations for future research in this area:

\begin{itemize}
\item{\textbf{Improving usability of privacy preserving XAI:}}
Explanations should be designed to meet the diverse information needs of users and integrate user-centric design principles \citep{aliExplainableArtificialIntelligence2023,kaplanUnifiedPracticalUsercentric2024}. The delivery of explanations is required to be in a format that is accessible and meaningful, taking into consideration the varying levels of expertise and requirements of diﬀerent user groups. This transition would prioritize providing need-to-know information to end-users based on their specific roles within the system. Furthermore, to enhance the eﬀectiveness of explanations, appropriate tools such as interactivity and visualization should be used to enhance the process of explanation and deepen users’ understanding \citep{boIncrementalXAIMemorable2024}. In addition, application of the 3C-principle of context, content, and consent \citep{brunotteContextContentConsent2023} can improve the usability of XAI tools by satisfying their requirements and expectations.\item{\textbf{Development of a standardised approach of privacy evaluation of XAI methods:}}
The development of a standardised approach to measure the leakage of privacy through explanations, will benefit developers in gauging the privacy of specific explanations so that appropriate preservation techniques can be integrated as needed. The development of quantitative privacy metrics can facilitate comparison between methods for adoption of those that are privacy safe and suitable for use in specific domains depending on the regulation requirement or the risk involved.  
\item{\textbf{Balancing trade-oﬀ in privacy, explainability and utility:}}
Determining the appropriate trade-oﬀ between the triad of privacy, explainability and utility, can help to achieve the right measure of balance between these properties. This could be achieved with the help of tools or tuning mechanisms. For instance, the use of compatibility matrix \citep{abbasiFurtherInsightsBalancing2024} or hyperparameters, such as the privacy budget $\epsilon$ in diﬀerential privacy or the parameter \textit
{k} in k-anonymity, is useful in tuning the desired level of these required properties. Similar tuning mechanisms can be integrated in other privacy preserving approaches to achieve the required trade-oﬀ. Metrics, such as trade-oﬀ score \citep{abbasiFurtherInsightsBalancing2024}, could be useful to quantify and monitor the balance of these properties enabling researchers and practitioners to adjust the parameters based on the trade-oﬀs involved.
\item{\textbf{Examining and improving trade-oﬀ in diﬀerent privacy preserving methods for XAI:}}
Diﬀerent privacy preservation methods applicable to XAI are discussed in Section ~\ref{section5}. Analysing the privacy-explainability-utility trade-oﬀ of these methods will identify eﬀective solutions and their limitations. While techniques, such as diﬀerential privacy and anonymization, have been mainly explored in XAI systems, other underutilized techniques such as use of knowledge integration and cryptographic protocols, could provide alternative approaches. Distributed privacy enhancing solutions, such as Fed-XAI, warrant further investigations to determine strategies to mitigate possible privacy leakages from XAI components. By systematically examining and comparing these various privacy preserving techniques, researchers can identify best practices and design hybrid approaches to eﬀectively balance diﬀerent properties.
\item{\textbf{Development of XAI methods that are privacy preserving by design:}}
The characteristics of privacy preserving XAI outlined in this review, can aid researchers and developers in building explainability methods that are privacy-enhanced by design \citep{bozorgpanahExplainableMachineLearning2024}. Furthermore, there is growing interest in neuro-symbolic approaches \citep{hitzlerNeurosymbolicApproachesArtificial2022} and semantic technologies \citep{seeligerSemanticWebTechnologies2019} as potential solutions as explainable-by-design strategies. Researchers can investigate the privacy safety of these techniques and integrate approaches to prioritize privacy.
\item{\textbf{Privacy preserving XAI for Gen-AI and LLMs:}}
XAI plays a vital role in fostering trustworthiness \citep{wangRationalityExplanationHuman2024} and ensuring ethical applications of Gen-AI and LLMs \citep{luoUnderstandingUtilizationSurvey2024}. However, as explainability is introduced in these models, it is important to ensure that it does not exacerbate the inherent privacy issues of these models or create new vulnerabilities. A privacy analysis of explainability methods in the early stages of development and the use of privacy attacks for auditing \citep{carliniExtractingTrainingData2021}, can boost the development of privacy-enhanced systems. Recently, Chain-of-Thought (CoT) \citep{wei2022chain} prompting has gained traction in eliciting the step-by-step reasoning of models and providing self-explanations in the process. However, the interaction with untrusted LLMs can be a threat to users' privacy, hence future work can be directed towards privacy preserved mechanisms \citep{bae2025privacy} in these models. Methods such as retrieval-augmented generation (RAG) for fine tuning of outputs by augmenting external data sources \citep{zengGoodBadExploring2024} and mechanistic interpretability (Section~\ref{section2.3.8}) can also be investigated as possible solutions towards achieving privacy preservation XAI. Thus, with the growing accessibility and widespread use of Gen-AI and LLMs, developing appropriate user-centric privacy preserving explainability techniques is an important avenue for further research. 
\item{\textbf{Evaluation of privacy preserving XAI characteristics:}}
The characteristics of privacy preserving XAI that we propose, aims to highlight the desirable qualities that XAI should exhibit to protect privacy while producing useful explanations to the target users. In further work, we will evaluate current XAI methods in light of these proposed characteristics to determine gaps in the methods and inform strategies for improvement. We will also enhance current methods so that the generated explanations better align with the proposed characteristics. We aim to improve the applicability of the characteristics through the evaluation of XAI methods.
\end{itemize}

\section{Conclusion}\label{section8}

XAI is an active field of research and a crucial pillar of Trustworthy
AI. It aims to bring logical explanations, a fundamental property of all
computer systems, to black-box AI models. Explainability of models is
essential to secure user trust in automated outcomes, especially in
critical domains where such outcomes have high impact on the lives of
individuals. Though explainability has emerged as a gold standard for
Trustworthy AI, previous works have highlighted potential privacy risks
of introducing transparency to black boxes. To the best of our
knowledge, there is a lack of detailed review on the tension between
privacy and explainability. In this article, we have focused on this gap
and conducted a scoping review to elicit details on the privacy risks
posed by XAI and the corresponding solutions for privacy preservation in
XAI. Our review draws attention to the intentional and unintentional
misuse of explanation interfaces and the pressing need for developing
XAI that is privacy preserving. In addition to reviewing the privacy
risks and the progress achieved by researchers in achieving privacy
improvement in XAI systems, we propose the characteristics of privacy
preserving XAI, to assist AI engineers and researchers in understanding
the requirements of XAI that achieves privacy with utility. We base
these characteristics on the identified risks, the encountered
performance issues, and the expected regulatory compliance. The
characteristics can be utilized for designing new explainability methods
and for evaluation of existing methods. Finally, we conclude the article
by identifying the open issues and challenges in the field and provide
recommendations for future work. Among the directions identified,
developing privacy metrics, creating privacy preserving explanations for
generative models and balancing the trade-off of privacy, utility and
explainability in existing and new XAI methods, will determine its
success as a foundation pillar of Trustworthy AI.

\subsubsection*{Broader Impact Statement}
This research was conducted to determine if two pillars of Trustworthy AI, namely, privacy and explainability, can co-exist. The privacy risks highlighted here have been consolidated from current literature to create awareness when using explainable methods in bringing transparency to black-boxes. 

\bibliography{main}

@inproceedings{kapishnikovXRAIBetterAttributions2019,
	address = {Seoul, Korea (South)},
	title = {{XRAI}: {Better} {Attributions} {Through} {Regions}},
	isbn = {978-1-7281-4803-8},
	shorttitle = {{XRAI}},
	url = {https://ieeexplore.ieee.org/document/9008576/},
	doi = {10.1109/ICCV.2019.00505},
	language = {en},
	urldate = {2023-02-24},
	booktitle = {2019 {IEEE}/{CVF} {International} {Conference} on {Computer} {Vision} ({ICCV})},
	publisher = {IEEE},
	author = {Kapishnikov, Andrei and Bolukbasi, Tolga and Viegas, Fernanda and Terry, Michael},
	month = oct,
	year = {2019},
	pages = {4947--4956},
}

@book{westinPrivacyFreedom1967,
	address = {New York},
	title = {Privacy and {Freedom}},
	publisher = {Atheneum},
	author = {Westin, Alan F.},
	year = {1967},
}

@article{warrenRightPrivacy1890,
	title = {The {Right} to {Privacy}},
	volume = {IV},
	number = {5},
	journal = {Harvard Law Review},
	author = {Warren, Samuel D. and Brandeis, Louis D.},
	month = dec,
	year = {1890},
}

@misc{icoExplainingDecisionsMade2020,
	title = {Explaining decisions made with {AI}},
	url = {https://ico.org.uk/for-organisations/guide-to-data-protection/key-dp-themes/explaining-decisions-made-with-ai/},
	language = {en},
	publisher = {Information Commissioner's Office},
	author = {ICO},
	year = {2020},
}

@inproceedings{shokriPrivacyRisksModel2021,
	address = {Virtual Event USA},
	title = {On the {Privacy} {Risks} of {Model} {Explanations}},
	isbn = {978-1-4503-8473-5},
	url = {https://dl.acm.org/doi/10.1145/3461702.3462533},
	doi = {10.1145/3461702.3462533},
	language = {en},
	urldate = {2023-02-24},
	booktitle = {Proceedings of the 2021 {AAAI}/{ACM} {Conference} on {AI}, {Ethics}, and {Society}},
	publisher = {ACM},
	author = {Shokri, Reza and Strobel, Martin and Zick, Yair},
	month = jul,
	year = {2021},
	pages = {231--241},
}

@article{rigakiSurveyPrivacyAttacks2023,
	title = {A {Survey} of {Privacy} {Attacks} in {Machine} {Learning}},
	volume = {56},
	doi = {https://doi.org/10.1145/3624010},
	language = {en},
	number = {4},
	urldate = {2023-02-24},
	journal = {ACM Computing Surveys},
	author = {Rigaki, Maria and Garcia, Sebastian},
	month = nov,
	year = {2023},
	pages = {1--34},
}

@article{zhangSurveyPrivacyInference2023,
	title = {A survey on privacy inference attacks and defenses in cloud-based {Deep} {Neural} {Network}},
	volume = {83},
	issn = {09205489},
	url = {https://linkinghub.elsevier.com/retrieve/pii/S0920548922000435},
	doi = {10.1016/j.csi.2022.103672},
	language = {en},
	urldate = {2023-02-24},
	journal = {Computer Standards \& Interfaces},
	author = {Zhang, Xiaoyu and Chen, Chao and Xie, Yi and Chen, Xiaofeng and Zhang, Jun and Xiang, Yang},
	month = jan,
	year = {2023},
	pages = {103672},
}

@article{jegorovaSurveyLeakagePrivacy2022,
	title = {Survey: {Leakage} and {Privacy} at {Inference} {Time}},
	issn = {0162-8828, 2160-9292, 1939-3539},
	shorttitle = {Survey},
	url = {https://ieeexplore.ieee.org/document/9987657/},
	doi = {10.1109/TPAMI.2022.3229593},
	language = {en},
	urldate = {2023-02-24},
	journal = {IEEE Transactions on Pattern Analysis and Machine Intelligence},
	author = {Jegorova, Marija and Kaul, Chaitanya and Mayor, Charlie and O'Neil, Alison Q. and Weir, Alexander and Murray-Smith, Roderick and Tsaftaris, Sotirios A.},
	year = {2022},
	pages = {1--20},
}

@misc{aivodjiModelExtractionCounterfactual2020,
	title = {Model extraction from counterfactual explanations},
	url = {http://arxiv.org/abs/2009.01884},
	language = {en},
	urldate = {2023-02-24},
	publisher = {arXiv},
	author = {Aïvodji, Ulrich and Bolot, Alexandre and Gambs, Sébastien},
	month = sep,
	year = {2020},
	note = {arXiv:2009.01884 [cs, stat]},
}

@inproceedings{milliModelReconstructionModel2019,
	address = {Atlanta GA USA},
	title = {Model {Reconstruction} from {Model} {Explanations}},
	isbn = {978-1-4503-6125-5},
	url = {https://dl.acm.org/doi/10.1145/3287560.3287562},
	doi = {10.1145/3287560.3287562},
	language = {en},
	urldate = {2023-02-24},
	booktitle = {Proceedings of the {Conference} on {Fairness}, {Accountability}, and {Transparency}},
	publisher = {ACM},
	author = {Milli, Smitha and Schmidt, Ludwig and Dragan, Anca D. and Hardt, Moritz},
	month = jan,
	year = {2019},
	pages = {1--9},
}

@inproceedings{wangDualCFEfficientModel2022,
	address = {Seoul Republic of Korea},
	title = {{DualCF}: {Efficient} {Model} {Extraction} {Attack} from {Counterfactual} {Explanations}},
	isbn = {978-1-4503-9352-2},
	shorttitle = {{DualCF}},
	url = {https://dl.acm.org/doi/10.1145/3531146.3533188},
	doi = {10.1145/3531146.3533188},
	language = {en},
	urldate = {2023-02-24},
	booktitle = {2022 {ACM} {Conference} on {Fairness}, {Accountability}, and {Transparency}},
	publisher = {ACM},
	author = {Wang, Yongjie and Qian, Hangwei and Miao, Chunyan},
	month = jun,
	year = {2022},
	pages = {1318--1329},
}

@article{vealeAlgorithmsThatRemember2018,
	title = {Algorithms that remember: model inversion attacks and data protection law},
	volume = {376},
	issn = {1364-503X, 1471-2962},
	shorttitle = {Governing artificial intelligence},
	url = {https://royalsocietypublishing.org/doi/10.1098/rsta.2018.0080},
	doi = {http://dx.doi.org/10.1098/rsta.2018.0083},
	language = {en},
	number = {2133},
	urldate = {2023-02-25},
	journal = {Philosophical Transactions of the Royal Society A: Mathematical, Physical and Engineering Sciences},
	author = {Veale, Michael and Binns, Reuben and Edwards, Lilian},
	month = nov,
	year = {2018},
	pages = {20180083},
}

@inproceedings{luoFeatureInferenceAttack2022,
	address = {Los Angeles CA USA},
	title = {Feature {Inference} {Attack} on {Shapley} {Values}},
	isbn = {978-1-4503-9450-5},
	url = {https://dl.acm.org/doi/10.1145/3548606.3560573},
	doi = {10.1145/3548606.3560573},
	language = {en},
	urldate = {2023-02-25},
	booktitle = {Proceedings of the 2022 {ACM} {SIGSAC} {Conference} on {Computer} and {Communications} {Security}},
	publisher = {ACM},
	author = {Luo, Xinjian and Jiang, Yangfan and Xiao, Xiaokui},
	month = nov,
	year = {2022},
	pages = {2233--2247},
}

@inproceedings{dudduInferringSensitiveAttributes2022,
	address = {New York, NY, USA},
	title = {Inferring {Sensitive} {Attributes} from {Model} {Explanations}},
	url = {https://dl.acm.org/doi/abs/10.1145/3511808.3557362},
	doi = {10.1145/3511808.3557362},
	language = {en},
	urldate = {2023-02-25},
	booktitle = {Proceedings of the 31st {ACM} {International} {Conference} on {Information} \& {Knowledge} {Management} ({CIKM} '22)},
	publisher = {Association for Computing Machinery},
	author = {Duddu, Vasisht and Boutet, Antoine},
	month = oct,
	year = {2022},
	pages = {416--425},
}

@inproceedings{yeomPrivacyRiskMachine2018,
	address = {Oxford},
	title = {Privacy {Risk} in {Machine} {Learning}: {Analyzing} the {Connection} to {Overfitting}},
	isbn = {978-1-5386-6680-7},
	shorttitle = {Privacy {Risk} in {Machine} {Learning}},
	url = {https://ieeexplore.ieee.org/document/8429311/},
	doi = {10.1109/CSF.2018.00027},
	language = {en},
	urldate = {2023-02-25},
	booktitle = {2018 {IEEE} 31st {Computer} {Security} {Foundations} {Symposium} ({CSF})},
	publisher = {IEEE},
	author = {Yeom, Samuel and Giacomelli, Irene and Fredrikson, Matt and Jha, Somesh},
	month = jul,
	year = {2018},
	pages = {268--282},
}

@inproceedings{mahloujifarPropertyInferencePoisoning2022,
	address = {San Francisco, CA, USA},
	title = {Property {Inference} from {Poisoning}},
	isbn = {978-1-6654-1316-9},
	url = {https://ieeexplore.ieee.org/document/9833623/},
	doi = {10.1109/SP46214.2022.9833623},
	language = {en},
	urldate = {2023-02-26},
	booktitle = {2022 {IEEE} {Symposium} on {Security} and {Privacy} ({SP})},
	publisher = {IEEE},
	author = {Mahloujifar, Saeed and Ghosh, Esha and Chase, Melissa},
	month = may,
	year = {2022},
	pages = {1120--1137},
}

@article{strobelDataPrivacyTrustworthy2022,
	title = {Data {Privacy} and {Trustworthy} {Machine} {Learning}},
	volume = {20},
	issn = {1540-7993, 1558-4046},
	url = {https://ieeexplore.ieee.org/document/9802763/},
	doi = {10.1109/MSEC.2022.3178187},
	language = {en},
	number = {5},
	urldate = {2023-02-27},
	journal = {IEEE Security \& Privacy},
	author = {Strobel, Martin and Shokri, Reza},
	month = sep,
	year = {2022},
	pages = {44--49},
}

@inproceedings{shokriMembershipInferenceAttacks2017,
	address = {San Jose, CA, USA},
	title = {Membership {Inference} {Attacks} {Against} {Machine} {Learning} {Models}},
	isbn = {978-1-5090-5533-3},
	url = {http://ieeexplore.ieee.org/document/7958568/},
	doi = {10.1109/SP.2017.41},
	language = {en},
	urldate = {2023-02-27},
	booktitle = {2017 {IEEE} {Symposium} on {Security} and {Privacy} ({SP})},
	publisher = {IEEE},
	author = {Shokri, Reza and Stronati, Marco and Song, Congzheng and Shmatikov, Vitaly},
	month = may,
	year = {2017},
	pages = {3--18},
}

@misc{doshi-velezRigorousScienceInterpretable2017,
	title = {Towards {A} {Rigorous} {Science} of {Interpretable} {Machine} {Learning}},
	url = {http://arxiv.org/abs/1702.08608},
	language = {en},
	urldate = {2023-03-01},
	publisher = {arXiv},
	author = {Doshi-Velez, Finale and Kim, Been},
	month = mar,
	year = {2017},
	note = {arXiv:1702.08608 [cs, stat]},
}

@inproceedings{hohmanGamutDesignProbe2019,
	address = {Glasgow Scotland Uk},
	title = {Gamut: {A} {Design} {Probe} to {Understand} {How} {Data} {Scientists} {Understand} {Machine} {Learning} {Models}},
	isbn = {978-1-4503-5970-2},
	shorttitle = {Gamut},
	url = {https://dl.acm.org/doi/10.1145/3290605.3300809},
	doi = {10.1145/3290605.3300809},
	language = {en},
	urldate = {2023-03-01},
	booktitle = {Proceedings of the 2019 {CHI} {Conference} on {Human} {Factors} in {Computing} {Systems}},
	publisher = {ACM},
	author = {Hohman, Fred and Head, Andrew and Caruana, Rich and DeLine, Robert and Drucker, Steven M.},
	month = may,
	year = {2019},
	pages = {1--13},
}

@article{machlevExplainableArtificialIntelligence2022,
	title = {Explainable {Artificial} {Intelligence} ({XAI}) techniques for energy and power systems: {Review}, challenges and opportunities},
	volume = {9},
	issn = {26665468},
	shorttitle = {Explainable {Artificial} {Intelligence} ({XAI}) techniques for energy and power systems},
	url = {https://linkinghub.elsevier.com/retrieve/pii/S2666546822000246},
	doi = {10.1016/j.egyai.2022.100169},
	language = {en},
	urldate = {2023-03-01},
	journal = {Energy and AI},
	author = {Machlev, R. and Heistrene, L. and Perl, M. and Levy, K.Y. and Belikov, J. and Mannor, S. and Levron, Y.},
	month = aug,
	year = {2022},
	pages = {100169},
}

@inproceedings{kohUnderstandingBlackboxPredictions2017,
	title = {Understanding {Black}-box {Predictions} via {Influence} {Functions}},
	volume = {70},
	url = {https://proceedings.mlr.press/v70/koh17a},
	language = {en},
	booktitle = {Proceedings of {Machine} {Learning} {Research}},
	author = {Koh, Pang Wei and Liang, Percy},
	month = jul,
	year = {2017},
	pages = {1885--1894},
}

@inproceedings{brunotteCanExplanationsSupport2021,
	address = {Notre Dame, IN, USA},
	title = {Can {Explanations} {Support} {Privacy} {Awareness}? {A} {Research} {Roadmap}},
	isbn = {978-1-6654-1898-0},
	shorttitle = {Can {Explanations} {Support} {Privacy} {Awareness}?},
	url = {https://ieeexplore.ieee.org/document/9582383/},
	doi = {10.1109/REW53955.2021.00032},
	language = {en},
	urldate = {2023-03-02},
	booktitle = {2021 {IEEE} 29th {International} {Requirements} {Engineering} {Conference} {Workshops} ({REW})},
	publisher = {IEEE},
	author = {Brunotte, Wasja and Chazette, Larissa and Korte, Kai},
	month = sep,
	year = {2021},
	pages = {176--180},
}

@techreport{tabassiAIRiskManagement2023,
	address = {Gaithersburg, MD},
	title = {{AI} {Risk} {Management} {Framework}: {AI} {RMF} (1.0)},
	shorttitle = {{AI} {Risk} {Management} {Framework}},
	url = {https://nvlpubs.nist.gov/nistpubs/ai/NIST.AI.100-1.pdf},
	language = {en},
	number = {NIST AI 100-1},
	urldate = {2023-03-02},
	institution = {National Institute of Standards and Technology},
	author = {Tabassi, Elham},
	year = {2023},
	doi = {10.6028/NIST.AI.100-1},
}

@inproceedings{ganjuPropertyInferenceAttacks2018,
	address = {Toronto Canada},
	title = {Property {Inference} {Attacks} on {Fully} {Connected} {Neural} {Networks} using {Permutation} {Invariant} {Representations}},
	isbn = {978-1-4503-5693-0},
	url = {https://dl.acm.org/doi/10.1145/3243734.3243834},
	doi = {10.1145/3243734.3243834},
	language = {en},
	urldate = {2023-03-03},
	booktitle = {Proceedings of the 2018 {ACM} {SIGSAC} {Conference} on {Computer} and {Communications} {Security}},
	publisher = {ACM},
	author = {Ganju, Karan and Wang, Qi and Yang, Wei and Gunter, Carl A. and Borisov, Nikita},
	month = oct,
	year = {2018},
	pages = {619--633},
}

@inproceedings{patelModelExplanationsDifferential2022,
	address = {Seoul Republic of Korea},
	title = {Model {Explanations} with {Differential} {Privacy}},
	isbn = {978-1-4503-9352-2},
	url = {https://dl.acm.org/doi/10.1145/3531146.3533235},
	doi = {10.1145/3531146.3533235},
	language = {en},
	urldate = {2023-03-06},
	booktitle = {2022 {ACM} {Conference} on {Fairness}, {Accountability}, and {Transparency}},
	publisher = {ACM},
	author = {Patel, Neel and Shokri, Reza and Zick, Yair},
	month = jun,
	year = {2022},
	pages = {1895--1904},
}

@article{harderInterpretableDifferentiallyPrivate2020,
	title = {Interpretable and {Differentially} {Private} {Predictions}},
	volume = {34},
	issn = {2374-3468, 2159-5399},
	url = {https://ojs.aaai.org/index.php/AAAI/article/view/5827},
	doi = {10.1609/aaai.v34i04.5827},
	language = {en},
	number = {04},
	urldate = {2023-03-07},
	journal = {Proceedings of the AAAI Conference on Artificial Intelligence},
	author = {Harder, Frederik and Bauer, Matthias and Park, Mijung},
	month = apr,
	year = {2020},
	pages = {4083--4090},
}

@inproceedings{dattaAlgorithmicTransparencyQuantitative2016,
	address = {San Jose, CA},
	title = {Algorithmic {Transparency} via {Quantitative} {Input} {Influence}: {Theory} and {Experiments} with {Learning} {Systems}},
	isbn = {978-1-5090-0824-7},
	shorttitle = {Algorithmic {Transparency} via {Quantitative} {Input} {Influence}},
	url = {http://ieeexplore.ieee.org/document/7546525/},
	doi = {10.1109/SP.2016.42},
	language = {en},
	urldate = {2023-03-07},
	booktitle = {2016 {IEEE} {Symposium} on {Security} and {Privacy} ({SP})},
	publisher = {IEEE},
	author = {Datta, Anupam and Sen, Shayak and Zick, Yair},
	month = may,
	year = {2016},
	pages = {598--617},
}

@article{rozanecKnowledgeGraphbasedRich2022,
	title = {Knowledge graph-based rich and confidentiality preserving {Explainable} {Artificial} {Intelligence} ({XAI})},
	volume = {81},
	issn = {15662535},
	url = {https://linkinghub.elsevier.com/retrieve/pii/S1566253521002414},
	doi = {10.1016/j.inffus.2021.11.015},
	language = {en},
	urldate = {2023-03-09},
	journal = {Information Fusion},
	author = {Rožanec, Jože M. and Fortuna, Blaž and Mladenić, Dunja},
	month = may,
	year = {2022},
	pages = {91--102},
}

@article{yanExplainableModelExtraction2022,
	title = {Towards explainable model extraction attacks},
	volume = {37},
	issn = {0884-8173, 1098-111X},
	url = {https://onlinelibrary.wiley.com/doi/10.1002/int.23022},
	doi = {10.1002/int.23022},
	language = {en},
	number = {11},
	urldate = {2023-03-17},
	journal = {International Journal of Intelligent Systems},
	author = {Yan, Anli and Hou, Ruitao and Liu, Xiaozhang and Yan, Hongyang and Huang, Teng and Wang, Xianmin},
	month = nov,
	year = {2022},
	pages = {9936--9956},
}

@inproceedings{zhaoExploitingExplanationsModel2021,
	address = {Montreal, QC, Canada},
	title = {Exploiting {Explanations} for {Model} {Inversion} {Attacks}},
	isbn = {978-1-6654-2812-5},
	url = {https://ieeexplore.ieee.org/document/9709977/},
	doi = {10.1109/ICCV48922.2021.00072},
	language = {en},
	urldate = {2023-03-17},
	booktitle = {2021 {IEEE}/{CVF} {International} {Conference} on {Computer} {Vision} ({ICCV})},
	publisher = {IEEE},
	author = {Zhao, Xuejun and Zhang, Wencan and Xiao, Xiaokui and Lim, Brian},
	month = oct,
	year = {2021},
	pages = {662--672},
}

@article{rawalRecentAdvancesTrustworthy2022,
	title = {Recent {Advances} in {Trustworthy} {Explainable} {Artificial} {Intelligence}: {Status}, {Challenges}, and {Perspectives}},
	volume = {3},
	issn = {2691-4581},
	shorttitle = {Recent {Advances} in {Trustworthy} {Explainable} {Artificial} {Intelligence}},
	url = {https://ieeexplore.ieee.org/document/9645355/},
	doi = {10.1109/TAI.2021.3133846},
	language = {en},
	number = {6},
	urldate = {2023-03-18},
	journal = {IEEE Transactions on Artificial Intelligence},
	author = {Rawal, Atul and McCoy, James and Rawat, Danda B. and Sadler, Brian M. and Amant, Robert St.},
	month = dec,
	year = {2022},
	pages = {852--866},
}

@article{gunningDARPAExplainableArtificial2019,
	title = {{DARPA}’s {Explainable} {Artificial} {Intelligence} {Program}},
	volume = {40},
	doi = {10.1609/aimag.v40i2.2850},
	language = {en},
	number = {2},
	journal = {AI Magazine},
	author = {Gunning, David and Aha, David W},
	month = jun,
	year = {2019},
	pages = {44--58},
}

@article{mcdermidArtificialIntelligenceExplainability2021,
	title = {Artificial intelligence explainability: the technical and ethical dimensions},
	volume = {379},
	issn = {1364-503X, 1471-2962},
	shorttitle = {Artificial intelligence explainability},
	url = {https://royalsocietypublishing.org/doi/10.1098/rsta.2020.0363},
	doi = {10.1098/rsta.2020.0363},
	language = {en},
	number = {2207},
	urldate = {2023-03-18},
	journal = {Philosophical Transactions of the Royal Society A: Mathematical, Physical and Engineering Sciences},
	author = {McDermid, John A. and Jia, Yan and Porter, Zoe and Habli, Ibrahim},
	month = oct,
	year = {2021},
	pages = {20200363},
}

@book{molnarInterpretableMachineLearning2023,
	title = {Interpretable {Machine} {Learning}},
	isbn = {979-8-4114-6333-0},
	author = {Molnar, Christoph},
	month = mar,
	year = {2023},
}

@inproceedings{biranHumanCentricJustificationMachine2017,
	address = {Melbourne, Australia},
	title = {Human-{Centric} {Justification} of {Machine} {Learning} {Predictions}},
	isbn = {978-0-9992411-0-3},
	url = {https://www.ijcai.org/proceedings/2017/202},
	doi = {10.24963/ijcai.2017/202},
	language = {en},
	urldate = {2023-03-18},
	booktitle = {Proceedings of the {Twenty}-{Sixth} {International} {Joint} {Conference} on {Artificial} {Intelligence}},
	publisher = {International Joint Conferences on Artificial Intelligence Organization},
	author = {Biran, Or and McKeown, Kathleen},
	month = aug,
	year = {2017},
	pages = {1461--1467},
}

@inproceedings{lundbergUnifiedApproachInterpreting2017,
	address = {Long Beach, California, USA},
	series = {{NIPS}'17},
	title = {A {Unified} {Approach} to {Interpreting} {Model} {Predictions}},
	isbn = {978-1-5108-6096-4},
	doi = {10.5555/3295222.3295230},
	language = {en},
	booktitle = {Proceedings of the 31st {International} {Conference} on {Neural} {Information} {Processing} {Systems}},
	publisher = {Curran Associates Inc.},
	author = {Lundberg, Scott M and Lee, Su-In},
	month = dec,
	year = {2017},
	pages = {4768--4777},
}

@article{liSurveyDatadrivenKnowledgeaware2020,
	title = {A {Survey} of {Data}-driven and {Knowledge}-aware {eXplainable} {AI}},
	issn = {1041-4347, 1558-2191, 2326-3865},
	url = {https://ieeexplore.ieee.org/document/9050829/},
	doi = {10.1109/TKDE.2020.2983930},
	language = {en},
	urldate = {2023-03-19},
	journal = {IEEE Transactions on Knowledge and Data Engineering},
	author = {Li, Xiao-Hui and Cao, Caleb Chen and Shi, Yuhan and Bai, Wei and Gao, Han and Qiu, Luyu and Wang, Cong and Gao, Yuanyuan and Zhang, Shenjia and Xue, Xun and Chen, Lei},
	year = {2020},
	pages = {1--1},
}

@article{dwivediExplainableAIXAI2023,
	title = {Explainable {AI} ({XAI}): {Core} {Ideas}, {Techniques}, and {Solutions}},
	volume = {55},
	issn = {0360-0300, 1557-7341},
	shorttitle = {Explainable {AI} ({XAI})},
	url = {https://dl.acm.org/doi/10.1145/3561048},
	doi = {10.1145/3561048},
	language = {en},
	number = {9},
	urldate = {2023-03-19},
	journal = {ACM Computing Surveys},
	author = {Dwivedi, Rudresh and Dave, Devam and Naik, Het and Singhal, Smiti and Omer, Rana and Patel, Pankesh and Qian, Bin and Wen, Zhenyu and Shah, Tejal and Morgan, Graham and Ranjan, Rajiv},
	month = sep,
	year = {2023},
	pages = {1--33},
}

@inproceedings{anconaBetterUnderstandingGradientbased2018,
	address = {Vancouver, BC, Canada},
	title = {Towards better understanding of gradient-based attribution methods for {Deep} {Neural} {Networks}},
	url = {https://openreview.net/forum?id=Sy21R9JAW},
	doi = {10.3929/ethz-b-000249929},
	language = {en},
	urldate = {2023-03-19},
	booktitle = {6th {International} {Conference} on {Learning} {Representations}, {ICLR} 2018},
	publisher = {OpenReview.net},
	author = {Ancona, Marco and Ceolini, Enea and Öztireli, Cengiz and Gross, Markus},
	month = apr,
	year = {2018},
	note = {arXiv:1711.06104 [cs, stat]},
}

@article{yangUnboxBlackboxMedical2022,
	title = {Unbox the black-box for the medical explainable {AI} via multi-modal and multi-centre data fusion: {A} mini-review, two showcases and beyond},
	volume = {77},
	issn = {15662535},
	shorttitle = {Unbox the black-box for the medical explainable {AI} via multi-modal and multi-centre data fusion},
	url = {https://linkinghub.elsevier.com/retrieve/pii/S1566253521001597},
	doi = {10.1016/j.inffus.2021.07.016},
	language = {en},
	urldate = {2023-03-19},
	journal = {Information Fusion},
	author = {Yang, Guang and Ye, Qinghao and Xia, Jun},
	month = jan,
	year = {2022},
	pages = {29--52},
}

@inproceedings{shrikumarLearningImportantFeatures2017,
	address = {Sydney, NSW, Australia},
	series = {{ICML}'17},
	title = {Learning {Important} {Features} {Through} {Propagating} {Activation} {Differences}},
	doi = {10.5555/3305890.3306006},
	language = {en},
	booktitle = {Proceedings of the 34th {International} {Conference} on {Machine} {Learning} - {Volume} 70},
	publisher = {JMLR.org},
	author = {Shrikumar, Avanti and Greenside, Peyton and Kundaje, Anshul},
	year = {2017},
	pages = {3145--3153},
}

@inproceedings{sundararajanAxiomaticAttributionDeep2017,
	address = {Sydney, NSW, Australia},
	series = {{ICML}'17},
	title = {Axiomatic {Attribution} for {Deep} {Networks}},
	doi = {10.5555/3305890.3306024},
	language = {en},
	booktitle = {Proceedings of the 34th {International} {Conference} on {Machine} {Learning} - {Volume} 70},
	publisher = {JMLR.org},
	author = {Sundararajan, Mukund and Taly, Ankur and Yan, Qiqi},
	year = {2017},
	pages = {3319--3328},
}

@inproceedings{simonyanDeepConvolutionalNetworks2014,
	title = {Deep {Inside} {Convolutional} {Networks}: {Visualising} {Image} {Classification} {Models} and {Saliency} {Maps}},
	shorttitle = {Deep {Inside} {Convolutional} {Networks}},
	language = {en},
	urldate = {2023-03-21},
	booktitle = {Workshop at {International} {Conference} on {Learning} {Representations}},
	author = {Simonyan, Karen and Vedaldi, Andrea and Zisserman, Andrew},
	month = apr,
	year = {2014},
	note = {arXiv:1312.6034 [cs]},
}

@inproceedings{sundararajanManyShapleyValues2020,
	address = {Virtual},
	title = {The {Many} {Shapley} {Values} for {Model} {Explanation}},
	language = {en},
	booktitle = {Proceedings of the 37th {International} {Conference} on {Machine} {Learning}},
	publisher = {PMLR},
	author = {Sundararajan, Mukund and Najmi, Amir},
	month = sep,
	year = {2020},
	pages = {9269--9278},
}

@article{guidottiCounterfactualExplanationsHow2022,
	title = {Counterfactual explanations and how to find them: literature review and benchmarking},
	issn = {1384-5810, 1573-756X},
	shorttitle = {Counterfactual explanations and how to find them},
	url = {https://link.springer.com/10.1007/s10618-022-00831-6},
	doi = {10.1007/s10618-022-00831-6},
	language = {en},
	urldate = {2023-03-22},
	journal = {Data Mining and Knowledge Discovery},
	author = {Guidotti, Riccardo},
	month = apr,
	year = {2022},
}

@inproceedings{kimExamplesAreNot2016,
	series = {{NIPS} 2016},
	title = {Examples are not enough, learn to criticize! {Criticism} for {Interpretability}},
	volume = {29},
	isbn = {978-1-5108-3881-9},
	language = {en},
	booktitle = {Advances in {Neural} {Information} {Processing} {Systems}},
	author = {Kim, Been and Khanna, Rajiv and Koyejo, Oluwasanmi},
	year = {2016},
}

@article{wachterCounterfactualExplanationsOpening2017,
	title = {Counterfactual {Explanations} {Without} {Opening} the {Black} {Box}: {Automated} {Decisions} and the {GDPR}},
	issn = {1556-5068},
	shorttitle = {Counterfactual {Explanations} {Without} {Opening} the {Black} {Box}},
	url = {https://www.ssrn.com/abstract=3063289},
	doi = {10.2139/ssrn.3063289},
	language = {en},
	urldate = {2023-03-22},
	journal = {SSRN Electronic Journal},
	author = {Wachter, Sandra and Mittelstadt, Brent and Russell, Chris},
	year = {2017},
}

@article{ribeiroAnchorsHighPrecisionModelAgnostic2018,
	title = {Anchors: {High}-{Precision} {Model}-{Agnostic} {Explanations}},
	volume = {32},
	issn = {2374-3468, 2159-5399},
	shorttitle = {Anchors},
	url = {https://ojs.aaai.org/index.php/AAAI/article/view/11491},
	doi = {10.1609/aaai.v32i1.11491},
	language = {en},
	number = {1},
	urldate = {2023-03-22},
	journal = {Proceedings of the AAAI Conference on Artificial Intelligence},
	author = {Ribeiro, Marco Tulio and Singh, Sameer and Guestrin, Carlos},
	month = apr,
	year = {2018},
}

@inproceedings{ribeiroWhyShouldTrust2016,
	address = {San Francisco California USA},
	title = {"{Why} {Should} {I} {Trust} {You}?": {Explaining} the {Predictions} of {Any} {Classifier}},
	copyright = {https://www.acm.org/publications/policies/copyright\_policy\#Background},
	shorttitle = {"{Why} {Should} {I} {Trust} {You}?},
	url = {https://dl.acm.org/doi/10.1145/2939672.2939778},
	doi = {10.1145/2939672.2939778},
	language = {en},
	urldate = {2025-07-09},
	booktitle = {Proceedings of the 22nd {ACM} {SIGKDD} {International} {Conference} on {Knowledge} {Discovery} and {Data} {Mining}},
	publisher = {ACM},
	author = {Ribeiro, Marco Tulio and Singh, Sameer and Guestrin, Carlos},
	month = aug,
	year = {2016},
	pages = {1135--1144},
}

@article{jiaPredictionWeaningMechanical2021,
	title = {Prediction of weaning from mechanical ventilation using {Convolutional} {Neural} {Networks}},
	volume = {117},
	issn = {09333657},
	url = {https://linkinghub.elsevier.com/retrieve/pii/S0933365721000804},
	doi = {10.1016/j.artmed.2021.102087},
	language = {en},
	urldate = {2023-03-22},
	journal = {Artificial Intelligence in Medicine},
	author = {Jia, Yan and Kaul, Chaitanya and Lawton, Tom and Murray-Smith, Roderick and Habli, Ibrahim},
	month = jul,
	year = {2021},
	pages = {102087},
}

@article{jimenez-lunaDrugDiscoveryExplainable2020,
	title = {Drug discovery with explainable artificial intelligence},
	volume = {2},
	issn = {2522-5839},
	url = {https://www.nature.com/articles/s42256-020-00236-4},
	doi = {10.1038/s42256-020-00236-4},
	language = {en},
	number = {10},
	urldate = {2023-03-23},
	journal = {Nature Machine Intelligence},
	author = {Jiménez-Luna, José and Grisoni, Francesca and Schneider, Gisbert},
	month = oct,
	year = {2020},
	pages = {573--584},
}

@inproceedings{dhurandharExplanationsBasedMissing2018,
	address = {Montreal, QC, Canada},
	series = {{NIPS}'18},
	title = {Explanations based on the {Missing}: {Towards} {Contrastive} {Explanations} with {Pertinent} {Negatives}},
	doi = {10.5555/3326943.3326998},
	language = {en},
	publisher = {Curran Associates Inc.},
	author = {Dhurandhar, Amit and Chen, Pin-Yu and Luss, Ronny and Tu, Chun-Chen and Ting, Paishun and Shanmugam, Karthikeyan and Das, Payel},
	year = {2018},
	pages = {590--601},
}

@article{tiddiKnowledgeGraphsTools2022,
	title = {Knowledge graphs as tools for explainable machine learning: {A} survey},
	volume = {302},
	issn = {00043702},
	shorttitle = {Knowledge graphs as tools for explainable machine learning},
	url = {https://linkinghub.elsevier.com/retrieve/pii/S0004370221001788},
	doi = {10.1016/j.artint.2021.103627},
	language = {en},
	urldate = {2023-03-23},
	journal = {Artificial Intelligence},
	author = {Tiddi, Ilaria and Schlobach, Stefan},
	month = jan,
	year = {2022},
	pages = {103627},
}

@article{hitzlerNeurosymbolicApproachesArtificial2022,
	title = {Neuro-symbolic approaches in artificial intelligence},
	volume = {9},
	issn = {2095-5138, 2053-714X},
	url = {https://academic.oup.com/nsr/article/doi/10.1093/nsr/nwac035/6542460},
	doi = {10.1093/nsr/nwac035},
	language = {en},
	number = {6},
	urldate = {2023-03-24},
	journal = {National Science Review},
	author = {Hitzler, Pascal and Eberhart, Aaron and Ebrahimi, Monireh and Sarker, Md Kamruzzaman and Zhou, Lu},
	month = jun,
	year = {2022},
	pages = {nwac035},
}

@article{paezPragmaticTurnExplainable2019,
	title = {The {Pragmatic} {Turn} in {Explainable} {Artificial} {Intelligence} ({XAI})},
	volume = {29},
	issn = {0924-6495, 1572-8641},
	url = {http://link.springer.com/10.1007/s11023-019-09502-w},
	doi = {10.1007/s11023-019-09502-w},
	language = {en},
	number = {3},
	urldate = {2023-03-24},
	journal = {Minds and Machines},
	author = {Páez, Andrés},
	month = sep,
	year = {2019},
	pages = {441--459},
}

@article{rajabiKnowledgegraphbasedExplainableAI2022,
	title = {Knowledge-graph-based explainable {AI}: {A} systematic review},
	issn = {0165-5515, 1741-6485},
	shorttitle = {Knowledge-graph-based explainable {AI}},
	url = {http://journals.sagepub.com/doi/10.1177/01655515221112844},
	doi = {10.1177/01655515221112844},
	language = {en},
	urldate = {2023-03-24},
	journal = {Journal of Information Science},
	author = {Rajabi, Enayat and Etminani, Kobra},
	month = sep,
	year = {2022},
	pages = {016555152211128},
}

@article{lecueRoleKnowledgeGraphs2020,
	title = {On the role of knowledge graphs in explainable {AI}},
	volume = {11},
	doi = {10.3233/SW-190374},
	number = {1},
	journal = {Semantic Web},
	author = {Lecue, Freddy},
	year = {2020},
	pages = {41--51},
}

@article{seeligerSemanticWebTechnologies2019,
	title = {Semantic {Web} {Technologies} for {Explainable} {Machine} {Learning} {Models}: {A} {Literature} {Review}},
	volume = {2465},
	url = {https://api.semanticscholar.org/CorpusID:204832199},
	language = {en},
	journal = {PROFILES/SEMEX@ ISWC},
	author = {Seeliger, Arne and Pfaﬀ, Matthias and Krcmar, Helmut},
	year = {2019},
	pages = {1--16},
}

@inproceedings{ilkouSymbolicVsSubsymbolic2020,
	title = {Symbolic {Vs} {Sub}-symbolic {AI} {Methods}: {Friends} or {Enemies}?},
	volume = {2699},
	language = {en},
	urldate = {2023-03-25},
	booktitle = {Proceedings of the {CIKM} 2020 {Workshops}},
	author = {Ilkou, Eleni and Koutraki, Maria},
	month = oct,
	year = {2020},
}

@inproceedings{cravenExtractingTreeStructuredRepresentations1995,
	address = {Cambridge, MA, USA},
	title = {Extracting {Tree}-{Structured} {Representations} of {Trained} {Networks}},
	language = {en},
	booktitle = {In {Proceedings} of the 8th {International} {Conference} on {Neural} {Information} {Processing} {Systems} ({NIPS}'95)},
	publisher = {MIT Press},
	author = {Craven, Mark and Shavlik, Jude W},
	month = nov,
	year = {1995},
	pages = {24--30},
}

@inproceedings{springenbergStrivingSimplicityAll2015,
	address = {San Diego, CA, USA},
	title = {Striving for {Simplicity}: {The} {All} {Convolutional} {Net}},
	shorttitle = {Striving for {Simplicity}},
	language = {en},
	urldate = {2023-03-28},
	publisher = {ICLR},
	author = {Springenberg, Jost Tobias and Dosovitskiy, Alexey and Brox, Thomas and Riedmiller, Martin},
	year = {2015},
	note = {arXiv:1412.6806 [cs]},
}

@misc{shrikumarNotJustBlack2017,
	title = {Not {Just} a {Black} {Box}: {Learning} {Important} {Features} {Through} {Propagating} {Activation} {Differences}},
	shorttitle = {Not {Just} a {Black} {Box}},
	url = {http://arxiv.org/abs/1605.01713},
	language = {en},
	urldate = {2023-03-28},
	publisher = {arXiv},
	author = {Shrikumar, Avanti and Greenside, Peyton and Shcherbina, Anna and Kundaje, Anshul},
	month = apr,
	year = {2017},
	note = {arXiv:1605.01713 [cs]},
}

@article{blanco-justiciaMachineLearningExplainability2020,
	title = {Machine learning explainability via microaggregation and shallow decision trees},
	volume = {194},
	issn = {09507051},
	url = {https://linkinghub.elsevier.com/retrieve/pii/S0950705120300368},
	doi = {10.1016/j.knosys.2020.105532},
	language = {en},
	urldate = {2023-03-29},
	journal = {Knowledge-Based Systems},
	author = {Blanco-Justicia, Alberto and Domingo-Ferrer, Josep and Martínez, Sergio and Sánchez, David},
	month = apr,
	year = {2020},
	pages = {105532},
}

@incollection{veugenPrivacyPreservingContrastiveExplanations2022,
	address = {Cham},
	title = {Privacy-{Preserving} {Contrastive} {Explanations} with {Local} {Foil} {Trees}},
	volume = {13301},
	isbn = {978-3-031-07688-6 978-3-031-07689-3},
	url = {https://link.springer.com/10.1007/978-3-031-07689-3_7},
	language = {en},
	urldate = {2023-03-29},
	booktitle = {Cyber {Security}, {Cryptology}, and {Machine} {Learning}},
	publisher = {Springer International Publishing},
	author = {Veugen, Thijs and Kamphorst, Bart and Marcus, Michiel},
	editor = {Dolev, Shlomi and Katz, Jonathan and Meisels, Amnon},
	year = {2022},
	doi = {10.1007/978-3-031-07689-3_7},
	note = {Series Title: Lecture Notes in Computer Science},
	pages = {88--98},
}

@article{montenegroPrivacyPreservingGenerativeAdversarial2021,
	title = {Privacy-{Preserving} {Generative} {Adversarial} {Network} for {Case}-{Based} {Explainability} in {Medical} {Image} {Analysis}},
	volume = {9},
	issn = {2169-3536},
	url = {https://ieeexplore.ieee.org/document/9598877/},
	doi = {10.1109/ACCESS.2021.3124844},
	language = {en},
	urldate = {2023-04-03},
	journal = {IEEE Access},
	author = {Montenegro, Helena and Silva, Wilson and Cardoso, Jaime S.},
	year = {2021},
	pages = {148037--148047},
}

@incollection{fiosinaInterpretablePrivacyPreservingCollaborative2022,
	address = {Cham},
	title = {Interpretable {Privacy}-{Preserving} {Collaborative} {Deep} {Learning} for {Taxi} {Trip} {Duration} {Forecasting}},
	volume = {1612},
	isbn = {978-3-031-17097-3 978-3-031-17098-0},
	url = {https://link.springer.com/10.1007/978-3-031-17098-0_20},
	language = {en},
	urldate = {2023-04-05},
	booktitle = {Smart {Cities}, {Green} {Technologies}, and {Intelligent} {Transport} {Systems}},
	publisher = {Springer International Publishing},
	author = {Fiosina, Jelena},
	editor = {Klein, Cornel and Jarke, Matthias and Helfert, Markus and Berns, Karsten and Gusikhin, Oleg},
	year = {2022},
	doi = {10.1007/978-3-031-17098-0_20},
	note = {Series Title: Communications in Computer and Information Science},
	pages = {392--411},
}

@article{liuWhenMachineLearning2022,
	title = {When {Machine} {Learning} {Meets} {Privacy}: {A} {Survey} and {Outlook}},
	volume = {54},
	issn = {0360-0300, 1557-7341},
	shorttitle = {When {Machine} {Learning} {Meets} {Privacy}},
	url = {https://dl.acm.org/doi/10.1145/3436755},
	doi = {10.1145/3436755},
	language = {en},
	number = {2},
	urldate = {2023-04-06},
	journal = {ACM Computing Surveys},
	author = {Liu, Bo and Ding, Ming and Shaham, Sina and Rahayu, Wenny and Farokhi, Farhad and Lin, Zihuai},
	month = mar,
	year = {2022},
	pages = {1--36},
}

@inproceedings{dowlinCryptoNetsApplyingNeural2016,
	address = {New York City, NY, USA},
	title = {{CryptoNets}: {Applying} {Neural} {Networks} to {Encrypted} {Data} with {High} {Throughput} and {Accuracy}},
	language = {en},
	publisher = {PMLR},
	author = {Dowlin, Nathan and Gilad-Bachrach, Ran and Laine, Kim and Lauter, Kristin and Naehrig, Michael and Wernsing, John},
	year = {2016},
	pages = {201--210},
}

@article{merhiAssessmentBarriersImpacting2022,
	title = {An {Assessment} of the {Barriers} {Impacting} {Responsible} {Artificial} {Intelligence}},
	issn = {1387-3326, 1572-9419},
	url = {https://link.springer.com/10.1007/s10796-022-10276-3},
	doi = {10.1007/s10796-022-10276-3},
	language = {en},
	urldate = {2023-04-11},
	journal = {Information Systems Frontiers},
	author = {Merhi, Mohammad I.},
	month = apr,
	year = {2022},
}

@article{trocinResponsibleAIDigital2021,
	title = {Responsible {AI} for {Digital} {Health}: a {Synthesis} and a {Research} {Agenda}},
	issn = {1387-3326, 1572-9419},
	shorttitle = {Responsible {AI} for {Digital} {Health}},
	url = {https://link.springer.com/10.1007/s10796-021-10146-4},
	doi = {10.1007/s10796-021-10146-4},
	language = {en},
	urldate = {2023-04-11},
	journal = {Information Systems Frontiers},
	author = {Trocin, Cristina and Mikalef, Patrick and Papamitsiou, Zacharoula and Conboy, Kieran},
	month = jun,
	year = {2021},
}

@inproceedings{shokriExploitingTransparencyMeasures2020,
	address = {New York, USA},
	title = {Exploiting {Transparency} {Measures} for {Membership} {Inference}: a {Cautionary} {Tale}},
	volume = {13},
	language = {en},
	booktitle = {The {AAAI} {Workshop} on {Privacy}-{Preserving} {Artificial} {Intelligence} ({PPAI})},
	publisher = {AAAI},
	author = {Shokri, Reza and Strobel, Martin and Zick, Yair},
	year = {2020},
}

@inproceedings{songMachineLearningModels2017,
	address = {Dallas Texas USA},
	title = {Machine {Learning} {Models} that {Remember} {Too} {Much}},
	isbn = {978-1-4503-4946-8},
	url = {https://dl.acm.org/doi/10.1145/3133956.3134077},
	doi = {10.1145/3133956.3134077},
	language = {en},
	urldate = {2023-04-11},
	booktitle = {Proceedings of the 2017 {ACM} {SIGSAC} {Conference} on {Computer} and {Communications} {Security}},
	publisher = {ACM},
	author = {Song, Congzheng and Ristenpart, Thomas and Shmatikov, Vitaly},
	month = oct,
	year = {2017},
	pages = {587--601},
}

@article{hitzlerNeuralsymbolicIntegrationSemantic2020,
	title = {Neural-symbolic integration and the {Semantic} {Web}},
	volume = {11},
	issn = {22104968, 15700844},
	url = {https://www.medra.org/servlet/aliasResolver?alias=iospress&doi=10.3233/SW-190368},
	doi = {10.3233/SW-190368},
	language = {en},
	number = {1},
	urldate = {2023-04-12},
	journal = {Semantic Web},
	author = {Hitzler, Pascal and Bianchi, Federico and Ebrahimi, Monireh and Sarker, Md Kamruzzaman},
	editor = {Janowicz, Krzysztof},
	month = jan,
	year = {2020},
	pages = {3--11},
}

@article{breimanRandomForests2001,
	title = {Random forests},
	volume = {45},
	journal = {Machine learning},
	author = {Breiman, Leo},
	year = {2001},
	pages = {5--32},
}

@article{mckaybowenPhilosophyDifferentialPrivacy2021,
	title = {The {Philosophy} of {Differential} {Privacy}},
	volume = {68},
	issn = {0002-9920, 1088-9477},
	url = {https://www.ams.org/notices/202110/rnoti-p1727.pdf},
	doi = {10.1090/noti2363},
	language = {en},
	number = {10},
	urldate = {2023-04-24},
	journal = {Notices of the American Mathematical Society},
	author = {McKay Bowen, Claire and Garfinkel, Simson},
	month = nov,
	year = {2021},
	pages = {1},
}

@incollection{dworkOurDataOurselves2006,
	address = {Berlin, Heidelberg},
	title = {Our {Data}, {Ourselves}: {Privacy} {Via} {Distributed} {Noise} {Generation}},
	volume = {4004},
	isbn = {978-3-540-34546-6 978-3-540-34547-3},
	shorttitle = {Our {Data}, {Ourselves}},
	url = {http://link.springer.com/10.1007/11761679_29},
	language = {en},
	urldate = {2023-04-24},
	booktitle = {Advances in {Cryptology} - {EUROCRYPT} 2006},
	publisher = {Springer Berlin Heidelberg},
	author = {Dwork, Cynthia and Kenthapadi, Krishnaram and McSherry, Frank and Mironov, Ilya and Naor, Moni},
	editor = {Vaudenay, Serge},
	year = {2006},
	doi = {10.1007/11761679_29},
	note = {Series Title: Lecture Notes in Computer Science},
	pages = {486--503},
}

@article{payrovnaziriExplainableArtificialIntelligence2020,
	title = {Explainable artificial intelligence models using real-world electronic health record data: a systematic scoping review},
	volume = {27},
	issn = {1527-974X},
	shorttitle = {Explainable artificial intelligence models using real-world electronic health record data},
	url = {https://academic.oup.com/jamia/article/27/7/1173/5838471},
	doi = {10.1093/jamia/ocaa053},
	language = {en},
	number = {7},
	urldate = {2023-04-28},
	journal = {Journal of the American Medical Informatics Association},
	author = {Payrovnaziri, Seyedeh Neelufar and Chen, Zhaoyi and Rengifo-Moreno, Pablo and Miller, Tim and Bian, Jiang and Chen, Jonathan H and Liu, Xiuwen and He, Zhe},
	month = jul,
	year = {2020},
	pages = {1173--1185},
}

@article{mohseniMultidisciplinarySurveyFramework2021,
	title = {A {Multidisciplinary} {Survey} and {Framework} for {Design} and {Evaluation} of {Explainable} {AI} {Systems}},
	volume = {11},
	issn = {2160-6455, 2160-6463},
	url = {https://dl.acm.org/doi/10.1145/3387166},
	doi = {10.1145/3387166},
	language = {en},
	number = {3-4},
	urldate = {2023-04-28},
	journal = {ACM Transactions on Interactive Intelligent Systems},
	author = {Mohseni, Sina and Zarei, Niloofar and Ragan, Eric D.},
	month = dec,
	year = {2021},
	pages = {1--45},
}

@article{capuanoExplainableArtificialIntelligence2022,
	title = {Explainable {Artificial} {Intelligence} in {CyberSecurity}: {A} {Survey}},
	volume = {10},
	issn = {2169-3536},
	shorttitle = {Explainable {Artificial} {Intelligence} in {CyberSecurity}},
	url = {https://ieeexplore.ieee.org/document/9877919/},
	doi = {10.1109/ACCESS.2022.3204171},
	language = {en},
	urldate = {2023-04-28},
	journal = {IEEE Access},
	author = {Capuano, Nicola and Fenza, Giuseppe and Loia, Vincenzo and Stanzione, Claudio},
	year = {2022},
	pages = {93575--93600},
}

@article{nassarBlockchainExplainableTrustworthy2020,
	title = {Blockchain for explainable and trustworthy artificial intelligence},
	volume = {10},
	issn = {1942-4787, 1942-4795},
	url = {https://onlinelibrary.wiley.com/doi/10.1002/widm.1340},
	doi = {10.1002/widm.1340},
	language = {en},
	number = {1},
	urldate = {2023-04-28},
	journal = {WIREs Data Mining and Knowledge Discovery},
	author = {Nassar, Mohamed and Salah, Khaled and ur Rehman, Muhammad Habib and Svetinovic, Davor},
	month = jan,
	year = {2020},
}

@inproceedings{konecnyFederatedLearningStrategies2016,
	address = {Barcelona, Spain},
	title = {Federated {Learning}: {Strategies} for {Improving} {Communication} {Efficiency}},
	shorttitle = {Federated {Learning}},
	url = {http://arxiv.org/abs/1610.05492},
	language = {en},
	urldate = {2023-04-28},
	author = {Konečný, Jakub and McMahan, H. Brendan and Yu, Felix X. and Richtárik, Peter and Suresh, Ananda Theertha and Bacon, Dave},
	year = {2016},
	note = {arXiv:1610.05492 [cs]},
}

@inproceedings{wangMeasureContributionParticipants2019,
	address = {Los Angeles, CA, USA},
	title = {Measure {Contribution} of {Participants} in {Federated} {Learning}},
	isbn = {978-1-7281-0858-2},
	url = {https://ieeexplore.ieee.org/document/9006179/},
	doi = {10.1109/BigData47090.2019.9006179},
	language = {en},
	urldate = {2023-04-29},
	booktitle = {2019 {IEEE} {International} {Conference} on {Big} {Data} ({Big} {Data})},
	publisher = {IEEE},
	author = {Wang, Guan and Dang, Charlie Xiaoqian and Zhou, Ziye},
	month = dec,
	year = {2019},
	pages = {2597--2604},
}

@article{clarkeInternetPrivacyConcerns1999,
	title = {Internet privacy concerns confirm the case for intervention},
	volume = {42},
	issn = {0001-0782, 1557-7317},
	url = {https://dl.acm.org/doi/10.1145/293411.293475},
	doi = {10.1145/293411.293475},
	language = {en},
	number = {2},
	urldate = {2023-05-03},
	journal = {Communications of the ACM},
	author = {Clarke, Roger},
	month = feb,
	year = {1999},
	pages = {60--67},
}

@article{curzonPrivacyArtificialIntelligence2021,
	title = {Privacy and {Artificial} {Intelligence}},
	volume = {2},
	issn = {2691-4581},
	url = {https://ieeexplore.ieee.org/document/9450036/},
	doi = {10.1109/TAI.2021.3088084},
	language = {en},
	number = {2},
	urldate = {2023-05-03},
	journal = {IEEE Transactions on Artificial Intelligence},
	author = {Curzon, James and Kosa, Tracy Ann and Akalu, Rajen and El-Khatib, Khalil},
	month = apr,
	year = {2021},
	pages = {96--108},
}

@article{deeksJudicialDemandExplainable2019,
	title = {The {Judicial} {Demand} {For} {Explainable} {Artificial} {Intelligence}},
	volume = {119},
	url = {https://www.jstor.org/stable/26810851},
	language = {en},
	number = {7},
	journal = {Columbia Law Review},
	author = {Deeks, Ashley},
	year = {2019},
	pages = {1829--1850},
}

@article{winikoffArtificialIntelligenceRight2021,
	title = {Artificial {Intelligence} and the {Right} to {Explanation} as a {Human} {Right}},
	volume = {25},
	issn = {1089-7801, 1941-0131},
	url = {https://ieeexplore.ieee.org/document/9420081/},
	doi = {10.1109/MIC.2020.3045821},
	language = {en},
	number = {2},
	urldate = {2023-05-08},
	journal = {IEEE Internet Computing},
	author = {Winikoff, Michael and Sardelic, Julija},
	month = mar,
	year = {2021},
	pages = {116--120},
}

@inproceedings{barcenaFedXAIFederatedLearning2022,
	address = {Italy},
	title = {Fed-{XAI}: {Federated} {Learning} of {Explainable} {Artificial} {Intelligence} {Models}},
	language = {en},
	booktitle = {{CEUR} {Workshop} {Proceedings} ({CEUR}-{WS}.org)},
	author = {Bárcena, José Luis Corcuera and Daole, Mattia and Ducange, Pietro and Marcelloni, Francesco and Renda, Alessandro and Ruffini, Fabrizio and Schiavo, Alessio},
	month = nov,
	year = {2022},
	pages = {104--117},
}

@article{chenEVFLExplainableVertical2022,
	title = {{EVFL}: {An} explainable vertical federated learning for data-oriented {Artificial} {Intelligence} systems},
	volume = {126},
	issn = {13837621},
	shorttitle = {{EVFL}},
	url = {https://linkinghub.elsevier.com/retrieve/pii/S1383762122000583},
	doi = {10.1016/j.sysarc.2022.102474},
	language = {en},
	urldate = {2023-05-11},
	journal = {Journal of Systems Architecture},
	author = {Chen, Peng and Du, Xin and Lu, Zhihui and Wu, Jie and Hung, Patrick C.K.},
	month = may,
	year = {2022},
	pages = {102474},
}

@article{mothukuriSurveySecurityPrivacy2021,
	title = {A survey on security and privacy of federated learning},
	volume = {115},
	issn = {0167739X},
	url = {https://linkinghub.elsevier.com/retrieve/pii/S0167739X20329848},
	doi = {10.1016/j.future.2020.10.007},
	language = {en},
	urldate = {2023-05-11},
	journal = {Future Generation Computer Systems},
	author = {Mothukuri, Viraaji and Parizi, Reza M. and Pouriyeh, Seyedamin and Huang, Yan and Dehghantanha, Ali and Srivastava, Gautam},
	month = feb,
	year = {2021},
	pages = {619--640},
}

@inproceedings{nasrComprehensivePrivacyAnalysis2019,
	address = {San Francisco, CA, USA},
	title = {Comprehensive {Privacy} {Analysis} of {Deep} {Learning}: {Passive} and {Active} {White}-box {Inference} {Attacks} against {Centralized} and {Federated} {Learning}},
	isbn = {978-1-5386-6660-9},
	shorttitle = {Comprehensive {Privacy} {Analysis} of {Deep} {Learning}},
	url = {https://ieeexplore.ieee.org/document/8835245/},
	doi = {10.1109/SP.2019.00065},
	language = {en},
	urldate = {2023-05-17},
	booktitle = {2019 {IEEE} {Symposium} on {Security} and {Privacy} ({SP})},
	publisher = {IEEE},
	author = {Nasr, Milad and Shokri, Reza and Houmansadr, Amir},
	month = may,
	year = {2019},
	pages = {739--753},
}

@misc{aidaActEnactConsumer2022,
	title = {An {Act} to enact the {Consumer} {Privacy} {Protection} {Act}, the {Personal} {Information} and {Data} {Protection} {Tribunal} {Act} and the {Artificial} {Intelligence} and {Data} {Act} and to make consequential and related amendments to other {Acts}},
	shorttitle = {Digital {Charter} {Implementation} {Act}},
	url = {https://www.justice.gc.ca/eng/csj-sjc/pl/charter-charte/c27_1.html},
	journal = {Bill C-27},
	author = {AIDA},
	month = nov,
	year = {2022},
}

@inproceedings{narettoEvaluatingPrivacyExposure2022,
	address = {Atlanta, GA, USA},
	title = {Evaluating the {Privacy} {Exposure} of {Interpretable} {Global} {Explainers}},
	doi = {10.1109/CogMI56440.2022.00012},
	publisher = {IEEE Computer Society},
	author = {Naretto, Francesca and Monreale, Anna and Giannotti, Fosca},
	year = {2022},
	pages = {13--19},
}

@article{gaudioDeepFixCXExplainablePrivacypreserving2023,
	title = {{DeepFixCX}: {Explainable} privacy‐preserving image compression for medical image analysis},
	issn = {1942-4787, 1942-4795},
	shorttitle = {{\textless}span style="font-variant},
	url = {https://onlinelibrary.wiley.com/doi/10.1002/widm.1495},
	doi = {10.1002/widm.1495},
	language = {en},
	urldate = {2023-05-27},
	journal = {WIREs Data Mining and Knowledge Discovery},
	author = {Gaudio, Alex and Smailagic, Asim and Faloutsos, Christos and Mohan, Shreshta and Johnson, Elvin and Liu, Yuhao and Costa, Pedro and Campilho, Aurélio},
	month = mar,
	year = {2023},
}

@article{yanExplanationLeaksExplanationguided2023,
	title = {Explanation leaks: {Explanation}-guided model extraction attacks},
	volume = {632},
	issn = {00200255},
	shorttitle = {Explanation leaks},
	url = {https://linkinghub.elsevier.com/retrieve/pii/S002002552300316X},
	doi = {10.1016/j.ins.2023.03.020},
	language = {en},
	urldate = {2023-05-29},
	journal = {Information Sciences},
	author = {Yan, Anli and Huang, Teng and Ke, Lishan and Liu, Xiaozhang and Chen, Qi and Dong, Changyu},
	month = jun,
	year = {2023},
	pages = {269--284},
}

@inproceedings{pawelczykPrivacyRisksAlgorithmic2023,
	address = {Valencia, Spain},
	series = {Proceedings of {Machine} {Learning} {Research}},
	title = {On the {Privacy} {Risks} of {Algorithmic} {Recourse}},
	volume = {206},
	url = {https://proceedings.mlr.press/v206/pawelczyk23a.html},
	language = {en},
	booktitle = {Proceedings of {The} 26th {International} {Conference} on {Artificial} {Intelligence} and {Statistics}},
	publisher = {PMLR},
	author = {Pawelczyk, Martin and Lakkaraju, Himabindu and Neel, Seth},
	year = {2023},
	pages = {9680--9696},
}

@incollection{muftuogluPrivacyPreservingMechanismsExplainability2022,
	address = {Cham},
	title = {Privacy-{Preserving} {Mechanisms} with {Explainability} in {Assistive} {AI} {Technologies}},
	volume = {28},
	isbn = {978-3-030-87131-4 978-3-030-87132-1},
	url = {https://link.springer.com/10.1007/978-3-030-87132-1_13},
	language = {en},
	urldate = {2023-05-31},
	booktitle = {Advances in {Assistive} {Technologies}},
	publisher = {Springer International Publishing},
	author = {Müftüoğlu, Z. and Kızrak, M. A. and Yıldırım, T.},
	editor = {Tsihrintzis, George A. and Virvou, Maria and Esposito, Anna and Jain, Lakhmi C.},
	year = {2022},
	doi = {10.1007/978-3-030-87132-1_13},
	note = {Series Title: Learning and Analytics in Intelligent Systems},
	pages = {287--309},
}

@inproceedings{fongInterpretableExplanationsBlack2017,
	address = {Venice},
	title = {Interpretable {Explanations} of {Black} {Boxes} by {Meaningful} {Perturbation}},
	isbn = {978-1-5386-1032-9},
	url = {http://ieeexplore.ieee.org/document/8237633/},
	doi = {10.1109/ICCV.2017.371},
	language = {en},
	urldate = {2023-05-31},
	booktitle = {2017 {IEEE} {International} {Conference} on {Computer} {Vision} ({ICCV})},
	publisher = {IEEE},
	author = {Fong, Ruth C. and Vedaldi, Andrea},
	month = oct,
	year = {2017},
	pages = {3449--3457},
}

@incollection{hoepmanPrivacyDesignStrategies2014,
	address = {Berlin, Heidelberg},
	title = {Privacy {Design} {Strategies}},
	volume = {428},
	isbn = {978-3-642-55414-8 978-3-642-55415-5},
	url = {http://link.springer.com/10.1007/978-3-642-55415-5_38},
	language = {en},
	urldate = {2023-05-31},
	booktitle = {{ICT} {Systems} {Security} and {Privacy} {Protection}},
	publisher = {Springer Berlin Heidelberg},
	author = {Hoepman, Jaap-Henk},
	editor = {Cuppens-Boulahia, Nora and Cuppens, Frédéric and Jajodia, Sushil and Abou El Kalam, Anas and Sans, Thierry},
	year = {2014},
	doi = {10.1007/978-3-642-55415-5_38},
	note = {Series Title: IFIP Advances in Information and Communication Technology},
	pages = {446--459},
}

@inproceedings{chazetteExploringExplainabilityDefinition2021,
	address = {Notre Dame, IN, USA},
	title = {Exploring {Explainability}: {A} {Definition}, a {Model}, and a {Knowledge} {Catalogue}},
	isbn = {978-1-6654-2856-9},
	shorttitle = {Exploring {Explainability}},
	url = {https://ieeexplore.ieee.org/document/9604587/},
	doi = {10.1109/RE51729.2021.00025},
	language = {en},
	urldate = {2023-06-15},
	booktitle = {2021 {IEEE} 29th {International} {Requirements} {Engineering} {Conference} ({RE})},
	publisher = {IEEE},
	author = {Chazette, Larissa and Brunotte, Wasja and Speith, Timo},
	month = sep,
	year = {2021},
	pages = {197--208},
}

@article{shahriarSurveyPrivacyRisks2023,
	title = {A {Survey} of {Privacy} {Risks} and {Mitigation} {Strategies} in the {Artificial} {Intelligence} {Life} {Cycle}},
	issn = {2169-3536},
	url = {https://ieeexplore.ieee.org/document/10155147/},
	doi = {10.1109/ACCESS.2023.3287195},
	language = {en},
	urldate = {2023-06-21},
	journal = {IEEE Access},
	author = {Shahriar, Sakib and Allana, Sonal and Fard, Mehdi Hazrati and Dara, Rozita},
	year = {2023},
	pages = {1--1},
}

@inproceedings{jeongLearningGenerateInversionResistant2022,
	address = {New Orleans},
	title = {Learning to {Generate} {Inversion}-{Resistant} {Model} {Explanations}},
	language = {en},
	author = {Jeong, Hoyong and Lee, Suyoung and Hwang, Sung Ju and Son, Sooel},
	year = {2022},
}

@article{chenAchievingTransparencyReport2022,
	title = {Achieving {Transparency} {Report} {Privacy} in {Linear} {Time}},
	volume = {14},
	issn = {1936-1955, 1936-1963},
	url = {https://dl.acm.org/doi/10.1145/3460001},
	doi = {10.1145/3460001},
	language = {en},
	number = {2},
	urldate = {2023-06-26},
	journal = {Journal of Data and Information Quality},
	author = {Chen, Chien-Lun and Golubchik, Leana and Pal, Ranjan},
	month = jun,
	year = {2022},
	pages = {1--56},
}

@inproceedings{fredriksonModelInversionAttacks2015,
	address = {Denver Colorado USA},
	title = {Model {Inversion} {Attacks} that {Exploit} {Confidence} {Information} and {Basic} {Countermeasures}},
	isbn = {978-1-4503-3832-5},
	url = {https://dl.acm.org/doi/10.1145/2810103.2813677},
	doi = {10.1145/2810103.2813677},
	language = {en},
	urldate = {2023-07-21},
	booktitle = {Proceedings of the 22nd {ACM} {SIGSAC} {Conference} on {Computer} and {Communications} {Security}},
	publisher = {ACM},
	author = {Fredrikson, Matt and Jha, Somesh and Ristenpart, Thomas},
	month = oct,
	year = {2015},
	pages = {1322--1333},
}

@inproceedings{sarkar2024explaining,
  title={Explaining LLM Decisions: Counterfactual Chain-of-Thought Approach},
  author={Sarkar, Pramir and Prakash, Ashish V and Singh, Jang Bahadur},
  booktitle={Advanced Computing and Communications Conference},
  pages={254--266},
  year={2024},
  organization={Springer}
}

@article{boedihardjoPrivacySyntheticData2023,
	title = {Privacy of {Synthetic} {Data}: {A} {Statistical} {Framework}},
	volume = {69},
	issn = {0018-9448, 1557-9654},
	shorttitle = {Privacy of {Synthetic} {Data}},
	url = {https://ieeexplore.ieee.org/document/9927488/},
	doi = {10.1109/TIT.2022.3216793},
	language = {en},
	number = {1},
	urldate = {2023-08-10},
	journal = {IEEE Transactions on Information Theory},
	author = {Boedihardjo, March and Strohmer, Thomas and Vershynin, Roman},
	month = jan,
	year = {2023},
	pages = {520--527},
}

@incollection{ullmanPCPsHardnessGenerating2011,
	address = {Berlin, Heidelberg},
	title = {{PCPs} and the {Hardness} of {Generating} {Private} {Synthetic} {Data}},
	volume = {6597},
	isbn = {978-3-642-19570-9 978-3-642-19571-6},
	url = {http://link.springer.com/10.1007/978-3-642-19571-6_24},
	language = {en},
	urldate = {2023-08-10},
	booktitle = {Theory of {Cryptography}},
	publisher = {Springer Berlin Heidelberg},
	author = {Ullman, Jonathan and Vadhan, Salil},
	editor = {Hutchison, David and Kanade, Takeo and Kittler, Josef and Kleinberg, Jon M. and Mattern, Friedemann and Mitchell, John C. and Naor, Moni and Nierstrasz, Oscar and Pandu Rangan, C. and Steffen, Bernhard and Sudan, Madhu and Terzopoulos, Demetri and Tygar, Doug and Vardi, Moshe Y. and Weikum, Gerhard and Ishai, Yuval},
	year = {2011},
	doi = {10.1007/978-3-642-19571-6_24},
	note = {Series Title: Lecture Notes in Computer Science},
	pages = {400--416},
}

@inproceedings{liuPrivacyPreservingSyntheticData2022,
	address = {Madrid Spain},
	title = {Privacy-{Preserving} {Synthetic} {Data} {Generation} for {Recommendation} {Systems}},
	isbn = {978-1-4503-8732-3},
	url = {https://dl.acm.org/doi/10.1145/3477495.3532044},
	doi = {10.1145/3477495.3532044},
	language = {en},
	urldate = {2023-08-10},
	booktitle = {Proceedings of the 45th {International} {ACM} {SIGIR} {Conference} on {Research} and {Development} in {Information} {Retrieval}},
	publisher = {ACM},
	author = {Liu, Fan and Cheng, Zhiyong and Chen, Huilin and Wei, Yinwei and Nie, Liqiang and Kankanhalli, Mohan},
	month = jul,
	year = {2022},
	pages = {1379--1389},
}

@inproceedings{weiszGeneralDesignPrinciples2023,
	title = {Toward {General} {Design} {Principles} for {Generative} {AI} {Applications}},
	language = {en},
	author = {Weisz, Justin D and Muller, Michael and He, Jessica and Houde, Stephanie},
	year = {2023},
}

@inproceedings{sunInvestigatingExplainabilityGenerative2022,
	address = {Helsinki Finland},
	title = {Investigating {Explainability} of {Generative} {AI} for {Code} through {Scenario}-based {Design}},
	isbn = {978-1-4503-9144-3},
	url = {https://dl.acm.org/doi/10.1145/3490099.3511119},
	doi = {10.1145/3490099.3511119},
	language = {en},
	urldate = {2023-08-10},
	booktitle = {27th {International} {Conference} on {Intelligent} {User} {Interfaces}},
	publisher = {ACM},
	author = {Sun, Jiao and Liao, Q. Vera and Muller, Michael and Agarwal, Mayank and Houde, Stephanie and Talamadupula, Kartik and Weisz, Justin D.},
	month = mar,
	year = {2022},
	pages = {212--228},
}

@article{meskoImperativeRegulatoryOversight2023,
	title = {The imperative for regulatory oversight of large language models (or generative {AI}) in healthcare},
	volume = {6},
	issn = {2398-6352},
	url = {https://www.nature.com/articles/s41746-023-00873-0},
	doi = {10.1038/s41746-023-00873-0},
	language = {en},
	number = {1},
	urldate = {2023-08-10},
	journal = {npj Digital Medicine},
	author = {Meskó, Bertalan and Topol, Eric J.},
	month = jul,
	year = {2023},
	pages = {120},
}

@article{arrietaExplainableArtificialIntelligence2020,
	title = {Explainable {Artificial} {Intelligence} ({XAI}): {Concepts}, {Taxonomies}, {Opportunities} and {Challenges} toward {Responsible} {AI}},
	volume = {58},
	issn = {1566-2535},
	shorttitle = {Explainable {Artificial} {Intelligence} ({XAI})},
	doi = {https://doi.org/10.1016/j.inffus.2019.12.012},
	language = {en},
	urldate = {2023-08-13},
	journal = {Information Fusion},
	author = {Arrieta, Alejandro Barredo and Díaz-Rodríguez, Natalia and Del Ser, Javier and Bennetot, Adrien and Tabik, Siham and Barbado, Alberto and García, Salvador and Gil-López, Sergio and Molina, Daniel and Benjamins, Richard and Chatila, Raja and Herrera, Francisco},
	year = {2020},
	pages = {82--115},
}

@article{blanco-justiciaCriticalReviewUse2023,
	title = {A {Critical} {Review} on the {Use} (and {Misuse}) of {Differential} {Privacy} in {Machine} {Learning}},
	volume = {55},
	issn = {0360-0300, 1557-7341},
	url = {https://dl.acm.org/doi/10.1145/3547139},
	doi = {10.1145/3547139},
	language = {en},
	number = {8},
	urldate = {2023-12-08},
	journal = {ACM Computing Surveys},
	author = {Blanco-Justicia, Alberto and Sánchez, David and Domingo-Ferrer, Josep and Muralidhar, Krishnamurty},
	month = aug,
	year = {2023},
	pages = {1--16},
}

@article{hedstromQuantusExplainableAI2023,
	title = {Quantus: {An} {Explainable} {AI} {Toolkit} for {Responsible} {Evaluation} of {Neural} {Network} {Explanations} and {Beyond}},
	volume = {24},
	language = {en},
	number = {34},
	journal = {Journal of Machine Learning Research},
	author = {Hedström, Anna and Weber, Leander and Bareeva, Dilyara and Krakowczyk, Daniel and Motzkus, Franz and Samek, Wojciech and Lapuschkin, Sebastian},
	year = {2023},
	pages = {1--11},
}

@article{huMembershipInferenceAttacks2022,
	title = {Membership {Inference} {Attacks} on {Machine} {Learning}: {A} {Survey}},
	volume = {54},
	issn = {0360-0300, 1557-7341},
	shorttitle = {Membership {Inference} {Attacks} on {Machine} {Learning}},
	url = {https://dl.acm.org/doi/10.1145/3523273},
	doi = {10.1145/3523273},
	language = {en},
	number = {11s},
	urldate = {2024-01-20},
	journal = {ACM Computing Surveys},
	author = {Hu, Hongsheng and Salcic, Zoran and Sun, Lichao and Dobbie, Gillian and Yu, Philip S. and Zhang, Xuyun},
	month = jan,
	year = {2022},
	pages = {1--37},
}

@article{zhaoExplainabilityLargeLanguage2024,
	title = {Explainability for {Large} {Language} {Models}: {A} {Survey}},
	issn = {2157-6904, 2157-6912},
	shorttitle = {Explainability for {Large} {Language} {Models}},
	url = {https://dl.acm.org/doi/10.1145/3639372},
	doi = {10.1145/3639372},
	language = {en},
	urldate = {2024-02-16},
	journal = {ACM Transactions on Intelligent Systems and Technology},
	author = {Zhao, Haiyan and Chen, Hanjie and Yang, Fan and Liu, Ninghao and Deng, Huiqi and Cai, Hengyi and Wang, Shuaiqiang and Yin, Dawei and Du, Mengnan},
	month = jan,
	year = {2024},
	pages = {3639372},
}

@inproceedings{bhattEvaluatingAggregatingFeaturebased2020,
	address = {Yokohama, Japan},
	title = {Evaluating and {Aggregating} {Feature}-based {Model} {Explanations}},
	isbn = {978-0-9992411-6-5},
	url = {https://www.ijcai.org/proceedings/2020/417},
	doi = {10.24963/ijcai.2020/417},
	language = {en},
	urldate = {2024-03-21},
	booktitle = {Proceedings of the {Twenty}-{Ninth} {International} {Joint} {Conference} on {Artificial} {Intelligence}},
	publisher = {International Joint Conferences on Artificial Intelligence Organization},
	author = {Bhatt, Umang and Weller, Adrian and Moura, José M. F.},
	month = jul,
	year = {2020},
	pages = {3016--3022},
}

@inproceedings{liuPleaseTellMe2024,
	address = {San Francisco, CA, USA},
	title = {Please {Tell} {Me} {More}: {Privacy} {Impact} of {Explainability} through the {Lens} of {Membership} {Inference} {Attack}},
	doi = {10.1109/SP54263.2024.00120},
	language = {en},
	author = {Liu, Han and Wu, Yuhao and Yu, Zhiyuan and Zhang, Ning},
	year = {2024},
	pages = {119--119},
}

@inproceedings{voFeaturebasedLearningDiverse2023,
	address = {Long Beach CA USA},
	title = {Feature-based {Learning} for {Diverse} and {Privacy}-{Preserving} {Counterfactual} {Explanations}},
	isbn = {979-8-4007-0103-0},
	url = {https://dl.acm.org/doi/10.1145/3580305.3599343},
	doi = {10.1145/3580305.3599343},
	language = {en},
	urldate = {2024-04-03},
	booktitle = {Proceedings of the 29th {ACM} {SIGKDD} {Conference} on {Knowledge} {Discovery} and {Data} {Mining}},
	publisher = {ACM},
	author = {Vo, Vy and Le, Trung and Nguyen, Van and Zhao, He and Bonilla, Edwin V. and Haffari, Gholamreza and Phung, Dinh},
	month = aug,
	year = {2023},
	pages = {2211--2222},
}

@article{bodriaBenchmarkingSurveyExplanation2023,
	title = {Benchmarking and survey of explanation methods for black box models},
	volume = {37},
	issn = {1384-5810, 1573-756X},
	url = {https://link.springer.com/10.1007/s10618-023-00933-9},
	doi = {10.1007/s10618-023-00933-9},
	language = {en},
	number = {5},
	urldate = {2024-04-04},
	journal = {Data Mining and Knowledge Discovery},
	author = {Bodria, Francesco and Giannotti, Fosca and Guidotti, Riccardo and Naretto, Francesca and Pedreschi, Dino and Rinzivillo, Salvatore},
	month = sep,
	year = {2023},
	pages = {1719--1778},
}

@article{yanExplanationbasedDatafreeModel2023,
	title = {Explanation-based data-free model extraction attacks},
	volume = {26},
	issn = {1386-145X, 1573-1413},
	url = {https://link.springer.com/10.1007/s11280-023-01150-6},
	doi = {10.1007/s11280-023-01150-6},
	language = {en},
	number = {5},
	urldate = {2024-04-06},
	journal = {World Wide Web},
	author = {Yan, Anli and Hou, Ruitao and Yan, Hongyang and Liu, Xiaozhang},
	month = sep,
	year = {2023},
	pages = {3081--3092},
}

@article{goethalsPrivacyIssueCounterfactual2023,
	title = {The {Privacy} {Issue} of {Counterfactual} {Explanations}: {Explanation} {Linkage} {Attacks}},
	volume = {14},
	issn = {2157-6904, 2157-6912},
	shorttitle = {The {Privacy} {Issue} of {Counterfactual} {Explanations}},
	url = {https://dl.acm.org/doi/10.1145/3608482},
	doi = {10.1145/3608482},
	language = {en},
	number = {5},
	urldate = {2024-04-07},
	journal = {ACM Transactions on Intelligent Systems and Technology},
	author = {Goethals, Sofie and Sörensen, Kenneth and Martens, David},
	month = oct,
	year = {2023},
	pages = {1--24},
}

@incollection{lopez-blancoFederatedLearningExplainable2023,
	address = {Cham},
	title = {Federated {Learning} of {Explainable} {Artificial} {Intelligence} ({FED}-{XAI}): {A} {Review}},
	volume = {740},
	isbn = {978-3-031-38332-8 978-3-031-38333-5},
	shorttitle = {Federated {Learning} of {Explainable} {Artificial} {Intelligence} ({FED}-{XAI})},
	url = {https://link.springer.com/10.1007/978-3-031-38333-5_32},
	language = {en},
	urldate = {2024-04-08},
	booktitle = {Distributed {Computing} and {Artificial} {Intelligence}, 20th {International} {Conference}},
	publisher = {Springer Nature Switzerland},
	author = {López-Blanco, Raúl and Alonso, Ricardo S. and González-Arrieta, Angélica and Chamoso, Pablo and Prieto, Javier},
	editor = {Ossowski, Sascha and Sitek, Pawel and Analide, Cesar and Marreiros, Goreti and Chamoso, Pablo and Rodríguez, Sara},
	year = {2023},
	doi = {10.1007/978-3-031-38333-5_32},
	note = {Series Title: Lecture Notes in Networks and Systems},
	pages = {318--326},
}

@article{wuFalconPrivacyPreservingInterpretable2023,
	title = {Falcon: {A} {Privacy}-{Preserving} and {Interpretable} {Vertical} {Federated} {Learning} {System}},
	volume = {16},
	issn = {2150-8097},
	shorttitle = {Falcon},
	url = {https://dl.acm.org/doi/10.14778/3603581.3603588},
	doi = {10.14778/3603581.3603588},
	language = {en},
	number = {10},
	urldate = {2024-04-09},
	journal = {Proceedings of the VLDB Endowment},
	author = {Wu, Yuncheng and Xing, Naili and Chen, Gang and Dinh, Tien Tuan Anh and Luo, Zhaojing and Ooi, Beng Chin and Xiao, Xiaokui and Zhang, Meihui},
	month = jun,
	year = {2023},
	pages = {2471--2484},
}

@article{zhuHorizontalFederatedLearning2022,
	title = {Horizontal {Federated} {Learning} of {Takagi}–{Sugeno} {Fuzzy} {Rule}-{Based} {Models}},
	volume = {30},
	copyright = {https://ieeexplore.ieee.org/Xplorehelp/downloads/license-information/IEEE.html},
	issn = {1063-6706, 1941-0034},
	url = {https://ieeexplore.ieee.org/document/9565342/},
	doi = {10.1109/TFUZZ.2021.3118733},
	language = {en},
	number = {9},
	urldate = {2024-04-10},
	journal = {IEEE Transactions on Fuzzy Systems},
	author = {Zhu, Xiubin and Wang, Dan and Pedrycz, Witold and Li, Zhiwu},
	month = sep,
	year = {2022},
	pages = {3537--3547},
}

@article{corcuerabarcenaEnablingFederatedLearning2023,
	title = {Enabling federated learning of explainable {AI} models within beyond-{5G}/{6G} networks},
	volume = {210},
	issn = {01403664},
	url = {https://linkinghub.elsevier.com/retrieve/pii/S0140366423002724},
	doi = {10.1016/j.comcom.2023.07.039},
	language = {en},
	urldate = {2024-04-10},
	journal = {Computer Communications},
	author = {Corcuera Bárcena, José Luis and Ducange, Pietro and Marcelloni, Francesco and Nardini, Giovanni and Noferi, Alessandro and Renda, Alessandro and Ruffini, Fabrizio and Schiavo, Alessio and Stea, Giovanni and Virdis, Antonio},
	month = oct,
	year = {2023},
	pages = {356--375},
}

@article{wuPrivacyPreservingVertical2020,
	title = {Privacy preserving vertical federated learning for tree-based models},
	volume = {13},
	issn = {2150-8097},
	url = {https://dl.acm.org/doi/10.14778/3407790.3407811},
	doi = {10.14778/3407790.3407811},
	language = {en},
	number = {12},
	urldate = {2024-04-10},
	journal = {Proceedings of the VLDB Endowment},
	author = {Wu, Yuncheng and Cai, Shaofeng and Xiao, Xiaokui and Chen, Gang and Ooi, Beng Chin},
	month = aug,
	year = {2020},
	pages = {2090--2103},
}

@inproceedings{jiaMemGuardDefendingBlackBox2019,
	address = {London United Kingdom},
	title = {{MemGuard}: {Defending} against {Black}-{Box} {Membership} {Inference} {Attacks} via {Adversarial} {Examples}},
	isbn = {978-1-4503-6747-9},
	shorttitle = {{MemGuard}},
	url = {https://dl.acm.org/doi/10.1145/3319535.3363201},
	doi = {10.1145/3319535.3363201},
	language = {en},
	urldate = {2024-04-10},
	booktitle = {Proceedings of the 2019 {ACM} {SIGSAC} {Conference} on {Computer} and {Communications} {Security}},
	publisher = {ACM},
	author = {Jia, Jinyuan and Salem, Ahmed and Backes, Michael and Zhang, Yang and Gong, Neil Zhenqiang},
	month = nov,
	year = {2019},
	pages = {259--274},
}

@article{kaplanUnifiedPracticalUsercentric2024,
	title = {A unified and practical user-centric framework for explainable artificial intelligence},
	volume = {283},
	issn = {09507051},
	url = {https://linkinghub.elsevier.com/retrieve/pii/S0950705123008572},
	doi = {10.1016/j.knosys.2023.111107},
	language = {en},
	urldate = {2024-04-11},
	journal = {Knowledge-Based Systems},
	author = {Kaplan, Sinan and Uusitalo, Hannu and Lensu, Lasse},
	month = jan,
	year = {2024},
	pages = {111107},
}

@misc{schneiderExplainableGenerativeAI2024,
	title = {Explainable {Generative} {AI} ({GenXAI}): {A} {Survey}, {Conceptualization}, and {Research} {Agenda}},
	shorttitle = {Explainable {Generative} {AI} ({GenXAI})},
	url = {http://arxiv.org/abs/2404.09554},
	language = {en},
	urldate = {2024-04-18},
	publisher = {arXiv},
	author = {Schneider, Johannes},
	month = apr,
	year = {2024},
	note = {arXiv:2404.09554 [cs]},
}

@misc{mishraPromptAidPromptExploration2023,
	title = {{PromptAid}: {Prompt} {Exploration}, {Perturbation}, {Testing} and {Iteration} using {Visual} {Analytics} for {Large} {Language} {Models}},
	shorttitle = {{PromptAid}},
	url = {http://arxiv.org/abs/2304.01964},
	language = {en},
	urldate = {2024-04-18},
	publisher = {arXiv},
	author = {Mishra, Aditi and Soni, Utkarsh and Arunkumar, Anjana and Huang, Jinbin and Kwon, Bum Chul and Bryan, Chris},
	month = apr,
	year = {2023},
	note = {arXiv:2304.01964 [cs]},
}

@inproceedings{duanPrivacyRiskIncontext2023,
	title = {On the {Privacy} {Risk} of {In}-context {Learning}},
	language = {en},
	booktitle = {61st {Annual} {Meeting} {Of} {The} {Association} {For} {Computational} {Linguistics}},
	author = {Duan, Haonan and Dziedzic, Adam and Yaghini, Mohammad and Papernot, Nicolas and Boenisch, Franziska},
	month = jul,
	year = {2023},
}

@inproceedings{carliniExtractingTrainingData2021,
	title = {Extracting {Training} {Data} from {Large} {Language} {Models}},
	isbn = {978-1-939133-24-3},
	url = {https://www.usenix.org/conference/usenixsecurity21/presentation/carlini-extracting},
	language = {en},
	booktitle = {30th {USENIX} {Security} {Symposium} ({USENIX} {Security} 21)},
	publisher = {USENIX Association},
	author = {Carlini, Nicholas and Tramèr, Florian and Wallace, Eric and Jagielski, Matthew and Herbert-Voss, Ariel and Lee, Katherine and Roberts, Adam and Brown, Tom and Song, Dawn and Erlingsson, Úlfar and Oprea, Alina and Raffel, Colin},
	month = aug,
	year = {2021},
	pages = {2633--2650},
}

@misc{luoUnderstandingUtilizationSurvey2024,
	title = {From {Understanding} to {Utilization}: {A} {Survey} on {Explainability} for {Large} {Language} {Models}},
	shorttitle = {From {Understanding} to {Utilization}},
	url = {http://arxiv.org/abs/2401.12874},
	language = {en},
	urldate = {2024-04-22},
	publisher = {arXiv},
	author = {Luo, Haoyan and Specia, Lucia},
	month = feb,
	year = {2024},
	note = {arXiv:2401.12874 [cs]},
}

@misc{zengGoodBadExploring2024,
	title = {The {Good} and {The} {Bad}: {Exploring} {Privacy} {Issues} in {Retrieval}-{Augmented} {Generation} ({RAG})},
	shorttitle = {The {Good} and {The} {Bad}},
	url = {http://arxiv.org/abs/2402.16893},
	language = {en},
	urldate = {2024-04-22},
	publisher = {arXiv},
	author = {Zeng, Shenglai and Zhang, Jiankun and He, Pengfei and Xing, Yue and Liu, Yiding and Xu, Han and Ren, Jie and Wang, Shuaiqiang and Yin, Dawei and Chang, Yi and Tang, Jiliang},
	month = feb,
	year = {2024},
	note = {arXiv:2402.16893 [cs]},
}

@article{alzubaidiRiskFreeTrustworthyArtificial2023,
	title = {Towards {Risk}-{Free} {Trustworthy} {Artificial} {Intelligence}: {Significance} and {Requirements}},
	volume = {2023},
	copyright = {https://creativecommons.org/licenses/by/4.0/},
	issn = {1098-111X, 0884-8173},
	shorttitle = {Towards {Risk}-{Free} {Trustworthy} {Artificial} {Intelligence}},
	url = {https://www.hindawi.com/journals/ijis/2023/4459198/},
	doi = {10.1155/2023/4459198},
	language = {en},
	urldate = {2024-04-30},
	journal = {International Journal of Intelligent Systems},
	author = {Alzubaidi, Laith and Al-Sabaawi, Aiman and Bai, Jinshuai and Dukhan, Ammar and Alkenani, Ahmed H. and Al-Asadi, Ahmed and Alwzwazy, Haider A. and Manoufali, Mohamed and Fadhel, Mohammed A. and Albahri, A. S. and Moreira, Catarina and Ouyang, Chun and Zhang, Jinglan and Santamaría, Jose and Salhi, Asma and Hollman, Freek and Gupta, Ashish and Duan, Ye and Rabczuk, Timon and Abbosh, Amin and Gu, Yuantong},
	editor = {El Kafhali, Said},
	month = oct,
	year = {2023},
	pages = {1--41},
}

@inproceedings{sandersonImplementingResponsibleAI2023,
	address = {Gold Coast, Australia},
	title = {Implementing {Responsible} {AI}: {Tensions} and {Trade}-{Offs} {Between} {Ethics} {Aspects}},
	copyright = {https://doi.org/10.15223/policy-029},
	isbn = {978-1-6654-8867-9},
	shorttitle = {Implementing {Responsible} {AI}},
	url = {https://ieeexplore.ieee.org/document/10191274/},
	doi = {10.1109/IJCNN54540.2023.10191274},
	language = {en},
	urldate = {2024-05-01},
	booktitle = {2023 {International} {Joint} {Conference} on {Neural} {Networks} ({IJCNN})},
	publisher = {IEEE},
	author = {Sanderson, Conrad and Douglas, David and Lu, Qinghua},
	month = jun,
	year = {2023},
	pages = {1--7},
}

@article{dharChallengesDeepLearning2023,
	title = {Challenges of {Deep} {Learning} in {Medical} {Image} {Analysis}—{Improving} {Explainability} and {Trust}},
	volume = {4},
	copyright = {https://ieeexplore.ieee.org/Xplorehelp/downloads/license-information/IEEE.html},
	issn = {2637-6415},
	url = {https://ieeexplore.ieee.org/document/10005626/},
	doi = {10.1109/TTS.2023.3234203},
	language = {en},
	number = {1},
	urldate = {2024-05-01},
	journal = {IEEE Transactions on Technology and Society},
	author = {Dhar, Tribikram and Dey, Nilanjan and Borra, Surekha and Sherratt, R. Simon},
	month = mar,
	year = {2023},
	pages = {68--75},
}

@incollection{muralidharElementsThatInfluence2023,
	address = {Cham},
	title = {Elements that {Influence} {Transparency} in {Artificial} {Intelligent} {Systems} - {A} {Survey}},
	volume = {14142},
	isbn = {978-3-031-42279-9 978-3-031-42280-5},
	url = {https://link.springer.com/10.1007/978-3-031-42280-5_21},
	language = {en},
	urldate = {2024-05-01},
	booktitle = {Human-{Computer} {Interaction} – {INTERACT} 2023},
	publisher = {Springer Nature Switzerland},
	author = {Muralidhar, Deepa and Belloum, Rafik and De Oliveira, Kathia Marçal and Ashok, Ashwin},
	editor = {Abdelnour Nocera, José and Kristín Lárusdóttir, Marta and Petrie, Helen and Piccinno, Antonio and Winckler, Marco},
	year = {2023},
	doi = {10.1007/978-3-031-42280-5_21},
	note = {Series Title: Lecture Notes in Computer Science},
	pages = {349--358},
}

@article{bozorgpanahExplainableMachineLearning2024,
	title = {Explainable machine learning models with privacy},
	volume = {13},
	issn = {2192-6352, 2192-6360},
	url = {https://link.springer.com/10.1007/s13748-024-00315-2},
	doi = {10.1007/s13748-024-00315-2},
	language = {en},
	number = {1},
	urldate = {2024-05-01},
	journal = {Progress in Artificial Intelligence},
	author = {Bozorgpanah, Aso and Torra, Vicenç},
	month = mar,
	year = {2024},
	pages = {31--50},
}

@article{zhaoEfficientPrivacypreservingTreebased2023,
	title = {Efficient and privacy-preserving tree-based inference via additive homomorphic encryption},
	volume = {650},
	issn = {00200255},
	url = {https://linkinghub.elsevier.com/retrieve/pii/S0020025523010654},
	doi = {10.1016/j.ins.2023.119480},
	language = {en},
	urldate = {2024-05-01},
	journal = {Information Sciences},
	author = {Zhao, Jiaqi and Zhu, Hui and Wang, Fengwei and Lu, Rongxing and Li, Hui},
	month = dec,
	year = {2023},
	pages = {119480},
}

@article{kuppaAdversarialXAIMethods2021,
	title = {Adversarial {XAI} {Methods} in {Cybersecurity}},
	volume = {16},
	copyright = {https://creativecommons.org/licenses/by-nc-nd/4.0/},
	issn = {1556-6013, 1556-6021},
	url = {https://ieeexplore.ieee.org/document/9555622/},
	doi = {10.1109/TIFS.2021.3117075},
	language = {en},
	urldate = {2024-05-01},
	journal = {IEEE Transactions on Information Forensics and Security},
	author = {Kuppa, Aditya and Le-Khac, Nhien-An},
	year = {2021},
	pages = {4924--4938},
}

@incollection{mochaourabDemonstratorCounterfactualExplanations2023,
	address = {Cham},
	title = {Demonstrator on {Counterfactual} {Explanations} for {Differentially} {Private} {Support} {Vector} {Machines}},
	volume = {13718},
	isbn = {978-3-031-26421-4 978-3-031-26422-1},
	url = {https://link.springer.com/10.1007/978-3-031-26422-1_52},
	language = {en},
	urldate = {2024-05-02},
	booktitle = {Machine {Learning} and {Knowledge} {Discovery} in {Databases}},
	publisher = {Springer Nature Switzerland},
	author = {Mochaourab, Rami and Sinha, Sugandh and Greenstein, Stanley and Papapetrou, Panagiotis},
	editor = {Amini, Massih-Reza and Canu, Stéphane and Fischer, Asja and Guns, Tias and Kralj Novak, Petra and Tsoumakas, Grigorios},
	year = {2023},
	doi = {10.1007/978-3-031-26422-1_52},
	note = {Series Title: Lecture Notes in Computer Science},
	pages = {662--666},
}

@article{martonExplainingNeuralNetworks2024,
	title = {Explaining neural networks without access to training data},
	issn = {0885-6125, 1573-0565},
	url = {https://link.springer.com/10.1007/s10994-023-06428-4},
	doi = {10.1007/s10994-023-06428-4},
	language = {en},
	urldate = {2024-05-02},
	journal = {Machine Learning},
	author = {Marton, Sascha and Lüdtke, Stefan and Bartelt, Christian and Tschalzev, Andrej and Stuckenschmidt, Heiner},
	month = jan,
	year = {2024},
}

@inproceedings{brunotteContextContentConsent2023,
	title = {Context, {Content}, {Consent} - {How} to {Design} {User}-{Centered} {Privacy} {Explanations} ({S})},
	url = {http://ksiresearchorg.ipage.com/seke/seke23paper/paper032.pdf},
	doi = {10.18293/SEKE2023-032},
	language = {en},
	urldate = {2024-05-03},
	author = {Brunotte, Wasja and Droste, Jakob and Schneider, Kurt},
	month = jul,
	year = {2023},
	pages = {86--89},
}

@article{triccoPRISMAExtensionScoping2018,
	title = {{PRISMA} {Extension} for {Scoping} {Reviews} ({PRISMA}-{ScR}): {Checklist} and {Explanation}},
	volume = {169},
	issn = {0003-4819, 1539-3704},
	shorttitle = {{PRISMA} {Extension} for {Scoping} {Reviews} ({PRISMA}-{ScR})},
	url = {https://www.acpjournals.org/doi/10.7326/M18-0850},
	doi = {10.7326/M18-0850},
	language = {en},
	number = {7},
	urldate = {2024-05-03},
	journal = {Annals of Internal Medicine},
	author = {Tricco, Andrea C. and Lillie, Erin and Zarin, Wasifa and O'Brien, Kelly K. and Colquhoun, Heather and Levac, Danielle and Moher, David and Peters, Micah D.J. and Horsley, Tanya and Weeks, Laura and Hempel, Susanne and Akl, Elie A. and Chang, Christine and McGowan, Jessie and Stewart, Lesley and Hartling, Lisa and Aldcroft, Adrian and Wilson, Michael G. and Garritty, Chantelle and Lewin, Simon and Godfrey, Christina M. and Macdonald, Marilyn T. and Langlois, Etienne V. and Soares-Weiser, Karla and Moriarty, Jo and Clifford, Tammy and Tunçalp, Özge and Straus, Sharon E.},
	month = oct,
	year = {2018},
	pages = {467--473},
}

@misc{elsevierb.v.EngineeringVillage,
	title = {Engineering Village: Search \& Discovery Platform},
	url = {https://www.engineeringvillage.com/},
	urldate = {2024-05-02},
	author = {{Engineering Village}},
}

@misc{Covidence,
	title = {Covidence Systematic Review Software},
	url = {https://www.covidence.org},
	urldate = {2024-05-05},
    author = {{Covidence}},
}

@article{nguyenXRandDifferentiallyPrivate2023,
	title = {{XRand}: {Differentially} {Private} {Defense} against {Explanation}-{Guided} {Attacks}},
	volume = {37},
	issn = {2374-3468, 2159-5399},
	shorttitle = {{XRand}},
	url = {https://ojs.aaai.org/index.php/AAAI/article/view/26401},
	doi = {10.1609/aaai.v37i10.26401},
	language = {en},
	number = {10},
	urldate = {2024-05-05},
	journal = {Proceedings of the AAAI Conference on Artificial Intelligence},
	author = {Nguyen, Truc and Lai, Phung and Phan, Hai and Thai, My T.},
	month = jun,
	year = {2023},
	pages = {11873--11881},
}

@article{elzeinPrivaTreeCollaborativePrivacyPreserving2024,
	title = {{PrivaTree}: {Collaborative} {Privacy}-{Preserving} {Training} of {Decision} {Trees} on {Biomedical} {Data}},
	volume = {21},
	copyright = {https://ieeexplore.ieee.org/Xplorehelp/downloads/license-information/IEEE.html},
	issn = {1545-5963, 1557-9964, 2374-0043},
	shorttitle = {{PrivaTree}},
	url = {https://ieeexplore.ieee.org/document/10153643/},
	doi = {10.1109/TCBB.2023.3286274},
	language = {en},
	number = {1},
	urldate = {2024-05-05},
	journal = {IEEE/ACM Transactions on Computational Biology and Bioinformatics},
	author = {El Zein, Yamane and Lemay, Mathieu and Huguenin, Kévin},
	month = jan,
	year = {2024},
	pages = {1--13},
}

@inproceedings{noriAccuracyInterpretabilityDifferential2021,
	title = {Accuracy, {Interpretability}, and {Differential} {Privacy} via {Explainable} {Boosting}},
	volume = {139},
	url = {https://proceedings.mlr.press/v139/nori21a.html},
	language = {en},
	booktitle = {Proceedings of the 38th {International} {Conference} on {Machine} {Learning}},
	author = {Nori, Harsha and Caruana, Rich and Bu, Zhiqi and Shen, Judy Hanwen and Kulkarni, Janardhan},
	year = {2021},
	pages = {8227--8237},
}

@article{cambriaSurveyXAINatural2023,
	title = {A survey on {XAI} and natural language explanations},
	volume = {60},
	issn = {03064573},
	url = {https://linkinghub.elsevier.com/retrieve/pii/S0306457322002126},
	doi = {10.1016/j.ipm.2022.103111},
	language = {en},
	number = {1},
	urldate = {2024-05-06},
	journal = {Information Processing \& Management},
	author = {Cambria, Erik and Malandri, Lorenzo and Mercorio, Fabio and Mezzanzanica, Mario and Nobani, Navid},
	month = jan,
	year = {2023},
	pages = {103111},
}

@article{wangRationalityExplanationHuman2024,
	title = {The rationality of explanation or human capacity? {Understanding} the impact of explainable artificial intelligence on human-{AI} trust and decision performance},
	volume = {61},
	issn = {03064573},
	shorttitle = {The rationality of explanation or human capacity?},
	url = {https://linkinghub.elsevier.com/retrieve/pii/S030645732400092X},
	doi = {10.1016/j.ipm.2024.103732},
	language = {en},
	number = {4},
	urldate = {2024-05-06},
	journal = {Information Processing \& Management},
	author = {Wang, Ping and Ding, Heng},
	month = jul,
	year = {2024},
	pages = {103732},
}

@article{aliExplainableArtificialIntelligence2023,
	title = {Explainable {Artificial} {Intelligence} ({XAI}): {What} we know and what is left to attain {Trustworthy} {Artificial} {Intelligence}},
	volume = {99},
	issn = {15662535},
	shorttitle = {Explainable {Artificial} {Intelligence} ({XAI})},
	url = {https://linkinghub.elsevier.com/retrieve/pii/S1566253523001148},
	doi = {10.1016/j.inffus.2023.101805},
	language = {en},
	urldate = {2024-05-06},
	journal = {Information Fusion},
	author = {Ali, Sajid and Abuhmed, Tamer and El-Sappagh, Shaker and Muhammad, Khan and Alonso-Moral, Jose M. and Confalonieri, Roberto and Guidotti, Riccardo and Del Ser, Javier and Díaz-Rodríguez, Natalia and Herrera, Francisco},
	month = nov,
	year = {2023},
	pages = {101805},
}

@article{zhangExplainableArtificialIntelligence2022,
	title = {An explainable artificial intelligence approach for financial distress prediction},
	volume = {59},
	issn = {03064573},
	url = {https://linkinghub.elsevier.com/retrieve/pii/S0306457322001030},
	doi = {10.1016/j.ipm.2022.102988},
	language = {en},
	number = {4},
	urldate = {2024-05-06},
	journal = {Information Processing \& Management},
	author = {Zhang, Zijiao and Wu, Chong and Qu, Shiyou and Chen, Xiaofang},
	month = jul,
	year = {2022},
	pages = {102988},
}

@article{karimiSurveyAlgorithmicRecourse2023,
	title = {A {Survey} of {Algorithmic} {Recourse}: {Contrastive} {Explanations} and {Consequential} {Recommendations}},
	volume = {55},
	issn = {0360-0300, 1557-7341},
	shorttitle = {A {Survey} of {Algorithmic} {Recourse}},
	url = {https://dl.acm.org/doi/10.1145/3527848},
	doi = {10.1145/3527848},
	language = {en},
	number = {5},
	urldate = {2024-05-07},
	journal = {ACM Computing Surveys},
	author = {Karimi, Amir-Hossein and Barthe, Gilles and Schölkopf, Bernhard and Valera, Isabel},
	month = may,
	year = {2023},
	pages = {1--29},
}

@article{elmestariPreservingDataPrivacy2024,
	title = {Preserving data privacy in machine learning systems},
	volume = {137},
	issn = {01674048},
	url = {https://linkinghub.elsevier.com/retrieve/pii/S0167404823005151},
	doi = {10.1016/j.cose.2023.103605},
	language = {en},
	urldate = {2024-08-06},
	journal = {Computers \& Security},
	author = {El Mestari, Soumia Zohra and Lenzini, Gabriele and Demirci, Huseyin},
	month = feb,
	year = {2024},
	pages = {103605},
}

@inproceedings{zarifzadehLowCostHighPowerMembership2024,
	title = {Low-{Cost} {High}-{Power} {Membership} {Inference} {Attacks}},
	url = {https://openreview.net/forum?id=sT7UJh5CTc},
	language = {en},
	author = {Zarifzadeh, Sajjad and Liu, Philippe and Shokri, Reza},
	year = {2024},
}

@inproceedings{boIncrementalXAIMemorable2024,
	address = {Honolulu HI USA},
	title = {Incremental {XAI}: {Memorable} {Understanding} of {AI} with {Incremental} {Explanations}},
	isbn = {979-8-4007-0330-0},
	shorttitle = {Incremental {XAI}},
	url = {https://dl.acm.org/doi/10.1145/3613904.3642689},
	doi = {10.1145/3613904.3642689},
	language = {en},
	urldate = {2024-08-10},
	booktitle = {Proceedings of the {CHI} {Conference} on {Human} {Factors} in {Computing} {Systems}},
	publisher = {ACM},
	author = {Bo, Jessica Y and Hao, Pan and Lim, Brian Y},
	month = may,
	year = {2024},
	pages = {1--17},
}

@article{pitropakisTaxonomySurveyAttacks2019,
	title = {A taxonomy and survey of attacks against machine learning},
	volume = {34},
	issn = {15740137},
	url = {https://linkinghub.elsevier.com/retrieve/pii/S1574013718303289},
	doi = {10.1016/j.cosrev.2019.100199},
	language = {en},
	urldate = {2025-01-09},
	journal = {Computer Science Review},
	author = {Pitropakis, Nikolaos and Panaousis, Emmanouil and Giannetsos, Thanassis and Anastasiadis, Eleftherios and Loukas, George},
	month = nov,
	year = {2019},
	pages = {100199},
}

@incollection{guerra-manzanaresPrivacyPreservingMachineLearning2023,
	address = {Cham},
	title = {Privacy-{Preserving} {Machine} {Learning} for {Healthcare}: {Open} {Challenges} and {Future} {Perspectives}},
	volume = {13932},
	isbn = {978-3-031-39538-3 978-3-031-39539-0},
	shorttitle = {Privacy-{Preserving} {Machine} {Learning} for {Healthcare}},
	url = {https://link.springer.com/10.1007/978-3-031-39539-0_3},
	language = {en},
	urldate = {2025-01-11},
	booktitle = {Trustworthy {Machine} {Learning} for {Healthcare}},
	publisher = {Springer Nature Switzerland},
	author = {Guerra-Manzanares, Alejandro and Lopez, L. Julian Lechuga and Maniatakos, Michail and Shamout, Farah E.},
	editor = {Chen, Hao and Luo, Luyang},
	year = {2023},
	doi = {10.1007/978-3-031-39539-0_3},
	note = {Series Title: Lecture Notes in Computer Science},
	pages = {25--40},
}

@inproceedings{carliniMembershipInferenceAttacks2022,
	address = {San Francisco, CA, USA},
	title = {Membership {Inference} {Attacks} {From} {First} {Principles}},
	copyright = {https://doi.org/10.15223/policy-009},
	isbn = {978-1-6654-1316-9},
	url = {https://ieeexplore.ieee.org/document/9833649/},
	doi = {10.1109/SP46214.2022.9833649},
	language = {en},
	urldate = {2025-01-13},
	booktitle = {2022 {IEEE} {Symposium} on {Security} and {Privacy} ({SP})},
	publisher = {IEEE},
	author = {Carlini, Nicholas and Chien, Steve and Nasr, Milad and Song, Shuang and Terzis, Andreas and Tramèr, Florian},
	month = may,
	year = {2022},
	pages = {1897--1914},
}

@inproceedings{choquette-chooLabelOnlyMembershipInference2021,
	series = {Proceedings of {Machine} {Learning} {Research}},
	title = {Label-{Only} {Membership} {Inference} {Attacks}},
	volume = {139},
	url = {https://proceedings.mlr.press/v139/choquette-choo21a.html},
	language = {en},
	booktitle = {Proceedings of the 38th {International} {Conference} on {Machine} {Learning}},
	publisher = {PMLR},
	author = {Choquette-Choo, Christopher A and Tramer, Florian and Carlini, Nicholas and Papernot, Nicolas},
	month = jul,
	year = {2021},
	pages = {1964--1974},
}

@article{maLabelOnlyMembershipInference2024,
	title = {Label-{Only} {Membership} {Inference} {Attack} {Based} on {Model} {Explanation}},
	volume = {56},
	issn = {1573-773X},
	url = {https://link.springer.com/10.1007/s11063-024-11682-1},
	doi = {10.1007/s11063-024-11682-1},
	language = {en},
	number = {5},
	urldate = {2025-01-14},
	journal = {Neural Processing Letters},
	author = {Ma, Yao and Zhai, Xurong and Yu, Dan and Yang, Yuli and Wei, Xingyu and Chen, Yongle},
	month = sep,
	year = {2024},
	pages = {236},
}

@inproceedings{cohenMembershipInferenceAttack2024,
	address = {Waikoloa, HI, USA},
	title = {Membership {Inference} {Attack} {Using} {Self} {Influence} {Functions}},
	copyright = {https://doi.org/10.15223/policy-029},
	isbn = {979-8-3503-1892-0},
	url = {https://ieeexplore.ieee.org/document/10484386/},
	doi = {10.1109/WACV57701.2024.00482},
	language = {en},
	urldate = {2025-01-14},
	booktitle = {2024 {IEEE}/{CVF} {Winter} {Conference} on {Applications} of {Computer} {Vision} ({WACV})},
	publisher = {IEEE},
	author = {Cohen, Gilad and Giryes, Raja},
	month = jan,
	year = {2024},
	pages = {4880--4889},
}

@inproceedings{ferryProbabilisticDatasetReconstruction2024,
	address = {Toronto, ON, Canada},
	title = {Probabilistic {Dataset} {Reconstruction} from {Interpretable} {Models}},
	copyright = {https://doi.org/10.15223/policy-029},
	isbn = {979-8-3503-4950-4},
	url = {https://ieeexplore.ieee.org/document/10516646/},
	doi = {10.1109/SaTML59370.2024.00009},
	language = {en},
	urldate = {2025-01-15},
	booktitle = {2024 {IEEE} {Conference} on {Secure} and {Trustworthy} {Machine} {Learning} ({SaTML})},
	publisher = {IEEE},
	author = {Ferry, Julien and Aïvodji, Ulrich and Gambs, Sébastien and Huguet, Marie-José and Siala, Mohamed},
	month = apr,
	year = {2024},
	pages = {1--17},
}

@inproceedings{miuraMEGEXDataFreeModel2024,
	address = {Singapore Singapore},
	title = {{MEGEX}: {Data}-{Free} {Model} {Extraction} {Attack} {Against} {Gradient}-{Based} {Explainable} {AI}},
	isbn = {979-8-4007-0691-2},
	shorttitle = {{MEGEX}},
	url = {https://dl.acm.org/doi/10.1145/3665451.3665533},
	doi = {10.1145/3665451.3665533},
	language = {en},
	urldate = {2025-01-18},
	booktitle = {Proceedings of the 2nd {ACM} {Workshop} on {Secure} and {Trustworthy} {Deep} {Learning} {Systems}},
	publisher = {ACM},
	author = {Miura, Takayuki and Shibahara, Toshiki and Yanai, Naoto},
	month = jul,
	year = {2024},
	pages = {56--66},
}

@article{tranComprehensiveSurveyTaxonomy2024,
	title = {A comprehensive survey and taxonomy on privacy-preserving deep learning},
	volume = {576},
	issn = {09252312},
	url = {https://linkinghub.elsevier.com/retrieve/pii/S0925231224001164},
	doi = {10.1016/j.neucom.2024.127345},
	language = {en},
	urldate = {2025-01-23},
	journal = {Neurocomputing},
	author = {Tran, Anh-Tu and Luong, The-Dung and Huynh, Van-Nam},
	month = apr,
	year = {2024},
	pages = {127345},
}

@article{yinComprehensiveSurveyPrivacypreserving2022,
	title = {A {Comprehensive} {Survey} of {Privacy}-preserving {Federated} {Learning}: {A} {Taxonomy}, {Review}, and {Future} {Directions}},
	volume = {54},
	issn = {0360-0300, 1557-7341},
	shorttitle = {A {Comprehensive} {Survey} of {Privacy}-preserving {Federated} {Learning}},
	url = {https://dl.acm.org/doi/10.1145/3460427},
	doi = {10.1145/3460427},
	language = {en},
	number = {6},
	urldate = {2025-01-23},
	journal = {ACM Computing Surveys},
	author = {Yin, Xuefei and Zhu, Yanming and Hu, Jiankun},
	month = jul,
	year = {2022},
	pages = {1--36},
}

@inproceedings{severiExplanationGuidedBackdoorPoisoning2021,
	title = {Explanation-{Guided} {Backdoor} {Poisoning} {Attacks} {Against} {Malware} {Classiﬁers}},
	language = {en},
	booktitle = {30th {USENIX} {Security} {Symposium} ({USENIX} {Security} 21)},
	author = {Severi, Giorgio and Meyer, Jim and Oprea, Alina and Coull, Scott},
	year = {2021},
}

@article{fletcherDecisionTreeClassification2020,
	title = {Decision {Tree} {Classification} with {Differential} {Privacy}: {A} {Survey}},
	volume = {52},
	issn = {0360-0300, 1557-7341},
	shorttitle = {Decision {Tree} {Classification} with {Differential} {Privacy}},
	url = {https://dl.acm.org/doi/10.1145/3337064},
	doi = {10.1145/3337064},
	language = {en},
	number = {4},
	urldate = {2025-01-24},
	journal = {ACM Computing Surveys},
	author = {Fletcher, Sam and Islam, Md. Zahidul},
	month = jul,
	year = {2020},
	pages = {1--33},
}

@inproceedings{liDifferentialPrivacyPreservation2020,
	address = {Guangzhou, China},
	title = {Differential {Privacy} {Preservation} in {Interpretable} {Feedforward}-{Designed} {Convolutional} {Neural} {Networks}},
	copyright = {https://ieeexplore.ieee.org/Xplorehelp/downloads/license-information/IEEE.html},
	isbn = {978-1-6654-0392-4},
	url = {https://ieeexplore.ieee.org/document/9343157/},
	doi = {10.1109/TrustCom50675.2020.00089},
	language = {en},
	urldate = {2025-01-24},
	booktitle = {2020 {IEEE} 19th {International} {Conference} on {Trust}, {Security} and {Privacy} in {Computing} and {Communications} ({TrustCom})},
	publisher = {IEEE},
	author = {Li, De and Wang, Jinyan and Tan, Zhou and Li, Xianxian and Hu, Yuhang},
	month = dec,
	year = {2020},
	pages = {631--638},
}

@article{baekDifferentiallyPrivateExplainable2024,
	title = {Differentially private and explainable boosting machine with enhanced utility},
	volume = {607},
	issn = {09252312},
	url = {https://linkinghub.elsevier.com/retrieve/pii/S0925231224011950},
	doi = {10.1016/j.neucom.2024.128424},
	language = {en},
	urldate = {2025-01-27},
	journal = {Neurocomputing},
	author = {Baek, Incheol and Chung, Yon Dohn},
	month = nov,
	year = {2024},
	pages = {128424},
}

@article{sweeneyAchievingKanonymityPrivacy2002,
	title = {Achieving k-anonymity privacy protection using generalization and suppression},
	volume = {10},
	issn = {0218-4885},
	number = {05},
	journal = {International Journal of Uncertainty, Fuzziness and Knowledge-Based Systems},
	author = {Sweeney, Latanya},
	year = {2002},
	note = {Publisher: World Scientific},
	pages = {571--588},
}

@article{sweeneyKanonymityModelProtecting2002,
	title = {k-anonymity: {A} model for protecting privacy},
	volume = {10},
	issn = {0218-4885},
	number = {05},
	journal = {International journal of uncertainty, fuzziness and knowledge-based systems},
	author = {Sweeney, Latanya},
	year = {2002},
	note = {Publisher: World Scientific},
	pages = {557--570},
}

@article{montenegroAnonymizingMedicalCasebased2024,
	title = {Anonymizing medical case-based explanations through disentanglement},
	volume = {95},
	issn = {13618415},
	url = {https://linkinghub.elsevier.com/retrieve/pii/S1361841524001348},
	doi = {10.1016/j.media.2024.103209},
	language = {en},
	urldate = {2025-02-04},
	journal = {Medical Image Analysis},
	author = {Montenegro, Helena and Cardoso, Jaime S.},
	month = jul,
	year = {2024},
	pages = {103209},
}

@inproceedings{johnsonHeartSpotPrivatizedExplainable2022,
	address = {Ioannina, Greece},
	title = {{HeartSpot}: {Privatized} and {Explainable} {Data} {Compression} for {Cardiomegaly} {Detection}},
	copyright = {https://doi.org/10.15223/policy-029},
	isbn = {978-1-6654-8791-7},
	shorttitle = {{HeartSpot}},
	url = {https://ieeexplore.ieee.org/document/9926777/},
	doi = {10.1109/BHI56158.2022.9926777},
	language = {en},
	urldate = {2025-02-04},
	booktitle = {2022 {IEEE}-{EMBS} {International} {Conference} on {Biomedical} and {Health} {Informatics} ({BHI})},
	publisher = {IEEE},
	author = {Johnson, Elvin and Mohan, Shreshta and Gaudio, Alex and Smailagic, Asim and Faloutsos, Christos and Campilho, Aurelio},
	month = sep,
	year = {2022},
	pages = {01--04},
}

@article{anCounterfactualExplanationWill2024,
	title = {Counterfactual {Explanation} at {Will}, with {Zero} {Privacy} {Leakage}},
	volume = {2},
	issn = {2836-6573},
	url = {https://dl.acm.org/doi/10.1145/3654933},
	doi = {10.1145/3654933},
	language = {en},
	number = {3},
	urldate = {2025-02-05},
	journal = {Proceedings of the ACM on Management of Data},
	author = {An, Shuai and Cao, Yang},
	month = may,
	year = {2024},
	pages = {1--29},
}

@article{zhangSurveyTrustworthyFederated2024,
	title = {A {Survey} of {Trustworthy} {Federated} {Learning}: {Issues}, {Solutions}, and {Challenges}},
	volume = {15},
	issn = {2157-6904, 2157-6912},
	shorttitle = {A {Survey} of {Trustworthy} {Federated} {Learning}},
	url = {https://dl.acm.org/doi/10.1145/3678181},
	doi = {10.1145/3678181},
	language = {en},
	number = {6},
	urldate = {2025-02-06},
	journal = {ACM Transactions on Intelligent Systems and Technology},
	author = {Zhang, Yifei and Zeng, Dun and Luo, Jinglong and Fu, Xinyu and Chen, Guanzhong and Xu, Zenglin and King, Irwin},
	month = dec,
	year = {2024},
	pages = {1--47},
}

@inproceedings{camposLatentDiffusionModels2024,
	address = {Santiago de Compostela; Spain},
	title = {Latent {Diffusion} {Models} for {Privacy}-preserving {Medical} {Case}-based {Explanations}},
	volume = {3831},
	booktitle = {1st {Workshop} on {Explainable} {Artificial} {Intelligence} for the {Medical} {Domain}, {EXPLIMED} 2024},
	publisher = {CEUR-WS},
	author = {Campos, Filipe and Petrychenko, Liliana and Teixeira, Luís F and Silva, Wilson},
	year = {2024},
}

@article{terziyanExplainableAIIndustry2022,
	title = {Explainable {AI} for {Industry} 4.0: {Semantic} {Representation} of {Deep} {Learning} {Models}},
	volume = {200},
	issn = {18770509},
	shorttitle = {Explainable {AI} for {Industry} 4.0},
	url = {https://linkinghub.elsevier.com/retrieve/pii/S1877050922002290},
	doi = {10.1016/j.procs.2022.01.220},
	language = {en},
	urldate = {2025-02-07},
	journal = {Procedia Computer Science},
	author = {Terziyan, Vagan and Vitko, Oleksandra},
	year = {2022},
	pages = {216--226},
}

@inproceedings{molhoekSecureCounterfactualExplanations2024,
	address = {Venice, Italy},
	title = {Secure {Counterfactual} {Explanations} in a {Two}-party {Setting}},
	copyright = {https://doi.org/10.15223/policy-029},
	isbn = {978-1-7377497-6-9},
	url = {https://ieeexplore.ieee.org/document/10706413/},
	doi = {10.23919/FUSION59988.2024.10706413},
	language = {en},
	urldate = {2025-02-07},
	booktitle = {2024 27th {International} {Conference} on {Information} {Fusion} ({FUSION})},
	publisher = {IEEE},
	author = {Molhoek, M. and Laanen, J. Van},
	month = jul,
	year = {2024},
	pages = {1--10},
}

@article{guoADFLPoisoningAttack2022,
	title = {{ADFL}: {A} {Poisoning} {Attack} {Defense} {Framework} for {Horizontal} {Federated} {Learning}},
	volume = {18},
	copyright = {https://ieeexplore.ieee.org/Xplorehelp/downloads/license-information/IEEE.html},
	issn = {1551-3203, 1941-0050},
	shorttitle = {{ADFL}},
	url = {https://ieeexplore.ieee.org/document/9735274/},
	doi = {10.1109/TII.2022.3156645},
	language = {en},
	number = {10},
	urldate = {2025-02-11},
	journal = {IEEE Transactions on Industrial Informatics},
	author = {Guo, Jingjing and Li, Haiyang and Huang, Feiran and Liu, Zhiquan and Peng, Yanguo and Li, Xinghua and Ma, Jianfeng and Menon, Varun G. and Igorevich, Konstantin Kostromitin},
	month = oct,
	year = {2022},
	pages = {6526--6536},
}

@article{guoTFLDTTrustEvaluation2023,
	title = {{TFL}-{DT}: {A} {Trust} {Evaluation} {Scheme} for {Federated} {Learning} in {Digital} {Twin} for {Mobile} {Networks}},
	volume = {41},
	copyright = {https://creativecommons.org/licenses/by/4.0/legalcode},
	issn = {0733-8716, 1558-0008},
	shorttitle = {{TFL}-{DT}},
	url = {https://ieeexplore.ieee.org/document/10234616/},
	doi = {10.1109/JSAC.2023.3310094},
	language = {en},
	number = {11},
	urldate = {2025-02-11},
	journal = {IEEE Journal on Selected Areas in Communications},
	author = {Guo, Jingjing and Liu, Zhiquan and Tian, Siyi and Huang, Feiran and Li, Jiaxing and Li, Xinghua and Igorevich, Kostromitin Konstantin and Ma, Jianfeng},
	month = nov,
	year = {2023},
	pages = {3548--3560},
}

@inproceedings{daoleTrustworthyAIHeterogeneous2024,
	address = {Yokohama, Japan},
	title = {Trustworthy {AI} in {Heterogeneous} {Settings}: {Federated} {Learning} of {Explainable} {Classifiers}},
	copyright = {https://doi.org/10.15223/policy-029},
	isbn = {979-8-3503-1954-5},
	shorttitle = {Trustworthy {AI} in {Heterogeneous} {Settings}},
	url = {https://ieeexplore.ieee.org/document/10612109/},
	doi = {10.1109/FUZZ-IEEE60900.2024.10612109},
	language = {en},
	urldate = {2025-02-11},
	booktitle = {2024 {IEEE} {International} {Conference} on {Fuzzy} {Systems} ({FUZZ}-{IEEE})},
	publisher = {IEEE},
	author = {Daole, Mattia and Ducange, Pietro and Marcelloni, Francesco and Renda, Alessandro},
	month = jun,
	year = {2024},
	pages = {1--9},
}

@incollection{ezzeddineDifferentialPrivacyAnomaly2024,
	address = {Cham},
	title = {Differential {Privacy} for {Anomaly} {Detection}: {Analyzing} the {Trade}-{Off} {Between} {Privacy} and {Explainability}},
	volume = {2155},
	isbn = {978-3-031-63799-5 978-3-031-63800-8},
	shorttitle = {Differential {Privacy} for {Anomaly} {Detection}},
	url = {https://link.springer.com/10.1007/978-3-031-63800-8_15},
	language = {en},
	urldate = {2025-02-12},
	booktitle = {Explainable {Artificial} {Intelligence}},
	publisher = {Springer Nature Switzerland},
	author = {Ezzeddine, Fatima and Saad, Mirna and Ayoub, Omran and Andreoletti, Davide and Gjoreski, Martin and Sbeity, Ihab and Langheinrich, Marc and Giordano, Silvia},
	editor = {Longo, Luca and Lapuschkin, Sebastian and Seifert, Christin},
	year = {2024},
	doi = {10.1007/978-3-031-63800-8_15},
	note = {Series Title: Communications in Computer and Information Science},
	pages = {294--318},
}

@inproceedings{berningTradeoffPrivacyQuality2024,
	address = {Vienna Austria},
	title = {The {Trade}-off {Between} {Privacy} \& {Quality} for {Counterfactual} {Explanations}},
	isbn = {979-8-4007-1718-5},
	url = {https://dl.acm.org/doi/10.1145/3664476.3670897},
	doi = {10.1145/3664476.3670897},
	language = {en},
	urldate = {2025-02-12},
	booktitle = {Proceedings of the 19th {International} {Conference} on {Availability}, {Reliability} and {Security}},
	publisher = {ACM},
	author = {Berning, Sjoerd and Dunning, Vincent and Spagnuelo, Dayana and Veugen, Thijs and Van Der Waa, Jasper},
	month = jul,
	year = {2024},
	pages = {1--9},
}

@inproceedings{abbasiFurtherInsightsBalancing2024,
	address = {Vienna Austria},
	title = {Further {Insights}: {Balancing} {Privacy}, {Explainability}, and {Utility} in {Machine} {Learning}-based {Tabular} {Data} {Analysis}},
	isbn = {979-8-4007-1718-5},
	shorttitle = {Further {Insights}},
	url = {https://dl.acm.org/doi/10.1145/3664476.3670901},
	doi = {10.1145/3664476.3670901},
	language = {en},
	urldate = {2025-02-12},
	booktitle = {Proceedings of the 19th {International} {Conference} on {Availability}, {Reliability} and {Security}},
	publisher = {ACM},
	author = {Abbasi, Wisam and Mori, Paolo and Saracino, Andrea},
	month = jul,
	year = {2024},
	pages = {1--10},
}

@article{liminfuRuleGenerationNeural1994,
	title = {Rule generation from neural networks},
	volume = {24},
	copyright = {https://ieeexplore.ieee.org/Xplorehelp/downloads/license-information/IEEE.html},
	issn = {00189472},
	url = {http://ieeexplore.ieee.org/document/299696/},
	doi = {10.1109/21.299696},
	language = {en},
	number = {8},
	urldate = {2025-02-13},
	journal = {IEEE Transactions on Systems, Man, and Cybernetics},
	author = {{LiMin Fu}},
	month = aug,
	year = {1994},
	pages = {1114--1124},
}

@article{benitezAreArtificialNeural1997,
	title = {Are artificial neural networks black boxes?},
	volume = {8},
	copyright = {https://ieeexplore.ieee.org/Xplorehelp/downloads/license-information/IEEE.html},
	issn = {1045-9227, 1941-0093},
	url = {https://ieeexplore.ieee.org/document/623216/},
	doi = {10.1109/72.623216},
	language = {en},
	number = {5},
	urldate = {2025-02-13},
	journal = {IEEE Transactions on Neural Networks},
	author = {Benitez, J.M. and Castro, J.L. and Requena, I.},
	month = sep,
	year = {1997},
	pages = {1156--1164},
}

@article{milareApproachExplainNeural2002,
	title = {{An} {Approach} {To} {Explain} {Neural} {Networks} {Using} {Symbolic} {Algorithms}},
	volume = {02},
	issn = {1469-0268, 1757-5885},
	url = {https://www.worldscientific.com/doi/abs/10.1142/S1469026802000695},
	doi = {10.1142/S1469026802000695},
	language = {en},
	number = {04},
	urldate = {2025-02-13},
	journal = {International Journal of Computational Intelligence and Applications},
	author = {Milaré, Claudia R. and De L. F. De Carvalho, André C. P. and Monard, Maria C.},
	month = dec,
	year = {2002},
	pages = {365--376},
}

@inproceedings{torresExtractingTreesTrained2005,
	address = {Rio de Janeiro, Brazil},
	title = {Extracting trees from trained {SVM} models using a {TREPAN} based approach},
	isbn = {978-0-7695-2457-3},
	url = {http://ieeexplore.ieee.org/document/1587773/},
	doi = {10.1109/ICHIS.2005.41},
	language = {en},
	urldate = {2025-02-13},
	booktitle = {Fifth {International} {Conference} on {Hybrid} {Intelligent} {Systems} ({HIS}'05)},
	publisher = {IEEE},
	author = {Torres, D.E.D. and Rocco, C.M.S.},
	year = {2005},
	pages = {6 pp.},
}

@misc{gdprArt22GDPR2016,
	title = {Art. 22 {GDPR}},
	url = {https://gdpr-info.eu/art-22-gdpr/},
	journal = {Art. 22 GDPR},
	author = {GDPR},
	month = apr,
	year = {2016},
}

@article{majeedAnonymizationTechniquesPrivacy2021,
	title = {Anonymization {Techniques} for {Privacy} {Preserving} {Data} {Publishing}: {A} {Comprehensive} {Survey}},
	volume = {9},
	copyright = {https://creativecommons.org/licenses/by/4.0/legalcode},
	issn = {2169-3536},
	shorttitle = {Anonymization {Techniques} for {Privacy} {Preserving} {Data} {Publishing}},
	url = {https://ieeexplore.ieee.org/document/9298747/},
	doi = {10.1109/ACCESS.2020.3045700},
	language = {en},
	urldate = {2025-02-14},
	journal = {IEEE Access},
	author = {Majeed, Abdul and Lee, Sungchang},
	year = {2021},
	pages = {8512--8545},
}

@incollection{tomaCombinationsAIModels2024,
	address = {Cham},
	title = {Combinations of {AI} {Models} and {XAI} {Metrics} {Vulnerable} to {Record} {Reconstruction} {Risk}},
	copyright = {https://www.springernature.com/gp/researchers/text-and-data-mining},
	isbn = {978-3-031-69650-3 978-3-031-69651-0},
	url = {https://link.springer.com/10.1007/978-3-031-69651-0_22},
	language = {en},
	urldate = {2025-07-10},
	booktitle = {Lecture {Notes} in {Computer} {Science}},
	publisher = {Springer Nature Switzerland},
	author = {Toma, Ryotaro and Kikuchi, Hiroaki},
	year = {2024},
	doi = {10.1007/978-3-031-69651-0_22},
	note = {ISSN: 0302-9743, 1611-3349},
	pages = {329--343},
}

@article{schneider2024explainable,
  title={Explainable Generative AI (GenXAI): a survey, conceptualization, and research agenda},
  author={Schneider, Johannes},
  journal={Artificial Intelligence Review},
  volume={57},
  number={11},
  pages={289},
  year={2024},
  publisher={Springer}
}

@article{wei2022chain,
  title={Chain-of-thought prompting elicits reasoning in large language models},
  author={Wei, Jason and Wang, Xuezhi and Schuurmans, Dale and Bosma, Maarten and Xia, Fei and Chi, Ed and Le, Quoc V and Zhou, Denny and others},
  journal={Advances in neural information processing systems},
  volume={35},
  pages={24824--24837},
  year={2022}
}

@article{
bereska2024mechanistic,
title={Mechanistic Interpretability for {AI} Safety - A Review},
author={Leonard Bereska and Stratis Gavves},
journal={Transactions on Machine Learning Research},
issn={2835-8856},
year={2024},
url={https://openreview.net/forum?id=ePUVetPKu6},
note={Survey Certification, Expert Certification}
}

@article{sajjad2022neuron,
  title={Neuron-level interpretation of deep nlp models: A survey},
  author={Sajjad, Hassan and Durrani, Nadir and Dalvi, Fahim},
  journal={Transactions of the Association for Computational Linguistics},
  volume={10},
  pages={1285--1303},
  year={2022},
  publisher={MIT Press One Broadway, 12th Floor, Cambridge, Massachusetts 02142, USA~…}
}

@article{nguyen2025privacy,
  title={Privacy-preserving explainable AI: a survey},
  author={Nguyen, Thanh Tam and Huynh, Thanh Trung and Ren, Zhao and Nguyen, Thanh Toan and Nguyen, Phi Le and Yin, Hongzhi and Nguyen, Quoc Viet Hung},
  journal={Science China Information Sciences},
  volume={68},
  number={1},
  pages={111101},
  year={2025},
  publisher={Springer}
}

@article{baniecki2024adversarial,
title = {Adversarial attacks and defenses in explainable artificial intelligence: A survey},
journal = {Information Fusion},
volume = {107},
pages = {102303},
year = {2024},
issn = {1566-2535},
doi = {https://doi.org/10.1016/j.inffus.2024.102303},
url = {https://www.sciencedirect.com/science/article/pii/S1566253524000812},
author = {Hubert Baniecki and Przemyslaw Biecek},
}

@InProceedings{Noppel2024adversarial,
author="Noppel, Maximilian
and Wressnegger, Christian",
editor="Hotho, Andreas
and Rudolph, Sebastian",
title="A Brief Systematization of Explanation-Aware Attacks",
booktitle="KI 2024: Advances in Artificial Intelligence",
year="2024",
publisher="Springer Nature Switzerland",
address="Cham",
pages="350--354",
isbn="978-3-031-70893-0"
}

@inproceedings{Noppel2024SoK,
  author={Noppel, Maximilian and Wressnegger, Christian},
  booktitle={2024 IEEE Symposium on Security and Privacy (SP)}, 
  title={SoK: Explainable Machine Learning in Adversarial Environments}, 
  year={2024},
  volume={},
  number={},
  pages={2441-2459},
  doi={10.1109/SP54263.2024.00021}}

@inproceedings{kim2018interpretability,
  title={Interpretability beyond feature attribution: Quantitative testing with concept activation vectors (tcav)},
  author={Kim, Been and Wattenberg, Martin and Gilmer, Justin and Cai, Carrie and Wexler, James and Viegas, Fernanda and others},
  booktitle={International conference on machine learning},
  pages={2668--2677},
  year={2018},
  organization={PMLR}
}

@article{crabbe2022concept,
  title={Concept activation regions: A generalized framework for concept-based explanations},
  author={Crabb{\'e}, Jonathan and van der Schaar, Mihaela},
  journal={Advances in Neural Information Processing Systems},
  volume={35},
  pages={2590--2607},
  year={2022}
}

@article{ghorbani2019towards,
  title={Towards automatic concept-based explanations},
  author={Ghorbani, Amirata and Wexler, James and Zou, James Y and Kim, Been},
  journal={Advances in neural information processing systems},
  volume={32},
  year={2019}
}

@inproceedings{koh2020concept,
  title={Concept bottleneck models},
  author={Koh, Pang Wei and Nguyen, Thao and Tang, Yew Siang and Mussmann, Stephen and Pierson, Emma and Kim, Been and Liang, Percy},
  booktitle={International conference on machine learning},
  pages={5338--5348},
  year={2020},
  organization={PMLR}
}

@article{olah2022mechanistic,
  title = {Mechanistic Interpretability, Variables, and the Importance of Interpretable Bases},
  journal={Transformer Circuits Thread},
  url = {https://transformer-circuits.pub/2022/mech-interp-essay/index.html},
  author= {Olah, Chris},
  year={2022},
}

@article{belinkov2022probes,
    author = {Belinkov, Yonatan},
    title = {Probing Classifiers: Promises, Shortcomings, and Advances},
    journal = {Computational Linguistics},
    volume = {48},
    number = {1},
    pages = {207-219},
    year = {2022},
    month = {04},
    issn = {0891-2017},
    doi = {10.1162/coli_a_00422},
    url = {https://doi.org/10.1162/coli_a_00422},
    eprint = {https://direct.mit.edu/coli/article-pdf/48/1/207/2006605/coli_a_00422.pdf},
}

@article{dalvi2019neuronact, 
title={What Is One Grain of Sand in the Desert? Analyzing Individual Neurons in Deep NLP Models}, 
volume={33}, 
url={https://ojs.aaai.org/index.php/AAAI/article/view/4592}, 
DOI={10.1609/aaai.v33i01.33016309}, 
number={01}, 
journal={Proceedings of the AAAI Conference on Artificial Intelligence}, 
author={Dalvi, Fahim and Durrani, Nadir and Sajjad, Hassan and Belinkov, Yonatan and Bau, Anthony and Glass, James}, 
year={2019}, 
month={07}, 
pages={6309-6317} 
}

@inproceedings{antverg2022neuronact,
  title={On the Pitfalls of Analyzing Individual Neurons in Language Models},
  author={Antverg, Omer and Belinkov, Yonatan},
  booktitle={International Conference on Learning Representations},
  year={2022}
}

@article{zimmermann2021mechint,
  title={How well do feature visualizations support causal understanding of CNN activations?},
  author={Zimmermann, Roland S and Borowski, Judy and Geirhos, Robert and Bethge, Matthias and Wallis, Thomas and Brendel, Wieland},
  journal={Advances in Neural Information Processing Systems},
  volume={34},
  pages={11730--11744},
  year={2021}
}

@inproceedings{alain2017probing,
language = {English},
copyright = {Compilation and indexing terms, Copyright 2025 Elsevier Inc.},
copyright = {Compendex},
title = {UNDERSTANDING INTERMEDIATE LAYERS USING LINEAR CLASSIFIER PROBES},
journal = {5th International Conference on Learning Representations, ICLR 2017 - Workshop Track Proceedings},
author = {Alain, Guillaume and Bengio, Yoshua},
year = {2017},
address = {Toulon, France},
note = {Black boxes;Hidden units;Intermediate layers;Linear classifiers;Middle layer;Neural network model;Potential problems;Training phasis;},
}

@inproceedings{cunningham2024sae,
language = {English},
copyright = {Compilation and indexing terms, Copyright 2025 Elsevier Inc.},
copyright = {Compendex},
title = {Sparse Autoencoders Find Highly Interpretable Features in Language Models},
journal = {12th International Conference on Learning Representations, ICLR 2024},
author = {Cunningham, Hoagy and Ewart, Aidan and Riggs, Logan and Huben, Robert and Sharkey, Lee},
year = {2024},
pages = {et al; Google Deepmind; Google Research; Meta; Microsoft - },
address = {Hybrid, Vienna, Austria},
key = {Neurons},
note = {Auto encoders;Automated methods;Counterfactuals;Interpretability;Language model;Learn+;Neural-networks;Object identification;Over-complete;Sets of features;},
}

@article{heo2019fooling,
  title={Fooling neural network interpretations via adversarial model manipulation},
  author={Heo, Juyeon and Joo, Sunghwan and Moon, Taesup},
  journal={Advances in neural information processing systems},
  volume={32},
  year={2019}
}

@article{dombrowski2019explanations,
  title={Explanations can be manipulated and geometry is to blame},
  author={Dombrowski, Ann-Kathrin and Alber, Maximillian and Anders, Christopher and Ackermann, Marcel and M{\"u}ller, Klaus-Robert and Kessel, Pan},
  journal={Advances in neural information processing systems},
  volume={32},
  year={2019}
}

@inproceedings{noppel2023backdoor,
  author={Noppel, Maximilian and Peter, Lukas and Wressnegger, Christian},
  booktitle={2023 IEEE Symposium on Security and Privacy (SP)}, 
  title={Disguising Attacks with Explanation-Aware Backdoors}, 
  year={2023},
  volume={},
  number={},
  pages={664-681},
  doi={10.1109/SP46215.2023.10179308}}

@inproceedings{zhang2020interpretable,
  title={Interpretable deep learning under fire},
  author={Zhang, Xinyang and Wang, Ningfei and Shen, Hua and Ji, Shouling and Luo, Xiapu and Wang, Ting},
  booktitle={29th $\{$USENIX$\}$ security symposium ($\{$USENIX$\}$ security 20)},
  year={2020}
}

@inproceedings{spartalis2023balancing,
author = {Spartalis, Christoforos N. and Semertzidis, Theodoros and Daras, Petros},
title = {Balancing XAI with Privacy and Security Considerations},
year = {2023},
isbn = {978-3-031-54128-5},
publisher = {Springer-Verlag},
address = {Berlin, Heidelberg},
url = {https://doi.org/10.1007/978-3-031-54129-2_7},
doi = {10.1007/978-3-031-54129-2_7},
booktitle = {Computer Security. ESORICS 2023 International Workshops: CPS4CIP, ADIoT, SecAssure, WASP, TAURIN, PriST-AI, and SECAI, The Hague, The Netherlands, September 25–29, 2023, Revised Selected Papers, Part II},
pages = {111–124},
numpages = {14},
location = {The Hague, The Netherlands}
}

@article{bae2025privacy,
  title={Privacy-Preserving LLM Interaction with Socratic Chain-of-Thought Reasoning and Homomorphically Encrypted Vector Databases},
  author={Bae, Yubeen and Kim, Minchan and Lee, Jaejin and Kim, Sangbum and Kim, Jaehyung and Choi, Yejin and Mireshghallah, Niloofar},
  journal={arXiv preprint arXiv:2506.17336},
  year={2025}
}
\bibliographystyle{tmlr}

\end{document}